%% file: sample-sigconf-authordraft.tex
\documentclass[sigconf]{acmart}

\AtBeginDocument{%
  }

\setcopyright{None}
\renewcommand\footnotetextcopyrightpermission[1]{}

\usepackage{xcolor}
\usepackage{array}
\usepackage{multirow}
\usepackage{amsmath}
\usepackage{bbold}

\begin{document}

\fancyhead{}
\title{DUDA: Distilled Unsupervised Domain Adaptation for Lightweight Semantic Segmentation}



\author{Beomseok Kang}
\affiliation{
  \institution{Georgia Institute of Technology}
  \city{Atlanta}
  \state{GA}
  \country{USA}}
\email{beomseok@gatech.edu}

\author{Niluthpol Chowdhury Mithun, Abhinav Rajvanshi, Han-Pang Chiu, Supun Samarasekera}
\affiliation{
  \institution{SRI International}
  \city{Princeton}
  \state{NJ}
  \country{USA}}
\email{{firstname.lastname}@sri.com}

\renewcommand{\shortauthors}{Kang et al.}

\input{sec/0_abstract}






\maketitle

\input{sec/1_introduction}
\input{sec/2_related_works}

\input{sec/3_proposed_methods}

\input{sec/4_experimental_results}
\input{sec/5_conclusion}



\bibliographystyle{ACM-Reference-Format}
\bibliography{sample-base}

\clearpage
\appendix
\input{sec/suppl}

\end{document}

%% file: sec/0_abstract.tex
\begin{abstract}
Unsupervised Domain Adaptation (UDA) is essential for enabling semantic segmentation in new domains without requiring costly pixel-wise annotations. State-of-the-art (SOTA) UDA methods primarily use self-training with architecturally identical teacher and student networks, relying on Exponential Moving Average (EMA) updates. However, these approaches face substantial performance degradation with lightweight models due to inherent architectural inflexibility leading to low-quality pseudo-labels. To address this, we propose Distilled Unsupervised Domain Adaptation (DUDA), a novel framework that combines EMA-based self-training with knowledge distillation (KD). Our method employs an auxiliary student network to bridge the architectural gap between heavyweight and lightweight models for EMA-based updates, resulting in improved pseudo-label quality. DUDA employs a strategic fusion of UDA and KD, incorporating innovative elements such as gradual distillation from large to small networks, inconsistency loss prioritizing poorly adapted classes, and learning with multiple teachers. Extensive experiments across four UDA benchmarks demonstrate DUDA's superiority in achieving SOTA performance with lightweight models, often surpassing the performance of heavyweight models from other approaches. 
\end{abstract}

%% file: sec/1_introduction.tex
\section{Introduction}
\label{sec:introduction}



\begin{figure}[t]
\centering
\vspace{0.2cm}
\includegraphics[width=\linewidth]{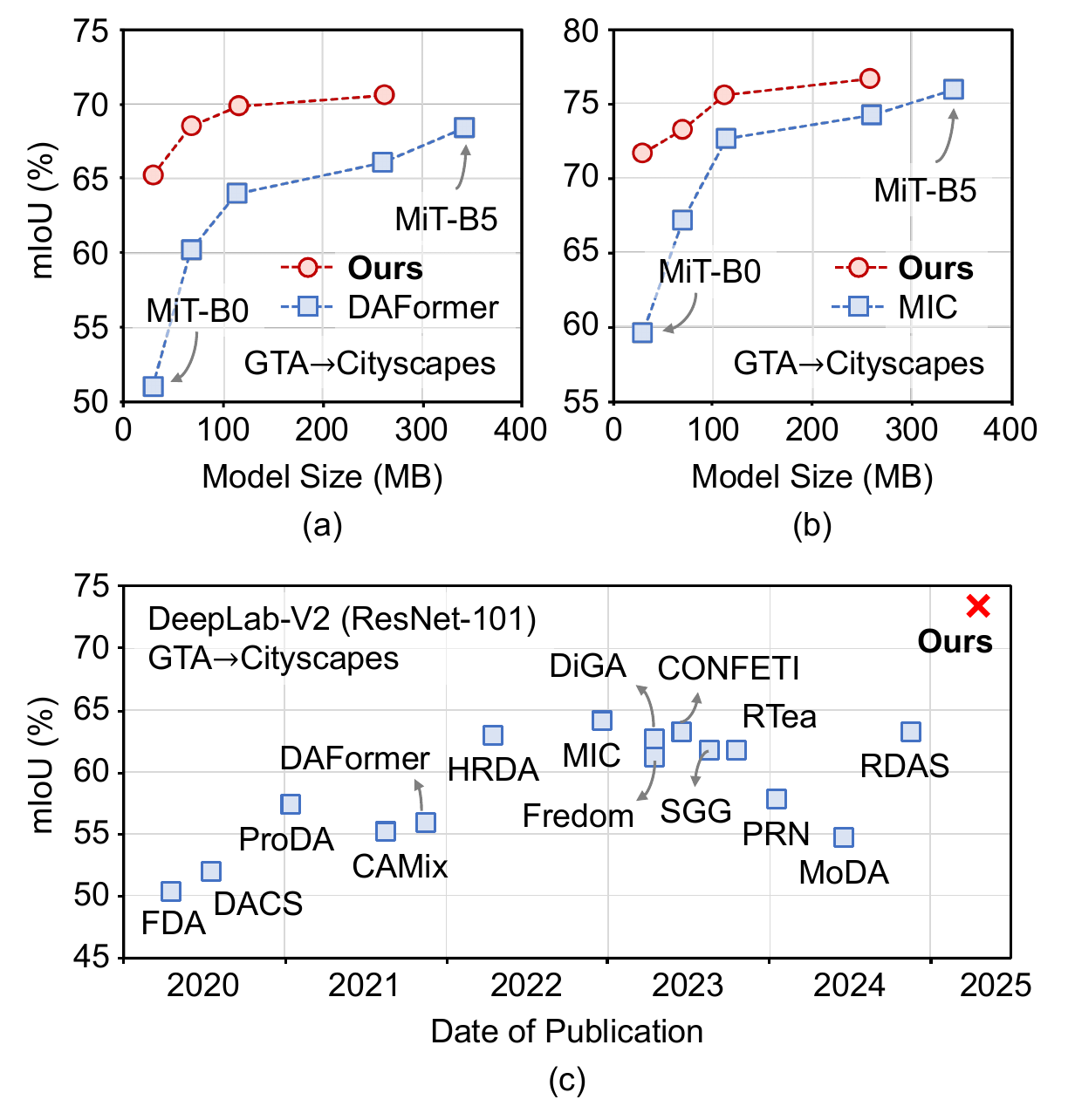}
\vspace{-0.3cm}
\caption{Semantic segmentation accuracy. (a) and (b) compare DUDA with DAFormer \cite{hoyer2022daformer} and MIC \cite{hoyer2023mic}, respectively. Various backbones are employed, such as MiT-B0, B1, B2, B4, and B5, from left to right data points. (c) shows a comparison of DUDA with recent UDA methods using DeepLab-V2 with a ResNet-101 backbone. Performance is measured on synthetic-to-real (GTA$\rightarrow$Cityscapes) adaptation. 
}
\vspace{-0.3cm}
\label{figure:memory_miou}
\end{figure}

Domain shift is a common challenge in multimedia applications involving visual understanding, originating from factors such as variations in camera sensors, changing environmental conditions (e.g., weather, lighting), and diverse visual styles across platforms \cite{sakaridis2021acdc, ren2024cross, wen2024cdea, loiseau2025reliability}. To mitigate performance degradation when systems encounter data distributions different from their training sets, domain adaptation approaches have gained significant attention. Unsupervised Domain Adaptation has emerged as a prominent research focus in recent years \cite{kalluri2024uda}, particularly within the domain of semantic segmentation, as it obviates the need for costly labeled target data  (\(i.e.\), pixel-wise annotation) \cite{hoyer2022daformer, hoyer2023mic, hoyer2022hrda}. State-of-the-art UDA semantic segmentation techniques are mostly based on self-training, where both the student and teacher networks share identical architecture and undergo interactive optimization \cite{tarvainen2017mean, hoyer2022daformer, hoyer2022hrda, hoyer2023mic}. However, existing techniques predominantly focus on improving accuracy, with limited attention to addressing efficiency concerns. Achieving robust performance with minimal computational overhead is critical for deploying these systems in practical, resource-constrained multimedia environments, such as autonomous driving and augmented reality.

The straightforward and common approach to developing efficient UDA models involves applying the SOTA methods to small (\textit{i.e.}, lightweight) networks~\cite{hoyer2023domain, hoyer2022daformer}. However, a significant accuracy degradation is observed, possibly resulting from having the teacher network small as well. For instance, in Figures \ref{figure:memory_miou}a and \ref{figure:memory_miou}b, synthetic-to-real adaptation experiments illustrate the performance degradation in lightweight models (\textit{e.g.}, MiT-B0) compared to heavyweight models (\textit{e.g.}, MiT-B5) \cite{xie2021segformer, hoyer2023mic, hoyer2022daformer}. Meanwhile, few recent studies have explored the UDA methods combined with network compression techniques, such as pruning and neural architecture search, in image classification tasks \cite{yu2019accelerating,feng2020admp,mengslimmable}. However, these methods alter the architecture of the student network, making EMA-based self-training inapplicable and thereby exhibiting a noticeable performance gap compared to SOTA models \cite{hoyer2023mic}. The inherent inflexibility in EMA-based self-training poses challenges in directly employing lightweight networks or applying network compression techniques. The exploration of methods to alleviate these constraints and achieve a balance between accuracy and efficiency in UDA semantic segmentation tasks has been rare.

In this paper, we introduce \textbf{D}istilled \textbf{U}nsupervised \textbf{D}omain \textbf{A}dap-tation (DUDA), a novel self-training method coupled with knowledge distillation (KD). DUDA specifically aims to improve the accuracy of lightweight models trained by the EMA-based self-training for UDA. Our key strategy is to incorporate a large (\textit{i.e.}, heavyweight) auxiliary student network between large teacher and small student networks to address their architectural mismatch. The three networks are jointly trained in a single framework by KD between the large and small networks and the EMA update between the large networks (Figure \ref{figure:duda_configuratoin}). It allows the training of lightweight student networks with highly reliable and consistent pseudo-labels, reaping the advantages of EMA-based self-training with large networks. Additionally, we define the inconsistency between the large teacher and small student networks to identify under-performing classes in an unsupervised manner. Further performance enhancement is achieved by applying non-uniform weighting to the loss functions associated with these identified classes. 

Our joint training approach \textit{distinctly differs from vanilla KD} applied independently post-UDA, which assigns uniform importance to every class and is prone to significant information loss in sequential distillation steps, leading to suboptimal performance. DUDA employs a novel and strategic fusion of UDA and KD, incorporating pre-adaptation (gradual distillation from large to small networks), inconsistency-based loss (prioritizing poorly-adapted classes), and multiple teachers (for enhanced learning). The key contributions of this work are as follows.

\begin{itemize}
\item{We present \textbf{D}istilled \textbf{U}nsupervised \textbf{D}omain \textbf{A}daptation (DUDA) for efficient and domain-adaptive semantic segmentation models. DUDA shows comparable accuracy (using lightweight models) to SOTA methods (using heavyweight models) in four UDA benchmarks.}

\item{DUDA explores UDA for lightweight semantic segmentation models, a largely unexplored area with significant practical implications.}

\item{DUDA can be seamlessly combined with other self-training UDA methods and demonstrates the noticeable accuracy improvement (10\%$\sim$) in lightweight Transformer-based models (Figures \ref{figure:memory_miou}a, \ref{figure:memory_miou}b, Table ~\ref{table:syn2real})}.

\item{DUDA excels in heterogeneous self-training between Transformer and CNN-based models, achieving SOTA accuracy in DeepLab-V2 by a large margin (Figure~\ref{figure:memory_miou}c, Table ~\ref{table:dlv2}}) and even surpasses supervised learning baselines (73.5 mIoU in DUDA$_\text{MIC}$ GTA\(\rightarrow\)Cityscapes vs. 71.4 in fully-supervised DeepLab-V2 \cite{chen2017deeplab} in Cityscapes).

\end{itemize}

%% file: sec/2_related_works.tex
\section{Related Works}
\label{sec:related_works}

\noindent\textbf{Unsupervised Domain Adaptation.} 
UDA Semantic segmentation methods fall into three categories: input, feature, and output-based \cite{schwonberg2023survey}. In the input space, image-to-image translation, particularly style transfer, is common, using Generative Adversarial Networks and diffusion-based methods \cite{murez2018image, dong2021and, peng2023diffusion}. Feature space methods align distributions between source and target domains, employing statistical or heuristic techniques like Maximum Classification Discrepancy and adversarial learning \cite{saito2018maximum, li2021bi, tsai2019domain, saha2021learning}. Output space strategies involve self-training with pseudo-labels from a teacher network's output to train a student network \cite{zhao2023learning, xie2023sepico}. Recently, EMA-based self-training~\cite{tarvainen2017mean} has been the most popular approach for UDA semantic segmentation and demonstrated SOTA accuracy across benchmarks \cite{hoyer2022hrda, zhao2023learning, yang2025micdrop, lim2024cross}. Moreover, several techniques, such as masked image consistency \cite{hoyer2023mic, wen2024cdea} for enhanced context learning and foundation models (\textit{e.g.}, vision-language \cite{ren2024cross, lim2024cross}, segment anything \cite{yan2023sam4udass}) for pseudo-label refinement, have further improved the self-training approach. Our DUDA is specifically designed to integrate seamlessly with these UDA methods.

\begin{figure*}[t]
\centering
\includegraphics[width=\textwidth]{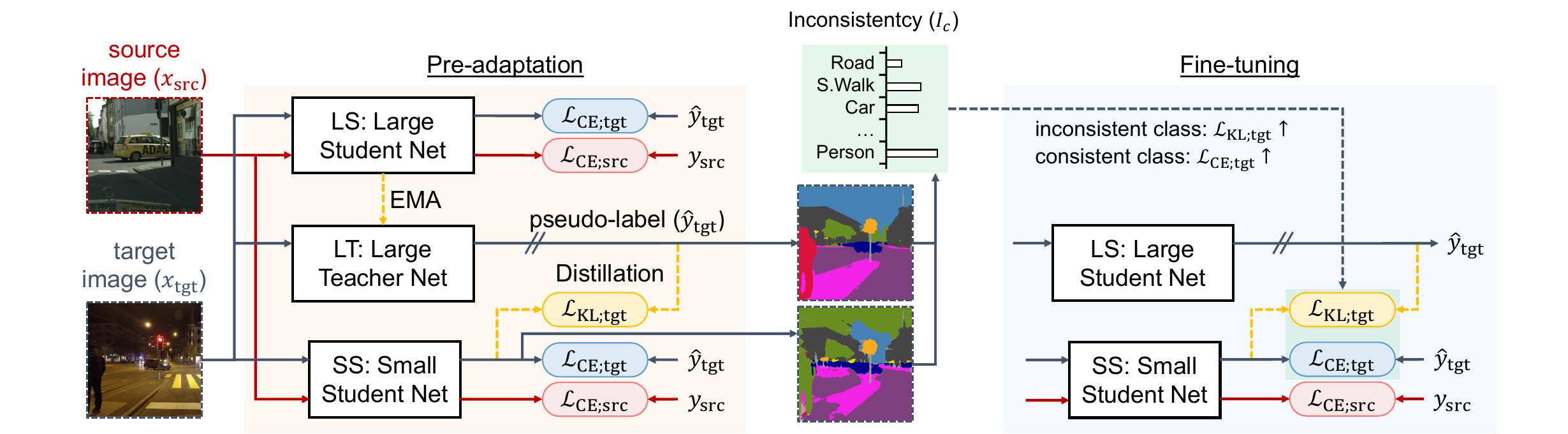}
\caption{A brief illustration of the proposed Distilled Unsupervised Domain Adaptation (DUDA) approach.}
\label{figure:duda_configuratoin}
\end{figure*}

Imbalanced distribution in source and target classes often degrades pseudo-label quality in self-training UDA methods \cite{lee2021unsupervised}. Accordingly, techniques like re-sampling \cite{gao2021dsp, hoyer2022daformer}, uncertainty estimation \cite{wang2021uncertainty,zheng2021rectifying, zou2019confidence}, re-weighting \cite{chen2019domain, zou2018unsupervised, wang2023cal}, and clustering \cite{zhang2019category, li2022class} have been proposed to mitigate this. Our inconsistency-based loss addresses the issue distinctly using predictions from pre-adapted teacher and student models to effectively fine-tune the student (Figure ~\ref{figure:duda_configuratoin}). 
Importantly, our loss complements these existing techniques.

\vspace{0.15cm}
\noindent\textbf{UDA for Lightweight Networks.} The efficiency of models can be enhanced by applying various compression techniques after domain adaptation (\textit{i.e.}, post-training compression), including quantization \cite{kuzmin2022fp8, oh2022non, yvinec2023spiq}, pruning \cite{chen2022mtp, bai2022dynamically, tang2023dynamic}, and knowledge distillation \cite{he2019knowledge, shu2021channel, zhu2023good, qiu2024make, beyer2022knowledge, son2021densely, jin2022semi}. However, our focus lies on techniques that concurrently address accuracy and efficiency during the domain adaptation process. Very few studies have investigated network compression and UDA together. Yu \textit{et al.} developed TCP \cite{yu2019accelerating}, a structured pruning framework for CNN, using Taylor-based importance estimation and Maximum Mean Discrepancy based feature alignment. Feng \textit{et al.} \cite{feng2020admp} also proposed a structured pruning method, coupled with KD-based self-training. Neural architecture search (NAS)-based approach \cite{mengslimmable} exhibited an attractive compression rate compared to the pruning methods. However, they are tailored for image classification tasks and alter network architectures, posing the challenge in potential incompatibility (\textit{i.e.}, misalignment between teacher and student networks) with EMA-based self-training. Knowledge distillation for UDA in semantic segmentation was explored by Kothandaraman \textit{et al.} \cite{kothandaraman2021domain}, utilizing an adversarial loss to reduce cross-domain discrepancy but exhibiting relatively poor performance (\textit{e.g.}, 33.8 mIoU on GTA$\to$Cityscapes). In summary, unsupervised domain adaptation for efficient semantic segmentation, particularly those compatible with EMA-based self-training, has been rarely investigated.

%% file: sec/3_proposed_methods.tex
\section{Proposed Methods}
\label{sec:proposed_methods}

\subsection{Background} 

\vspace{0.1cm}
\noindent\textbf{Unsupervised Domain Adaptation.} UDA in the context of semantic segmentation assumes a labeled source domain, characterized by pixel-wise annotations, and an unlabeled target domain. The primary objective is to enhance the segmentation performance of neural networks without leveraging target labels. In essence, our goal is to adapt a neural network \(f_{\theta}: \mathcal{R}^{H \times W \times 3} \rightarrow \mathcal{R}^{H \times W \times C}\) to the target domain utilizing source images (\(x_{\text{src}})\), source labels (\(y_{\text{src}}\)), and target images (\(x_\text{tgt}\)). \(H \times W\) represents the image dimensions, and \(C\) denotes the number of classes.

\vspace{0.1cm}
\noindent\textbf{Self-training.} Self-training stands out as a commonly employed technique in UDA to train the function \(f_{\theta}\). In particular, EMA-based self-training employs the same function with the two different sets of parameters \(f_{\theta;\text{tea}}\) and \(f_{\theta;\text{stu}}\), so that the EMA update can be defined by:
\begin{equation}
\small
\theta_{\text{tea}}(t+1) \leftarrow \alpha \theta_{\text{tea}}(t) + (1-\alpha) \theta_{\text{stu}}(t)
\label{equation:ema_update1}
\end{equation}
\noindent where \(\alpha\) is a smoothing factor \(\in (0,1]\). In addition, it optimizes the pixel-wise classification loss (\textit{i.e.},  cross-entropy loss) on labeled source images (\(\mathcal{L}_{\text{CE;src}}\)) and pseudo-labeled target images (\(\mathcal{L}_{\text{CE;tgt}}\)) as follows:
\begin{equation}
\small
\mathcal{L}_{\text{CE;src}}=-\sum_{i=1}^{H \times W}\sum_{c=1}^{C}{y_\text{src}^{(i, c)} \cdot \log{f_{\theta;\text{stu}}(x_{\text{src}})^{(i, c)}}}
\label{equation:ce_source}
\end{equation}
\begin{equation}
\small
\mathcal{L}_{\text{CE;tgt}}=-\sum_{i=1}^{H \times W}\sum_{c=1}^{C}{\hat{y}_\text{tgt}^{(i, c)} \cdot \log{f_{\theta;\text{stu}}(x_{\text{tgt}})^{(i, c)}}}
\label{equation:ce_target}
\end{equation}
\noindent where \(\hat{y}_\text{tgt}^{(i, c)}\) is pseudo-labels generated by the teacher \(f_{\theta;\text{tea}}\).

\subsection{Motivation of DUDA}
Our objective is to enhance the efficiency of the student network (\(f_{\theta;\text{stu}}\)) while ensuring the provision of high-quality pseudo-labels (\(\hat{y}_\text{tgt}\)). It is crucial to emphasize that the quality of pseudo-labels significantly influences UDA performance \cite{choi2019pseudo}. As shown in SegFormer \cite{xie2021segformer}, even small models can indeed reach good accuracy with supervision (\textit{e.g.}, mIoU in Cityscapes is MiT-B5: 82.4\%$\rightarrow$MiT-B0: 76.2\%). This suggests that poor performance of small models on the target domain is likely due to limitations in their pseudo-labels, not an inherent weakness (\textit{e.g.}, limited capacity). A smaller student model will reduce inference costs, however, applying the EMA update in Equation (\ref{equation:ema_update1}) necessitates the teacher network (\(f_{\theta;\text{tea}}\)) to be of the same architecture as the student, posing challenges in generating reliable pseudo-labels. 

To overcome this challenge, our proposed method keeps the \(f_{\theta;\text{stu}}\) network small and introduces two large networks \(g_{\phi;\text{tea}}\) and \(g_{\phi;\text{stu}}\), both sharing the same architecture. The DUDA framework thus consists of three distinct networks, a large teacher (\(g_{\phi;\text{tea}}\)), a large student (\(g_{\phi;\text{stu}}\)), and a small student network (\(f_{\theta;\text{stu}}\)). Each network in DUDA serves a specific purpose. For instance, the large teacher network generates pseudo-labels for both the large and small student networks, while the large student network facilitates the EMA update of the big teacher. This design allows any EMA-based self-training methods to be seamlessly applied to large networks, thereby generating robust training signals for the small student network. The remaining key challenge then becomes effectively guiding the small student, which we approach as a KD task between the large teacher and small student networks. By doing so, DUDA effectively integrates unsupervised domain adaptation and knowledge distillation within a unified framework.


\subsection{Training Procedures of DUDA}

Figure \ref{figure:duda_configuratoin} provides an overview of the proposed framework. The input to the networks are the source image (\(x_{\text{src}}\)) with corresponding source label (\(y_{\text{src}}\)) and  the target image (\(x_{\text{tgt}}\)). DUDA has two training stages, pre-adaptation and fine-tuning. The three networks are collaboratively trained in the pre-adaptation, and then, the small student network is further optimized to enhance the performance of low-performing classes. After training, the small network is employed for inference. We abbreviate the large teacher as LT, the large student as LS, and the small student as SS.

\vspace{0.1cm}
\noindent\textbf{Pre-adaptation.} 
For LS and SS networks, we basically optimize the cross-entropy losses. Each network is optimized by \(\mathcal{L}_{CE;\text{src}}\) and \(\mathcal{L}_{CE;\text{tgt}}\) as given in  Equation (\ref{equation:ce_source}) and Equation (\ref{equation:ce_target}). Note, the equations, in the case of LS network, are based on the prediction results, \(g_{\phi;\text{stu}}(x_{\text{src}})\) and \(g_{\phi;\text{stu}}(x_{\text{tgt}})\). The pseudo-labels are generated by the LT network as follows:
\begin{equation}
\small
    \hat{y}_{tgt} = \mathbb{1}(g_{\phi;\text{tea}}(x_{\text{tgt}})) \in \{0,1\}^{H \times W \times C}
\end{equation}
\noindent where \(\mathbb{1}(\cdot)\) is an one-hot encoding function. For LT network, we apply the EMA update (Equation (\ref{equation:ema_update1})) on \(\phi_\text{tea}\) with \(\phi_\text{stu}\).
We additionally incorporate a Kullback-Leibler (KL) divergence loss on the target domain to distill the representation information from LT network to SS network. KL divergence is empirically known to provide richer representation information combined with cross-entropy in KD \cite{kim2021comparing}. This enables SS network to learn not only hard labels represented in the one-hot vector but also the continuous output distribution, which potentially indicates the correlation between the object classes. Our KL divergence loss is given by:
\begin{equation}
\mathcal{L_\text{KL;tgt}}=
g_{\phi;\text{tea}}(x_{\text{tgt}}) \cdot \log{\frac{g_{\phi;\text{tea}}(x_{\text{tgt}})}{f_{\theta;\text{stu}}(x_{\text{tgt}})}}
\label{equation:kl_target}
\end{equation}
\noindent The target distribution in KL divergence can be derived from LS network as well. However, empirically, the performance of LT network shows less fluctuation than that of LS network during the pre-adaptation phase.

Finally, LS network is optimized by cross-entropy loss (\(\mathcal{L_\text{CE;src}}+\mathcal{L_\text{CE;tgt}}\)), and SS network is trained by the cross-entropy and KL divergence loss (\(\mathcal{L_\text{CE;src}}+\mathcal{L_\text{CE;tgt}}+\mathcal{L_\text{KL;tgt}}\)), where KD is simultaneously performed along with UDA.

\vspace{0.1cm}
\noindent\textbf{Inconsistent Prediction.} The disparity in performance between LT (or LS) and SS networks is primarily evident in minority (\textit{e.g.}, less frequent) classes. For instance, both LT (or LS) and SS networks excel in majority (\textit{e.g.}, frequently observed) classes like sky and road, whereas challenges arise with minority classes such as train and motorbike. Although KD generally helps reduce such a performance gap \cite{yuan2020revisiting}, the pre-adaptation procedure is not specifically designed to address the imbalanced performance issue. The difficulty lies in the unsupervised setting, where the identification of minority classes is not directly possible due to the absence of target labels.

We introduce a class-wise inconsistency measure (\(I_\text{c}\)), which quantifies the inconsistency in the prediction results of LT and SS networks, to approximately estimate SS network's class-wise performance on the target domain. The idea is to compare the intersection and union of the prediction results, \(g_{\phi;\text{tea}}(x_\text{tgt})\) and \(f_{\theta;\text{stu}}(x_\text{tgt})\), assuming their ratio can be used as an indicator of poor classes. A high ratio indicates that SS network's prediction is highly different from the LT network's. We define the inconsistency for the object class \(c\) as follows:
\begin{equation}
\small
I_{c} = \frac{\sum_{t}{i_{c}(t)}}{\sum_{t}{n_{c}(t)}}
\label{equation:inconsistency}
\end{equation}
\noindent where \(i_{c}(t)\) is the inconsistency for the class \(c\) at the \(t\)-th image during the pre-adaptation, and \(n_{c}(t)\) is a binary variable to indicate the presence of the class \(c\) in the scene. \(i_{c}(t)\) is mathematically defined by:
\begin{equation}
\small
i_{c}(t) = 1 - \frac{s_{c}(t)}{u_{c}(t)}
\label{equation:temp_inconsistency}
\end{equation}
\begin{equation}
\small
s_{c}(t) = \sum_{i}{\mathbb{1}(g_{\phi;\text{tea}}(x_\text{tgt}))^{(i, c)} \cdot \mathbb{1}(f_{\theta;\text{stu}}(x_\text{tgt}))^{(i, c)}}
\label{equation:intersection}
\end{equation}
\begin{equation}
\small
\begin{split}
u_{c}(t) = & \sum_{i}{\mathbb{1}(g_{\phi;\text{tea}}(x_\text{tgt}))^{(i, c)}} + \sum_{i}{\mathbb{1}(f_{\theta;\text{stu}}(x_\text{tgt}))^{(i, c)}} \\
& - \sum_{i}{\mathbb{1}(g_{\phi;\text{tea}}(x_\text{tgt}))^{(i, c)}} \cdot \mathbb{1}(f_{\theta;\text{tea}}(x_\text{tgt}))^{(i, c)}
\end{split}
\label{equation:union}
\end{equation}
\noindent where \(s_{c}(t)\) and \(u_{c}(t)\) represent the intersection and union of the prediction results between LT and SS networks, respectively, and \(i\) is a pixel index. \(n_{c}(t)\) is heuristically defined. For example, \(n_{c}(t)=0\) if \(u_{c}(t)\) is lower than 0.1\% of entire pixels, indicating that the prediction on the class \(c\) from both the networks is too small. Otherwise, \(n_{c}(t)=1\). The estimation of the inconsistency (\(I_{c}\)) is completed at the end of the pre-adapation stage, and it is used for improving the accuracy on poor classes during the fine-tuning stage.

\vspace{0.1cm}
\noindent\textbf{Fine-tuning.} 
Throughout the pre-adaptation phase, all three networks are trained together. However, the high-quality pseudo-labels are ensured towards the conclusion of the adaptation process. To further refine SS network using the matured pseudo-labels, we introduce the fine-tuning stage. LT and LS networks remain the same after the pre-adaptation (freeze). We observe that the performance of LT and LS networks saturate at a similar level, indicating that either of them can produce high-quality pseudo-labels. Since SS network has already acquired knowledge directly from LT network during the pre-adaptation stage, we leverage pseudo-labels from LS network in the fine-tuning phase.

KD is still applied during the fine-tuning. That is, the loss function for SS network has both the cross-entropy and KL divergence losses. Inconsistency (\(I_{c}\)) is utilized to balance these losses, with the hypothesis that emphasizing soft labels and de-emphasizing hard labels can enhance poor-performing classes.
For instance, the KL loss is given more weight if \(I_{c}\) is high. As a simple way to change their importance, \(I_{c}\) is used as the coefficient of the losses. Finally, the loss function is defined as:
\begin{equation}
\small
\mathcal{L}= \mathcal{L}_{\text{CE;src}} + \sum_{c}{(2-I'_{c}) \cdot \mathcal{L}_{\text{CE;tgt}}^{(c)}} + \sum_{c}{I'_{c} \cdot \mathcal{L}_{\text{KL;tgt}}^{(c)}}
\label{equation:balanced_loss}
\end{equation}
\begin{equation}
\small
I'_{c}=C \cdot \frac{I_{c}}{\sum_{c}{I_{c}}}
\end{equation}
\noindent where \(I'_{c}\) is the normalized inconsistency, and \(C\) is the number of classes (\textit{e.g.}, 19 for Cityscapes \cite{cordts2016cityscapes}). The coefficients are \(2-I'_{c}\) and \(I'_{c}\) to keep the sum of them (\textit{i.e.}, 2) same as the non-weighted losses. 

\subsection{Network Architectures}
DUDA is a model-agnostic framework and applicable to various recent architectures such as DeepLab \cite{chen2017deeplab} (ResNet-based \cite{he2016deep}) and SegFormer MiT (Transformer-based) \cite{xie2021segformer}. Considering the recent SOTA accuracy achieved by the Transformer-based architectures \cite{hoyer2022hrda, hoyer2023mic}, we employ MiT-B5 as a backbone in LT and LS networks and explore lightweight backbones, including MiT-B0, B1, B2, and B4, in SS network. An interesting question is whether DUDA is still effective if SS network is not the Transformer-based architecture. In this light, we define heterogeneous self-training, where LT and LS networks are Transformer-based, and SS network is CNN-based. Then, the former configuration is defined as homogeneous self-training. In heterogeneous self-training, we use a DeepLab-V2 (ResNet-based) \cite{chen2017deeplab} backbone in SS network. While DeepLab-V2 is based on a ResNet-101 \cite{he2016deep} backbone and is not essentially a lightweight design, we aim to understand how much we can improve it through the heterogeneous setting. To further evaluate the heterogeneous setting in lightweight designs, we perform our experiments by replacing the ResNet-101 backbone to ResNet-50 and ResNet-18. For decoder heads, we directly utilize the DAFormer and HRDA \cite{hoyer2022hrda} models, which have shown SOTA accuracy in prior works in UDA segmentation tasks.

%% file: sec/4_experimental_results.tex
\begin{table}
\centering
\renewcommand{\arraystretch}{1.5}
\setlength{\tabcolsep}{2.6pt}
\fontsize{7.7pt}{7.7pt}\selectfont

\caption{Comparison in accuracy and inference costs with recent SOTA UDA semantic segmentation methods. FLOPs and latency (ms) of decoder heads are omitted since the methods are based on the same decoder. FLOPs of the decoder head in HRDA is calculated by 202.3 GFLOPs. 
Note, the computational costs of DUDA remain the same as MIC with MiT-B0, B1, B2, and B4 backbone. All other methods are implemented by MiT-B5. Numbers in parentheses reflect comparisons to the best-reported prior works.
}
\vspace{-0.2cm}
\begin{tabular}{l|ccccc}

\hline
\multirow{2}{*}{Method} & \multicolumn{2}{c}{mIoU (\%) $\uparrow$} & \multicolumn{1}{c}{Memory $\downarrow$} & \multicolumn{1}{c}{GFLOPs $\downarrow$} & \multicolumn{1}{c}{Latency $\downarrow$}\\ 
& GTA$\rightarrow$CS & SYN$\rightarrow$CS & (MB) & Backbone & Backbone \\ 

\hline

HRDA \cite{hoyer2022hrda} & 73.8 & 65.8 & 342.8 & 248.6 & 55.1\\
DiGA \cite{shen2023diga} & 74.3 & 66.2 & 342.8 & 248.6 & 55.1\\
GANDA \cite{liao2023geometry} & 74.5 & 67.0 & 342.8 & 248.6 & 55.1\\
RTea \cite{zhao2023learning} & 74.7 & 66.5 & 342.8 & 248.6 & 55.1\\
BLV \cite{wang2023balancing} & 74.9 & 66.8 & 342.8 & 248.6 & 55.1\\
IR$^{2}$F \cite{gong2023continuous} & 74.4 & 67.7 & 342.8+$\alpha$ & 248.6 & 55.1\\
CDAC \cite{wang2023cdac} & 75.3 & \textbf{68.7} & 342.8 & 248.6 & 55.1 \\
PiPa \cite{chen2023pipa} & 75.6 & 68.2 & 342.8+$\alpha$ & 248.6 & 55.1 \\
MIC \cite{hoyer2023mic} & 75.9 & 67.3 & 342.8 & 248.6 & 55.1 \\

CDEA \cite{wen2024cdea} & \underline{76.3} & \underline{68.4} & 342.8 & 248.6 & 55.1 \\

\hline

\textbf{DUDA}\textbf{$_\text{MIC}$} (B4) & \textbf{76.7} (+0.4) & 68.3 (-0.4) & 260.4 & 191.9 & 46.0 \\ 
\textbf{DUDA}\textbf{$_\text{MIC}$} (B2) & 75.5 (-0.8) & 67.3 (-1.4) & 113.8 & 73.0 & 22.4 \\ 
\textbf{DUDA}\textbf{$_\text{MIC}$} (B1) & 73.3 (-3.0) & 66.8 (-1.9) & \underline{69.6} & \underline{37.7} & \underline{12.0} \\
\textbf{DUDA}\textbf{$_\text{MIC}$} (B0) & 71.7 (-4.6) & 65.3 (-3.4) & \textbf{29.2} & \textbf{11.7} & \textbf{8.9} \\ 

\hline
\end{tabular}
\label{table:overview_cs}
\vspace{-0.2cm}
\end{table}

\section{Experimental Results}
\label{sec:experimental_results}

\vspace{0.1cm}
\noindent\textbf{Dataset.} 
We explore common scenarios of domain adaptation: synthetic-to-real, day-to-nighttime, and clear-to-adverse weather adaptation. For the synthetic data, we use the GTA dataset \cite{richter2016playing}, comprising 24,966 images sized at 1914\(\times\)1052 pixels, and Synthia (SYN) dataset~\cite{ros2016synthia}, consisting of 9.4K images sized at 1280\(\times\)760 pixels. For the real-world data, we use Cityscapes (CS) dataset~\cite{cordts2016cityscapes} with 2,975 training images and 500 validation images at a resolution of 2048\(\times\)1024 pixels (daytime and clear weather), DarkZurich (DZur) dataset~\cite{sakaridis2020map} with 2,416 training, 50 validation, and 151 test images at 1920\(\times\)1080 pixels (nighttime), and ACDC  dataset~\cite{sakaridis2021acdc}, which includes 1,600 training, 406 validation, and 2,000 test images (fog, night, rain, and snow) at a resolution of 1,920\(\times\)1080 pixels. Following prior works, the accuracy in experimental results is evaluated on validation images in GTA$\rightarrow$CS and SYN$\rightarrow$CS adaptation and test images in CS$\rightarrow$DZur and CS$\rightarrow$ACDC adaptation.

\vspace{0.1cm}
\noindent\textbf{Baseline.} Prior UDA approaches have typically applied uniform methodologies across network architectures, often evaluating their effectiveness primarily on MiT-B5 and DeepLab-V2 ResNet-101 backbones \cite{wang2023cdac,chen2023pipa, hoyer2023mic,shen2023diga,li2023contrast}. Only a few studies, such as DAFormer and HRDA, have been applied to relatively small networks, including MiT-B3, B4, and ResNet-50 backbones. Therefore, we consider directly applying the recent SOTA, MIC (\textit{i.e.}, an extension of HRDA) and DAFormer, as our UDA-based baseline. Another intuitive baseline is to train large networks using UDA techniques, and then directly apply knowledge distillation to a small network, accordingly considered KD-based baseline. Since both the baselines are not implemented in prior works, we modify the DAFormer and MIC frameworks to incorporate small backbones for the UDA-based baseline and perform ablation studies in the DUDA framework for the KD-based baseline.

\vspace{0.1cm}
\noindent\textbf{Implementation Details.} Our implementation closely follows SOTA prior works, DAFormer \cite{hoyer2022daformer} and MIC \cite{hoyer2023mic}, including backbone and decoder head structures, as well as learning rate, batch size, optimizer, etc. The backbones are pre-trained on ImageNet-1k. Training strategies utilized in DAFormer \cite{hoyer2022daformer}, such as data augmentation, learning rate warm-up, ImageNet feature distance (FD), and rare class sampling (RCS), are integrated into DUDA, but omitted in Figure \ref{figure:duda_configuratoin} for simplicity. In addition, the MIC training strategy, such as masked loss, is applied within the DUDA framework when coupled with MIC. We allocate 40k iterations for the pre-adaptation phase and 80k iterations for the fine-tuning phase. We perform our experiments with 3 different random seeds.

\vspace{-0.5mm}
\subsection{Comparison with SOTA models}

\vspace{0.5mm}
\noindent\textbf{Lightweight Backbone.} We investigate the accuracy and inference computational costs of DUDA, comparing it with SOTA UDA methods. The recent SOTA models are mostly based on HRDA with a SegFormer backbone. MIC was explored in different UDA benchmarks, showing SOTA mIoU (\textit{e.g.}, 75.9\% in GTA$\rightarrow$CS adaptation). Accordingly, we primarily consider DUDA coupled with MIC (DUDA$_\text{MIC}$), incorporating MiT-B0, B1, B2, or B4 backbone, in comparison. Table \ref{table:overview_cs} provides a comprehensive comparison between DUDA and the recent UDA techniques to the accuracy, memory, FLOPs, and latency in synthetic-to-real adaptation. We assume the compared methods share the same computational costs since they mention their implementation follows HRDA with MiT-B5 backbone. For the FLOPs and latency estimation, a single input image of 512\(\times\)1024 pixels is considered with an overlapping sliding window process \cite{hoyer2022hrda} in a NVIDIA RTX A5000 GPU.

\vspace{0.1cm}
\noindent\textbf{Comparable Accuracy with Less Computation.} It is important to note that, the model size, FLOPs, and latency in Table \ref{table:overview_cs} represent the inference costs of SS network since the large networks are not employed during the inference. Given our smaller architectures are implemented by smaller backbones and utilize the same decoder design with DAFormer and HRDA, computational benefits are mainly obtained from lightweight backbone architectures. For example, MiT-B0 model is approximately \(\times\)12 smaller than MiT-B5 model, and its FLOPs is \(\times\)21 less (backbone) and \(\times\)2.2 less (total) than MiT-B5 with \(3\sim6\%\) accuracy degradation using DUDA. Prior UDA methods (\textit{e.g.}, MIC) show substantially more degradation in accuracy with small backbones (as discussed in the next section). We observe that using the MiT-B2 backbone (with about $\times3$ less model size and backbone FLOPs than MiT-B5), DUDA models achieve comparable performance to prior SOTA. Results in other datasets, including class-wise IoU are in Suppl. Training time and memory requirements are also available in Suppl. Our proposed method excels in inference, delivering similar accuracy with a significantly lower memory model.

\vspace{0.1cm}
\noindent\textbf{Baselines with extended Self Fine-tuning vs. DUDA.} We find that simply extending training iterations for SOTA baseline MIC (\textit{e.g.}, increasing from 40k to match DUDA’s total iterations) does not consistently lead to improved mIoU.
For example, MIC B5 performance is degraded with 120k iterations (\textit{i.e.}, self fine-tuning) to 63.8\% mIoU (vs. 67.3\% with MIC at 40k) in SYN$\rightarrow$CS, which is even worse than 65.3\% achieved with DUDA MIC-B0.

\subsection{Comparison with DeepLab-V2 Models}

\vspace{0.1cm}
\noindent{\textbf{Heterogeneous Self-training.}} DUDA's distinctive feature is the flexibility for teacher and student networks to have different architectures, allowing diverse combinations like Transformer-based teacher and CNN-based student. While most CNN-based UDA models often rely on DeepLab-V2 with a ResNet-101 backbone for superior accuracy, a significant performance gap persists compared to Transformer-based models with a MiT-B5 backbone \cite{hoyer2023mic}. This suggests that using MiT-B5 as a teacher could potentially enhance the accuracy of a DeepLab-V2 student. 

\begin{table}[t]
\centering
\renewcommand{\arraystretch}{1.5}
\setlength{\tabcolsep}{4.1pt}
\fontsize{7.7pt}{7.7pt}\selectfont

\caption{Comparison of DUDA with prior methods in ResNet-based networks. R101, R50, and R18 stand for ResNet-101, ResNet-50, and ResNet-18 backbone in DeepLab-V2 architecture. Note that all other methods are based on ResNet-101 backbone. *The results of DAFormer, except for GTA$\rightarrow$CS, are reproduced by ourselves. Inference computational costs are provided in Suppl.}
\vspace{-2mm}
\begin{tabular}{l|cccc}

\hline
\multirow{2}{*}{Method} & \multicolumn{4}{c}{mIoU (\%) $\uparrow$}  \\ 
& GTA$\rightarrow$CS & SYN$\rightarrow$CS & CS$\rightarrow$DZur & CS$\rightarrow$ACDC \\ 

\hline

ADVENT \cite{vu2019advent} & 45.5 & 41.2 & 29.7 & 32.7 \\
DACS \cite{tranheden2021dacs} & 52.1 & 48.3 & - & -\\
ProDA \cite{zhang2021prototypical} & 57.5 & 55.5 & - & - \\
MGCDA \cite{sakaridis2020map} & - & - & 42.5 & 48.7 \\
DANNet \cite{wu2021dannet} & - & - & 42.5 & 50.0 \\
DAFormer \cite{hoyer2022daformer} & 56.0 & 53.3* & 42.2* & 52.3* \\
GANDA \cite{liao2023geometry} & - & 56.3 & - & - \\
Fredom \cite{truong2023fredom} & 61.3 & 59.1 & - & - \\
RTea \cite{zhao2023learning} & 61.9 & 58.9 & - & - \\
DiGA \cite{shen2023diga} & 62.7 & 60.2 & - & - \\
SGG \cite{peng2023diffusion} & 61.9 & 61.0 & - & - \\
Jeong \textit{et al}. \cite{jeong2024revisiting} & 63.3 & 58.6 & - & - \\
CONFETI \cite{li2023contrast} & 63.3 & 58.7 & - & - \\
HRDA \cite{hoyer2022hrda} & 63.0 & 61.2 & 44.9 & 57.6 \\
MIC \cite{hoyer2023mic} & 64.2 & 62.8 & 49.4 & 60.4 \\

\hline
\textbf{DUDA}\textbf{$_\text{MIC}$} (R101) & \textbf{73.5} (+9.3) & \textbf{67.4} (+4.6) & \textbf{55.2} (+5.8) & \textbf{65.9} (+5.5) \\ 
\textbf{DUDA}\textbf{$_\text{MIC}$} (R50) & \underline{72.8} (+8.6) & \underline{67.2} (+4.4) & \underline{54.2} (+4.8) & \underline{63.9} (+3.5) \\ 
\textbf{DUDA}\textbf{$_\text{MIC}$} (R18) & 69.3 (+5.1) & 63.7 (+0.8) & 49.3 (-0.1) & 57.7 (-2.7) \\ 

\hline
\end{tabular}
\label{table:dlv2}
\vspace{-0.2cm}
\end{table}

\vspace{0.1cm}
\noindent{\textbf{SOTA Accuracy in ResNet-based Methods.}} Table \ref{table:dlv2} compares the recent UDA methods applied to ResNet-based models in the four different datasets. DUDA$_\text{MIC}$ surpasses the accuracy of baseline MIC with a DeepLab-V2 backbone (DeepLab-V2) by noticeable differences (\(4.6\sim9.3\%\) mIoU). In prior UDA Semantic Segmentation works, the performance gap between Transformer-based and ResNet-based architectures has been quite large. For example, DeepLab-V2 trained with MIC achieves mIoU scores of 64.2\%, 62.8\%, 49.4\%, and 60.4\% across the four datasets, whereas MiT-B5 trained with MIC demonstrates significantly higher performance with scores of 75.9\%, 67.3\%, 60.2\%, and 70.4\%
However, our method reduces the gap into a few mIoU percentages ($\sim$5\%) and even demonstrates the same accuracy ($+$0.1\%) in SYN$\rightarrow$CS.

ResNet-18 trained with DUDA achieves a substantial improvement over the previous state-of-the-art MIC-based ResNet-101 model by a huge margin in GTA$\rightarrow$CS, such as 69.3 mIoU for DUDA\(_\text{MIC}\) ResNet-18 and 64.2 mIoU for MIC ResNet-101. These results again highlight the significant effectiveness of our approach.

\vspace{0.1cm}
\noindent{\textbf{Accuracy against supervised DeepLab-V2.}} Interestingly, our ResNet-101 in GTA$\rightarrow$CS achieves even higher accuracy (73.5 mIoU) than the supervised accuracy reported in the original DeepLab-V2 paper (71.4 mIoU in CS) \cite{chen2017deeplab}. Note, that our training procedures are built on several recent techniques, such as masked loss, and the decoder head is different. However, these results indicate that the large performance gap between supervised learning and UDA, consistently observed in prior UDA works with DeepLab-V2, can be meaningfully reduced.

\vspace{0.1cm}
\noindent\textbf{Performance improvement with CNNs compared to Transformers.} DUDA's performance over SOTA methods is apparently much higher in DeepLab-V2 models. However, compared with the baseline performance (\textit{i.e.}, without DUDA), DUDA$_\text{MIC}$ boosts MiT-B0 by +12.2\% mIoU and ResNet-101 by +9.3\% mIoU, indicating DUDA is effective in both Transformer-based and ResNet-based models. To further clarify, the higher gains over SOTA in CNNs (Table \ref{table:dlv2} vs. Table \ref{table:overview_cs}) stems from a larger performance gap in the teacher (MiT-B5) and ResNet models, which DUDA effectively reduces.

\subsection{Comparison in Lightweight Networks}
Recent SOTA UDA semantic segmentation methods have not been explored in the same small architectures. In this section, we experimentally show DUDA leads to huge performance improvement across four UDA benchmarks (Table \ref{table:syn2real}), particularly in small backbone architectures (with the same inference costs). In addition to DUDA$_\text{MIC}$, we conduct experiments combining DUDA with DAFormer, denoted as DUDA$_\text{DAF}$, to highlight its applicability and effectiveness with other UDA techniques.

\begin{table}[t]
\centering
\renewcommand{\arraystretch}{1.4}
\setlength{\tabcolsep}{5pt}
\fontsize{7.7pt}{7.7pt}\selectfont

\caption {Comparison with DAFormer and MIC in lightweight networks. mIoU for SYN$\rightarrow$CS is averaged over 16 classes. Class-wise IoU is available in Suppl.}
\vspace{-2mm}
\begin{tabular}{l|c|cccc}

\hline
\multirow{2}{*}{Method} & & \multicolumn{4}{c}{mIoU (\%) $\uparrow$} \\ 
& & GTA$\rightarrow$CS & SYN$\rightarrow$CS & CS$\rightarrow$DZur & CS$\rightarrow$ACDC \\ 

\hline

DAFormer & \multirow{4}{*}{\rotatebox[origin=c]{90}{MiT-B0}} & 51.0 & 46.1 & 33.5 & 44.9 \\
\textbf{DUDA}\textbf{$_\text{DAF}$} & & \textbf{65.2} & \textbf{58.3} & \textbf{50.1} & \textbf{53.1} \\
MIC & & 59.5 & 53.0 & 37.5 & 51.7 \\
\textbf{DUDA}\textbf{$_\text{MIC}$} & & \textbf{71.7} & \textbf{65.3} & \textbf{54.9} & \textbf{64.4} \\

\hline
DAFormer & \multirow{4}{*}{\rotatebox[origin=c]{90}{MiT-B1}} & 60.2 & 55.4 & 36.7 & 49.4 \\
\textbf{DUDA}\textbf{$_\text{DAF}$} & & \textbf{68.5} & \textbf{60.5} & \textbf{51.8} & \textbf{54.7} \\
MIC & & 67.1 & 65.0 & 44.2 & 56.9 \\
\textbf{DUDA}\textbf{$_\text{MIC}$} & & \textbf{73.3} & \textbf{66.8} & \textbf{56.7} & \textbf{66.5} \\

\hline
DAFormer & \multirow{4}{*}{\rotatebox[origin=c]{90}{MiT-B2}} & 63.9 & 59.4 & 44.7 & 52.2 \\
\textbf{DUDA}\textbf{$_\text{DAF}$} & & \textbf{69.8} & \textbf{61.6} & \textbf{54.4} & \textbf{56.8} \\
MIC & & 72.7 & 66.4 & 56.5 & 60.9 \\
\textbf{DUDA}\textbf{$_\text{MIC}$} & & \textbf{75.5} & \textbf{67.3} & \textbf{59.6} & \textbf{68.6} \\
\hline

DAFormer & \multirow{4}{*}{\rotatebox[origin=c]{90}{MiT-B4}} & 66.1 & 59.9 & 46.9 & 55.5 \\
\textbf{DUDA}\textbf{$_\text{DAF}$} & & \textbf{70.5} & \textbf{62.0} & \textbf{54.4} & \textbf{57.7} \\
MIC & & 74.3 & 66.0 & 59.9 & 64.0 \\
\textbf{DUDA}\textbf{$_\text{MIC}$} & & \textbf{76.7} & \textbf{68.3} & \textbf{60.3} & \textbf{70.2} \\

\hline
\end{tabular}
\label{table:syn2real}
\vspace{-2mm}
\end{table}

\vspace{0.1cm}
\noindent{\textbf{Larger Improvement in Smaller Models.}} Table \ref{table:syn2real} presents a comparison of DUDA$_\text{MIC}$ against MIC and DUDA$_\text{DAF}$ against DAFomer across MiT-B0, B1, B2, and B4 backbones. In the GTA\(\rightarrow\)CS dataset, DUDA$_\text{DAF}$ and DUDA$_\text{MIC}$ achieve remarkable mIoU improvements of 14.2\% and 12.2\%, respectively, for the MiT-B0 backbone, and 8.3\% and 6.2\%, respectively, for the MiT-B1 backbone, relative to their counterparts without DUDA. 
The results imply that DUDA is particularly effective for small models, as shown in Figure \ref{figure:memory_miou} as well. This trend is consistently observed across other datasets as well. Overall, due to learning from higher-quality labels and an inconsistency-based balancing strategy, DUDA performs significantly better in most of the classes across benchmarks, with more substantial improvement generally observed in the minority classes.

\begin{figure*}
\centering
\includegraphics[width=0.9\textwidth]{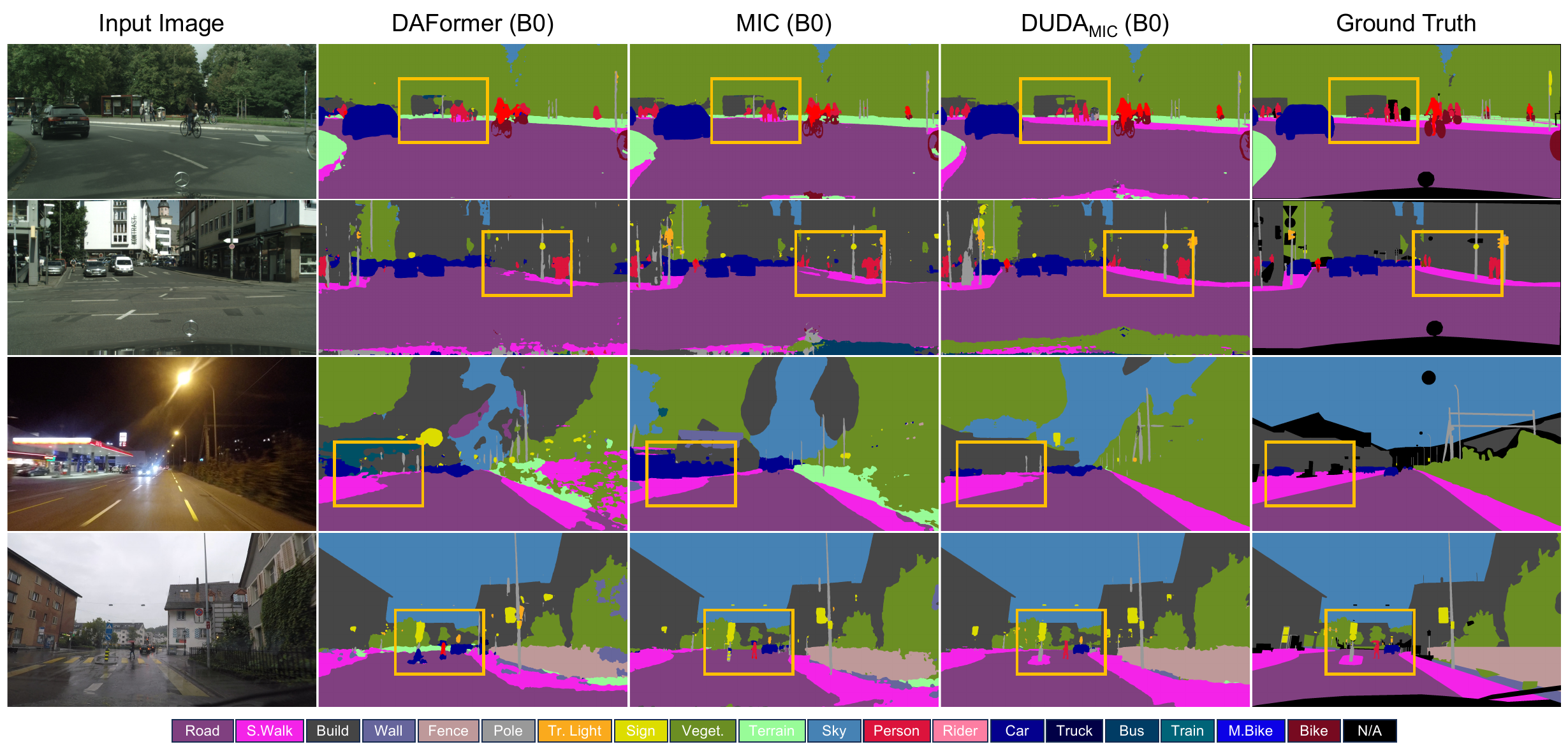}
\caption{Qualitative results in benchmarks: GTA$\rightarrow$CS (1st row); SYN$\rightarrow$CS (2nd row); CS$\rightarrow$DZur (3rd row); CS$\rightarrow$ACDC (4th row).}
\label{figure:pred_cityscapes}
\end{figure*}

\subsection{Ablation Study}
Table \ref{table:ablation_daformer} showcases the accuracy of MiT-B0 and MiT-B1 models trained by DUDA$_\text{DAF}$ with the absence of the pre-adaptation or fine-tuning with cross-entropy, KL divergence loss, or inconsistency-based loss balancing in the GTA\(\rightarrow\)CS dataset. Other rows incorporate three networks and utilize knowledge distillation from large networks to small one. The model lacking the pre-adaptation is exclusively trained during the fine-tuning stage using the pre-adapted large networks. This investigation aims to discern the performance disparity between pre-training the small student network with gradually evolving pseudo-labels and directly training the initialized small student network with matured pseudo-labels. More ablation studies, such as LS/LT in the pre-adpataiton or fine-tuning, the same experiments in MiT-B2 and MiT-B4 models, and MiT-B5 as both teacher (LS and LT) and student (SS) networks, are in Suppl.

\vspace{0.1cm}
\noindent{\textbf{Comparison with Baseline and vanilla KD.}} row-1 and row-3 (No Distillation and CE \checkmark) in Table \ref{table:ablation_daformer} are our baseline methods; row-1 as an UDA baseline and row-3 as a vanilla KD baseline. Row-1 (No Distillation) indicates DAFormer-based UDA applied to small teacher and student networks without knowledge distillation. Our primary observation is that the accuracy enhancement in the DUDA framework mainly stems from the cross-entropy loss. In other words, generating pseudo-labels from the large network emerges as the pivotal component in DUDA. The introduction of other components incrementally boosts both mIoU and mAcc, surpassing the vanilla KD baseline. 

\vspace{0.1cm}
\noindent\textbf{Benefits of Pre-adaptation and Fine-tuning.} Pre-adaptation is crucial because standard knowledge distillation (KD) fine-tuning struggles with large capacity differences between teacher and student models. A large teacher’s complex representation can overwhelm the student initially. Pre-adaptation addresses this by gradually distilling knowledge throughout UDA, allowing the student to adapt progressively. Then, fine-tuning boosts performance by using matured pseudo-labels from well-trained teacher and inconsistency weighting to prioritize under-performing classes. To validate this, we train an initialized student using a pre-trained teacher in the fine-tuning setup (\textit{i.e.}, fine-tune alone). Our experiments demonstrate incorporating the pre-adaptation stage enhances the KD process. For example, DAFormer base MiT-B0 in GTA$\rightarrow$CS, fine-tune alone (120k) vs pre-adaptation (40k)+fine-tune (80k) is 64.6\% vs. 65.2\%. Similarly, in MiT-B1, 68.0\% vs. 68.5\%.

\begin{table}[t]
\centering
\renewcommand{\arraystretch}{1.5}
\setlength{\tabcolsep}{3.5pt}
\fontsize{7.7pt}{7.7pt}\selectfont

\caption{Ablation Studies of DUDA on GTA$\to$CS with DAFormer as the base. The basic model is configured as DAFormer without DUDA, and subsequently, we incrementally introduce pre-adaptation and fine-tuning with cross-entropy, KL divergence, and inconsistency-based balanced losses in the DUDA setup.}
\vspace{-2mm}
\begin{tabular}{l|c|ccc|cc}

\hline
\multirow{2}{*}{Method} & \multicolumn{1}{c|}{Pre-} & \multicolumn{3}{c|}{Fine-tuning} & \multirow{2}{*}{mIoU (\%)} & \multirow{2}{*}{mAcc (\%)} \\
& adaptation & CE & KL & Inconsistency & & \\
\hline

MiT-B0 & \multicolumn{4}{c|}{No Distillation} & 51.0 & 62.5 \\
MiT-B0 & \checkmark & & & & 59.6 & 69.6 \\
MiT-B0 & & \checkmark & & & 62.3 & 71.7 \\
MiT-B0 & & \checkmark & \checkmark & & 63.7 & 72.9 \\
MiT-B0 & \checkmark & \checkmark & \checkmark & & 64.4 & 73.5 \\
MiT-B0 & \checkmark & \checkmark & \checkmark & \checkmark & 65.2 & 75.2 \\
\hline

MiT-B1 & \multicolumn{4}{c|}{No Distillation} & 60.2 & 70.8 \\
MiT-B1 & \checkmark & & & & 64.7 & 75.6 \\
MiT-B1 & & \checkmark & & & 66.8 & 76.2 \\
MiT-B1 & & \checkmark & \checkmark & & 67.9 & 77.2 \\
MiT-B1 & \checkmark & \checkmark & \checkmark & & 68.4 & 77.6 \\
MiT-B1 & \checkmark & \checkmark & \checkmark & \checkmark & 68.5 & 78.9 \\
\hline

\end{tabular}
\label{table:ablation_daformer}
\vspace{-2mm}
\end{table}

\subsection{Visualization of Prediction Results} 
Figure \ref{figure:pred_cityscapes} display predictions from MiT-B0 models trained by DUDA$_\text{MIC}$ compared to DAFormer, and MIC. We observe DUDA predictions are significantly better. We highlight some interesting regions with orange boxes. From synthetic-to-real adaptation results in row-1 and row-2, we observe that DUDA accurately predicts sitting persons, sidewalks, and persons, contrasting with unstable predictions from DAFormer and MIC. In the nighttime image prediction in row-3 after CS$\rightarrow$DZur adaptation, DAFormer fails to segment the sidewalk appropriately, and MIC incorrectly predicts many cars near the gas station. In the rainy image in row-4 after CS$\rightarrow$ACDC adaptation, both DAFormer and MIC fail to classify the small island of the sidewalk, whereas our DUDA model succeeds. While DUDA demonstrates substantial improvements over other methods, there remains room for further enhancement in certain cases (\textit{e.g.}, row-3). We believe incorporating additional guidance (\textit{e.g.}, depth cues \cite{yang2025micdrop}, vision-language model assistance \cite{lim2024cross}) could further enhance DUDA’s performance, which we leave as future work.

\begin{figure}
\centering
\includegraphics[width=\linewidth]{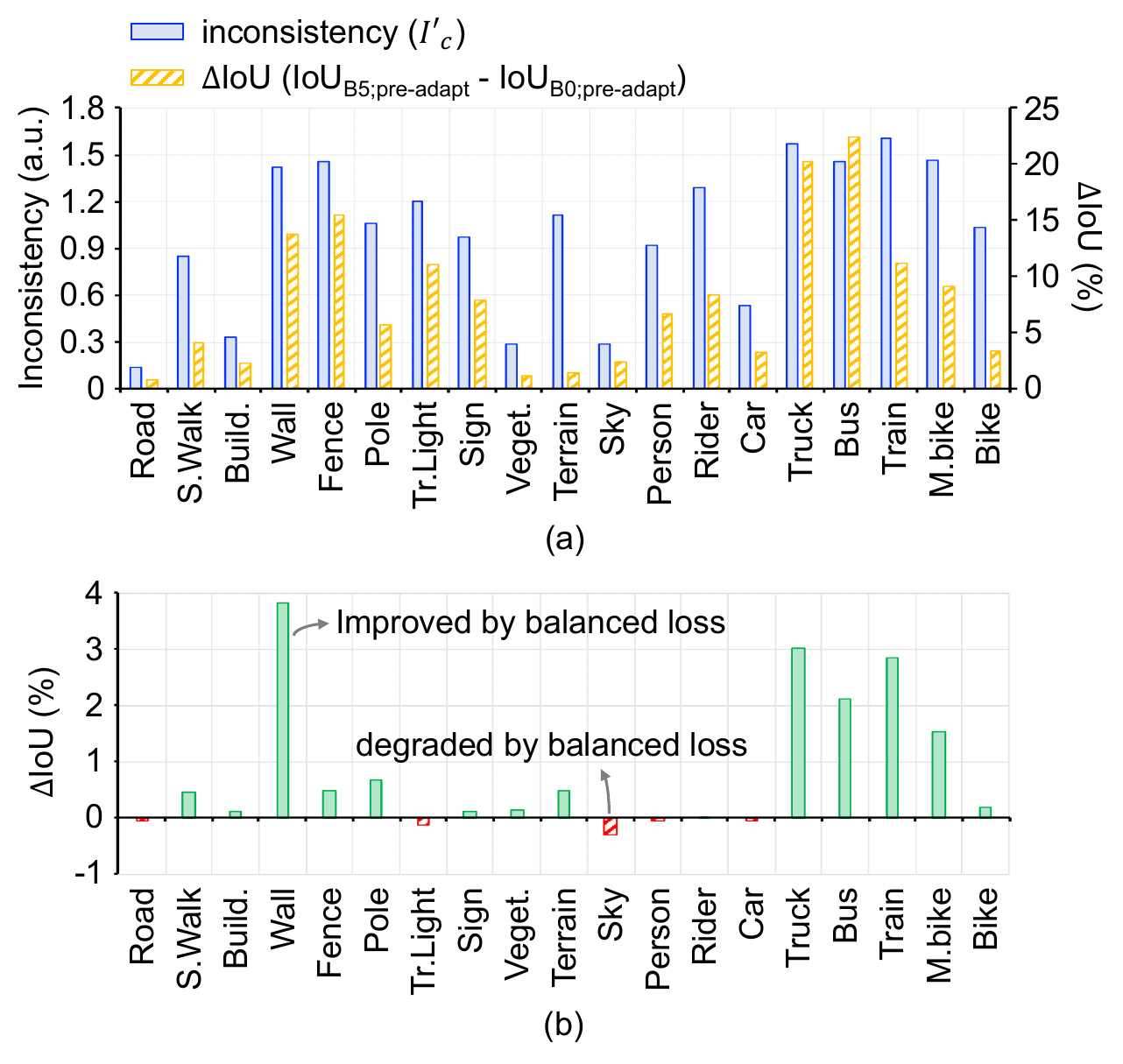}
\caption{Class-wise inconsistency and accuracy changes resulted from inconsistency-based loss balancing on GTA$\rightarrow$CS. (a) compares the class-wise true IoU disparity and the normalized inconsistency ($I'_{c}$) between MiT-B5 and MiT-B0 after the pre-adaptation by DUDA$_\text{DAF}$. (b) shows the accuracy difference between the MiT-B0 models after fine-tuning with and without the loss balancing. \textcolor{red}{}}
\label{figure:inconsistency}
\end{figure}

\subsection{Analysis on Inconsistency Prediction}

This section empirically investigates two key assumptions in the inconsistency ($I_{c}$)-based balancing of cross-entropy and KL divergence losses. First, the proposed inconsistency can approximately identify underperforming classes. Equation (\ref{equation:temp_inconsistency}) illustrates this estimation as the ratio of the intersection to union. In this context, $I_{c}$ can be regarded as the normalized IoU measured on pseudo-labels for the class \(c\) during the pre-adaptation. Figure \ref{figure:inconsistency}a displays the class-wise inconsistency values and the true IoU difference between LT (MiT-B5) and SS (MiT-B0) networks after the pre-adaptation. 
It is important to note that since the inconsistency is computed on pseudo-labels, its reliability hinges on the quality of these labels. For instance, significant differences between pseudo-labels and true labels may yield similarly low accuracy values for both the teacher and student, even with the high inconsistency. Second, increasing the importance of the KL loss contributes to improving poor classes. Figure \ref{figure:inconsistency}b outlines the class-wise accuracy for two MiT-B0 models fine-tuned with and without the inconsistency-based loss balancing, leveraging the inconsistency distribution from Figure \ref{figure:inconsistency}a for training. The balanced loss generally enhances the accuracy of classes with high inconsistency.

%% file: sec/5_conclusion.tex
\section{Conclusion}
\label{sec:conclusion}
Our work offers a novel solution that significantly augments the effectiveness of UDA for semantic segmentation. Experiments on four different UDA benchmarks demonstrate that our approach brings the performance of lightweight models close to that of heavyweight models. This achievement heralds a considerable leap forward by unlocking substantial enhancements in efficiency and flexibility, especially when employing smaller models on edge devices.  We also show that the proposed framework DUDA is model-agnostic and can even be employed in a heterogeneous setting where CNN-based models are adapted from Transformer-based models. We believe that many resource-constrained applications requiring semantic segmentation can benefit from our work. We foresee the potential of our approach to effectively tackle a range of challenges in UDA, thereby contributing to the advancement of the field.

\vspace{0.1cm}
\noindent\textbf{Limitations.} DUDA utilizes multiple large models to support the training of the target lightweight model, leading to a higher memory requirement during training. Nonetheless, the inference cost remains the same as the lightweight baseline, while yielding substantial performance improvements. Alternatively, DUDA can train a significantly smaller model, maintaining comparable SOTA mIoU. 

%% file: sec/suppl.tex
\section{Overview}
\label{sec:overview}

This is the supplementary material to support our manuscript ``DUDA: Distilled Unsupervised Domain Adaptation
for Lightweight Semantic Segmentation". It contains additional and detailed results, particularly related to Sec. 4 (Experimental Results) of the main article, that couldn't be included in the main article due to space constraints. In Sec. \ref{sec:compared_methods}, we note references compared in Figure 1 in the main paper. In Sec. \ref{sec:training_cost}, we provide the training time of DUDA$_\text{MIC}$ with the various backbone models as an additional implementation detail. In Sec. \ref{sec:ablation_study}, in addition to Table 4 of the main paper, we provide more detailed ablation studies using MiT-B0 backbone across the four different datasets. In Sec. \ref{sec:additional_quantitative_results}, we compare DUDA models with the SOTA UDA methods with the class-wise IoU across four UDA different benchmarks. We also provide experiments comparing DUDA and its baselines (\textit{i.e.}, DAFormer, MIC) with MiT-B2 and MiT-B4 backbones. Additionally, we provide class-wise IoU scores for the compared ResNet-based methods in Table 2 of the main paper. 

\section{Compared Methods}
\label{sec:compared_methods}
We compare with the following methods in Figure 1 in the main paper: FDA \cite{yang2020fda},
DACS \cite{tranheden2021dacs}, 
ProDA \cite{zhang2021prototypical}, 
CAMix \cite{zhou2022context},
DAFormer \cite{hoyer2022daformer},
HRDA \cite{hoyer2022hrda}
MIC \cite{hoyer2023mic},
DiGA \cite{shen2023diga}, 
Fredom \cite{truong2023fredom}, 
SGG \cite{peng2023diffusion}, 
CONFETI \cite{li2023contrast}, 
RTea \cite{zhao2023learning}, 
PRN \cite{zhao2024unsupervised}, MoDA \cite{pan2024moda}, and RDAS (\textit{i.e.}, Revisiting Domain Adaptive Semantic Segmentation) \cite{jeong2024revisiting}.


\begin{table}[t]
\centering
\renewcommand{\arraystretch}{1.4}
\setlength{\tabcolsep}{2pt}
\fontsize{8.7pt}{8.7pt}\selectfont

\caption{Training time of DUDA$_\text{MIC}$ in the various backbones. The training is performed in a single NVIDIA A6000 or A5000 GPU. *Training time is measured in NVIDIA A5000 GPU.}

\begin{tabular}{llcc}
\hline
\multirow{2}{*}{Method} & \multirow{2}{*}{Backbone} & \multicolumn{2}{c}{Training Time (iterations)} \\
& & Pre-adaptation (40k) & Fine-tuning (80k) \\ 
\hline
DUDA\textbf{$_\text{MIC}$} & MiT-B0 & 39 hours & *41 hours \\
DUDA\textbf{$_\text{MIC}$} & MiT-B1 & 40 hours & *45 hours \\
DUDA\textbf{$_\text{MIC}$} & MiT-B2 & 43 hours &  46 hours \\
DUDA\textbf{$_\text{MIC}$} & MiT-B4 & 49 hours & 58 hours \\
DUDA\textbf{$_\text{MIC}$} & DeepLab-V2 & 54 hours & *73 hours \\
\hline
\end{tabular}

\label{table:training_time}
\end{table}

\begin{table}[t]
\centering
\renewcommand{\arraystretch}{1.4}
\setlength{\tabcolsep}{3.5pt}
\fontsize{7.6pt}{7.6pt}\selectfont

\caption{Ablation Studies of DUDA on the four different adaptation scenarios with DAFormer as the base across the four datasets. The basic model is configured as DAFormer without DUDA, and subsequently, we incrementally introduce cross-entropy, KL divergence, pre-adaptation, and inconsistency-based balanced losses in the DUDA setup. mIoU for Synthia$\rightarrow$Cityscapes is averaged over 16 classes.}


\begin{tabular}{l|c|ccc|cc}

\hline
\multirow{2}{*}{Method} & \multicolumn{1}{c|}{Pre-} & \multicolumn{3}{c|}{Fine-tuning} & \multirow{2}{*}{mIoU (\%)} & \multirow{2}{*}{mAcc (\%)} \\
& adaptation & CE & KL & Inconsistency & & \\

\hline
\multicolumn{7}{c}{\textbf{Synthetic-to-Real: GTA$\rightarrow$Cityscapes (Val.)}} \\

\hline

MiT-B0 & \multicolumn{4}{c|}{No Distillation} & 51.00 & 62.49 \\
MiT-B0 & & \checkmark & & & 62.34 & 71.74 \\
MiT-B0 & & \checkmark & \checkmark & & 63.67 & 72.94 \\
MiT-B0 & \checkmark & \checkmark & \checkmark & & 64.38 & 73.50 \\
MiT-B0 & \checkmark & \checkmark & \checkmark & \checkmark & 65.19 & 75.18 \\

\hline

\multicolumn{7}{c}{\textbf{Synthetic-to-Real: Synthia$\rightarrow$Cityscapes (Val.)}} \\

\hline

MiT-B0 & \multicolumn{4}{c|}{No Distillation} & 46.09 & 58.25 \\
MiT-B0 & & \checkmark & & & 57.06 & 67.44 \\
MiT-B0 & & \checkmark & \checkmark & & 57.01 & 68.44 \\
MiT-B0 & \checkmark & \checkmark & \checkmark & & 57.72 & 69.20 \\
MiT-B0 & \checkmark & \checkmark & \checkmark & \checkmark & 58.31 & 71.12 \\
\hline

\multicolumn{7}{c}{\textbf{Day-to-Nighttime: Cityscapes$\rightarrow$DarkZurich (Val.)}} \\

\hline

MiT-B0 & \multicolumn{4}{c|}{No Distillation} & 23.89 & 40.08 \\
MiT-B0 & & \checkmark & & & 33.18 & 49.02 \\
MiT-B0 & & \checkmark & \checkmark & & 34.30 & 50.49 \\
MiT-B0 & \checkmark & \checkmark & \checkmark & & 35.18 & 51.07 \\
MiT-B0 & \checkmark & \checkmark & \checkmark & \checkmark & 35.29 & 51.84 \\

\hline

\multicolumn{7}{c}{\textbf{Clear-to-Adverse-Weather: Cityscapes→ACDC (Val.)}} \\

\hline

MiT-B0 & \multicolumn{4}{c|}{No Distillation} & 43.79 & 58.06 \\
MiT-B0 & & \checkmark & & & 49.68 & 62.91 \\
MiT-B0 & & \checkmark & \checkmark & & 51.84 & 65.00 \\
MiT-B0 & \checkmark & \checkmark & \checkmark & & 53.52 & 66.11 \\
MiT-B0 & \checkmark & \checkmark & \checkmark & \checkmark & 53.86 & 67.45 \\

\hline

\end{tabular}
\label{table:ablation_study_daformer_4data}
\end{table}

\begin{table}[t]
\centering
\renewcommand{\arraystretch}{1.4}
\setlength{\tabcolsep}{3.5pt}
\fontsize{7.6pt}{7.6pt}\selectfont

\caption{Ablation Studies of DUDA on the four different adaptation scenarios with MIC as the base across the four datasets. The basic model is configured as MIC without DUDA, and subsequently, we incrementally introduce cross-entropy, KL divergence, pre-adaptation, and inconsistency-based balanced losses in the DUDA setup. mIoU for Synthia$\rightarrow$Cityscapes is averaged over 16 classes.}


\begin{tabular}{l|c|ccc|cc}

\hline
\multirow{2}{*}{Method} & \multicolumn{1}{c|}{Pre-} & \multicolumn{3}{c|}{Fine-tuning} & \multirow{2}{*}{mIoU (\%)} & \multirow{2}{*}{mAcc (\%)} \\
& adaptation & CE & KL & Inconsistency & & \\

\hline
\multicolumn{7}{c}{\textbf{Synthetic-to-Real: GTA$\rightarrow$Cityscapes (Val.)}} \\

\hline

MiT-B0 & \multicolumn{4}{c|}{No Distillation} & 59.54 & 69.29 \\
MiT-B0 & & \checkmark & & & 71.37 & 79.99 \\
MiT-B0 & & \checkmark & \checkmark & & 70.80 & 79.82 \\
MiT-B0 & \checkmark & \checkmark & \checkmark & & 71.63 & 80.27 \\
MiT-B0 & \checkmark & \checkmark & \checkmark & \checkmark & 71.71 & 81.04 \\

\hline

\multicolumn{7}{c}{\textbf{Synthetic-to-Real: Synthia$\rightarrow$Cityscapes (Val.)}} \\

\hline

MiT-B0 & \multicolumn{4}{c|}{No Distillation} & 52.96 & 64.26\\
MiT-B0 & & \checkmark & & & 64.79 & 75.03 \\
MiT-B0 & & \checkmark & \checkmark & & 64.56 & 74.93 \\
MiT-B0 & \checkmark & \checkmark & \checkmark & & 65.02 & 75.17 \\
MiT-B0 & \checkmark & \checkmark & \checkmark & \checkmark & 65.25 & 76.04 \\
\hline

\multicolumn{7}{c}{\textbf{Day-to-Nighttime: Cityscapes$\rightarrow$DarkZurich (Val.)}} \\

\hline

MiT-B0 & \multicolumn{4}{c|}{No Distillation} & 31.38 & 47.12 \\
MiT-B0 & & \checkmark & & & 40.35 & 60.00 \\
MiT-B0 & & \checkmark & \checkmark & & 40.93 & 60.74 \\
MiT-B0 & \checkmark & \checkmark & \checkmark & & 41.28 & 60.31 \\
MiT-B0 & \checkmark & \checkmark & \checkmark & \checkmark & 40.83 & 60.75 \\

\hline

\multicolumn{7}{c}{\textbf{Clear-to-Adverse-Weather: Cityscapes→ACDC (Val.)}} \\

\hline

MiT-B0 & \multicolumn{4}{c|}{No Distillation} & 52.82 & 64.89 \\
MiT-B0 & & \checkmark & & & 63.86 & 74.23 \\
MiT-B0 & & \checkmark & \checkmark & & 64.13 & 74.43 \\
MiT-B0 & \checkmark & \checkmark & \checkmark & & 65.48 & 75.22 \\
MiT-B0 & \checkmark & \checkmark & \checkmark & \checkmark & 65.39 & 75.73 \\

\hline

\end{tabular}
\label{table:ablation_study_mic_4data}
\end{table}

\section{Training and Inference Cost}
\label{sec:training_cost}

\noindent\textbf{Training Cost.} We primarily measure the training time for each of the two training procedures. Our training is performed on a single GPU, either NVIDIA A5000 or A6000, and other training parameters, such as batch size, are the same as introduced in the main text. Table \ref{table:training_time} summarizes the training time for different backbone models trained by DUDA$_\text{MIC}$ using the GTA$\rightarrow$Cityscapes dataset. Since DUDA is operated on three different networks (LT, LS, and SS) and consists of the two training stages (pre-adaptation and fine-tuning), the training cost is more expensive than that of its baseline, such as HRDA \cite{hoyer2022hrda} and MIC \cite{hoyer2023mic}. As LT and LS networks are identical regardless of the SS network's backbone, the increase in the training time (both pre-adaptation and fine-tuning) has resulted from the larger SS networks (from top to bottom rows). 

\noindent\textbf{Inference Cost.} We acknowledge the elevated memory requirement of DUDA during training due to the auxiliary large network, however, it incurs no increase in inference cost. Notably, the primary obstacle from memory issues predominantly emerges at inference time. Considering the other UDA methods~\cite{zhang2021prototypical,liu2021bapa,yang2020fda} take several days~\cite{hoyer2022hrda}, the training speed of DUDA is rather similar to them. However, our focus is to obtain SOTA UDA performance in lightweight models (SS networks), not fast training. We can rather expect fast inference as a by-product of lightweight models since DUDA reduces the memory by 1.3$\sim$11.7 times and FLOPs of the backbone by 1.3$\sim$21.2, keeping the mIoU comparable (Table 1 of the main paper). Note, the memory and FLOPs of ResNet models are as follows: ResNet-18 (46.2MB and 176.0 GFLOPs), ResNet-50 (99.7MB and 366.1GFLOPs), and ResNet-101 (175.7MB and 638.7 GFLOPs). FLOPs are measured using the same method as in Table 1 in the main paper.

In summary, while DUDA's training cost is higher than that of the baseline methods, the inference cost remains the same.

\begin{table}[t]
\centering
\renewcommand{\arraystretch}{1.4}
\setlength{\tabcolsep}{3.5pt}
\fontsize{7.6pt}{7.6pt}\selectfont

\caption{Ablation Studies of DUDA in GTA$\rightarrow$Cityscapes with DAFormer. Experimental setups are same with Table \ref{table:ablation_study_daformer_4data} but MiT-B2 and MiT-B4 models are used for students.}


\begin{tabular}{l|c|ccc|c}

\hline
\multirow{2}{*}{Method} & \multicolumn{1}{c|}{Pre-} & \multicolumn{3}{c|}{Fine-tuning} & \multirow{2}{*}{mIoU (\%)} \\
& adaptation & CE & KL & Inconsistency & \\

\hline
\multicolumn{6}{c}{\textbf{Synthetic-to-Real: GTA$\rightarrow$Cityscapes (Val.)}} \\

\hline

MiT-B2 & \multicolumn{4}{c|}{No Distillation} & 63.9 \\
MiT-B2 & & \checkmark & & & 68.4 \\
MiT-B2 & & \checkmark & \checkmark & & 68.5 \\
MiT-B2 & \checkmark & \checkmark & \checkmark & & 69.5 \\
MiT-B2 & \checkmark & \checkmark & \checkmark & \checkmark & 69.8 \\

\hline

\multicolumn{6}{c}{\textbf{Synthetic-to-Real: GTA$\rightarrow$Cityscapes (Val.)}} \\

\hline

MiT-B4 & \multicolumn{4}{c|}{No Distillation} & 66.1 \\
MiT-B4 & & \checkmark & & & 70.0 \\
MiT-B4 & & \checkmark & \checkmark & & 70.2 \\
MiT-B4 & \checkmark & \checkmark & \checkmark & & 70.4 \\
MiT-B4 & \checkmark & \checkmark & \checkmark & \checkmark & 70.5 \\
\hline

\end{tabular}
\label{table:ablation_study_daformer_b2_b4}
\end{table}

\begin{table*}[t]
\centering
\renewcommand{\arraystretch}{1.45}
\setlength{\tabcolsep}{4pt}
\fontsize{6.5pt}{6.5pt}\selectfont

\caption {Comparison of DUDA with DAFormer and MIC in MiT-B0 and B1 backbones in class-wise IoU. mIoU for Synthia$\rightarrow$Cityscapes is averaged over 16 classes.}


\resizebox{1.0\textwidth}{!}{
\begin{tabular}{>{\arraybackslash}p{1.4cm}|>{\centering\arraybackslash}m{0.10cm}|>{\centering\arraybackslash}m{0.40cm}>{\centering\arraybackslash}m{0.60cm}>{\centering\arraybackslash}m{0.50cm}>{\centering\arraybackslash}m{0.40cm}>{\centering\arraybackslash}m{0.50cm}>{\centering\arraybackslash}m{0.40cm}>{\centering\arraybackslash}m{0.70cm}>{\centering\arraybackslash}m{0.40cm}>{\centering\arraybackslash}m{0.50cm}>{\centering\arraybackslash}m{0.70cm}>{\centering\arraybackslash}m{0.30cm}>{\centering\arraybackslash}m{0.60cm}>{\centering\arraybackslash}m{0.50cm}>{\centering\arraybackslash}m{0.30cm}>{\centering\arraybackslash}m{0.50cm}>{\centering\arraybackslash}m{0.30cm}>{\centering\arraybackslash}m{0.50cm}>{\centering\arraybackslash}m{0.60cm}>{\centering\arraybackslash}m{0.40cm}|>{\centering\arraybackslash}m{0.50cm}}
\hline

Method & & Road & S.Walk & Build. & Wall & Fence & Pole & Tr.Light & Sign & Veget. & Terrain & Sky & Person & Rider & Car & Truck & Bus & Train & M.bike & Bike & mIoU \\

\hline
\multicolumn{22}{c}{\textbf{Synthetic-to-Real: GTA$\rightarrow$Cityscapes (Val.)}} \\
\hline

DAFormer \cite{hoyer2022daformer} & \multirow{4}{*}{\rotatebox[origin=c]{90}{MiT-B0}} & 92.2 & 55.9 & 85.6 & 25.2 & 22.3 & 40.0 & 39.5 & 46.2 & 87.3 & 43.6 & 87.7 & 63.4 & 31.8 & 85.4 & 36.4 & 40.6 & 1.4 & 31.8 & 52.9 & 51.0 \\

\textbf{DUDA}\textbf{$_\text{DAF}$} & & 96.3 & 76.7 & 88.5 & 43.0 & 41.7 & 48.5 & 49.5 & 59.6 & 89.7 & 44.4 & 91.7 & 68.9 & 40.3 & 91.2 & 72.6 & 72.6 & 52.8 & 52.4 & 61.2 & 65.2 \\

MIC \cite{hoyer2023mic} & & 95.0 & 68.0 & 88.4 & 44.4 & 29.4 & 48.6 & 48.5 & 62.5 & 89.8 & 46.0 & 92.7 & 71.0 & 36.6 & 87.5 & 49.2 & 57.7 & 0.4 & 53.6 & 62.1 & 59.5 \\

\textbf{DUDA}\textbf{$_\text{MIC}$} & & 97.1 & 78.3 & 90.6 & 56.3 & 51.0 & 56.8 & 58.3 & 68.1 & 91.2 & 49.2 & 93.7 & 76.4 & 50.0 & 93.3 & 76.7 & 82.4 & 71.2 & 56.7 & 65.5 & 71.7 \\

\hline

DAFormer \cite{hoyer2022daformer} & \multirow{4}{*}{\rotatebox[origin=c]{90}{MiT-B1}} & 94.4 & 64.8 & 87.1 & 34.1 & 27.2 & 44.7 & 47.7 & 55.0 & 88.7 & 47.3 & 90.3 & 66.4 & 32.1 & 89.2 & 59.9 & 55.6 & 52.0 & 47.6 & 58.8 & 60.2 \\

\textbf{DUDA}\textbf{$_\text{DAF}$} & & 96.7 & 75.8 & 89.2 & 47.7 & 46.8 & 50.9 & 52.9 & 63.5 & 90.2 & 45.3 & 92.7 & 70.9 & 42.8 & 92.3 & 76.0 & 79.1 & 69.8 & 56.2 & 62.2 & 68.5 \\

MIC \cite{hoyer2023mic} & & 95.8 & 72.0 & 89.9 & 54.4 & 40.3 & 55.8 & 59.7 & 70.2 & 90.9 & 50.5 & 93.8 & 75.3 & 46.1 & 91.5 & 62.8 & 64.5 & 35.4 & 60.6 & 65.5 & 67.1 \\

\textbf{DUDA}\textbf{$_\text{MIC}$} & & 97.3 & 79.6 & 91.0 & 54.4 & 53.9 & 59.0 & 62.2 & 70.8 & 91.6 & 50.2 & 93.9 & 78.3 & 53.7 & 94.2 & 82.3 & 84.5 & 70.8 & 58.1 & 67.0 & 73.3 \\

\hline
\multicolumn{22}{c}{\textbf{Synthetic-to-Real: Synthia$\rightarrow$Cityscapes (Val.)}} \\
\hline

DAFormer \cite{hoyer2022daformer} & \multirow{4}{*}{\rotatebox[origin=c]{90}{MiT-B0}} & 57.2 & 21.6 & 84.1 & 9.9 & 1.0 & 40.2 &	34.4 & 40.6 & 84.1 & - & 86.5 & 65.4 & 32.3 & 81.6 & - & 36.6 & - & 7.4 & 54.5 & 46.1 \\

\textbf{DUDA}\textbf{$_\text{DAF}$} & & 75.9 & 31.3 & 88.1 & 41.9 & 7.6 & 48.4 & 50.4 & 52.2 & 84.3 & - & 91.4 & 70.3 & 43.7 & 86.9 & - & 54.6 & - & 45.1 & 61.0 & 58.3 \\

MIC \cite{hoyer2023mic} & & 83.8 & 39.0 & 86.1 & 0.2 & 0.9 & 48.1 &	52.0 & 49.5 & 85.9 & - & 93.5 & 71.6 & 35.1 & 86.3 & - & 47.9 & - & 7.3 & 60.1 & 53.0 \\

\textbf{DUDA}\textbf{$_\text{MIC}$} & & 85.7 & 52.5 & 88.9 & 43.9 &	8.6 & 56.8 & 62.0 & 61.2 & 82.9 & - & 94.5 & 78.4 & 53.1 & 89.7 & - & 62.1 & - & 60.7 & 63.0 & 65.3 \\

\hline

DAFormer \cite{hoyer2022daformer} & \multirow{4}{*}{\rotatebox[origin=c]{90}{MiT-B1}} & 85.8 & 36.6 & 85.9 & 32.0 & 2.3 & 43.1 & 47.0 & 47.5 & 86.1 & - & 92.1 & 69.9 & 37.2 & 84.2 & - & 37.2 & - & 39.8 & 59.5 & 55.4 \\

\textbf{DUDA}\textbf{$_\text{DAF}$} & & 77.5 & 33.7 & 88.6 & 43.2 & 8.9 & 51.0 & 53.6 & 56.1 & 84.1 & - & 90.9 & 72.8 & 48.9 & 86.9 & - & 59.1 & - & 49.7 & 63.4 & 60.5 \\

MIC \cite{hoyer2023mic} & & 94.8 & 69.5	& 87.2 & 38.7 &	1.5 & 55.0 & 60.3 & 57.8 & 87.5 & - & 94.3 & 76.5 &	46.0 & 88.9 & - & 61.4 & - & 58.2 &	63.2 & 65.0 \\

\textbf{DUDA}\textbf{$_\text{MIC}$} & & 85.2 & 52.4 & 89.5 & 46.9 & 7.8 & 59.7 & 65.4 & 63.7 & 82.4 & - & 95.0 & 80.2 &	57.9 & 90.1 & - & 64.8 & - & 62.9 & 64.3 & 66.8 \\

\hline
\multicolumn{22}{c}{\textbf{Day-to-Nighttime: Cityscapes$\rightarrow$DarkZurich (Test)}} \\
\hline

DAFormer \cite{hoyer2022daformer} & \multirow{4}{*}{\rotatebox[origin=c]{90}{MiT-B0}} & 88.4 & 51.1 & 61.9 & 25.8 & 11.3 & 45.0 & 29.8 & 13.8 & 44.9 & 12.0 & 38.5 & 31.8 & 10.0 & 68.2 & 19.7 & 0.0 & 56.1 & 9.2 & 19.2 & 33.5 \\

\textbf{DUDA}\textbf{$_\text{DAF}$} & & 92.5 & 64.3 & 71.7 & 41.0 & 16.1 & 50.4 & 43.4 & 45.9 & 57.9 & 37.6 & 64.9 & 47.3 & 45.9 & 76.9 & 49.6 & 0.5 & 78.8 & 34.2 & 33.0 & 50.1 \\

MIC \cite{hoyer2023mic} & & 91.5 & 59.2 & 65.0 & 44.0 & 14.8 & 46.4 & 10.7 & 33.3 & 52.6 & 34.5 & 51.5 & 43.6 & 17.0 & 52.2 & 0.0 & 0.0 & 62.8 & 7.5 & 26.1 & 37.5 \\

\textbf{DUDA}\textbf{$_\text{MIC}$} & & 94.7 & 73.7 & 79.5 & 49.5 & 17.7 & 57.3 & 32.0 & 49.0 & 57.1 & 39.7 & 68.2 & 58.2 & 49.1 & 79.9 & 78.9 & 1.8 & 86.2 & 31.4 & 38.7 & 54.9 \\

\hline

DAFormer \cite{hoyer2022daformer} & \multirow{4}{*}{\rotatebox[origin=c]{90}{MiT-B1}} & 91.0 & 55.2 & 50.8 & 35.2 & 12.5 & 38.4 & 30.4 & 29.6 & 29.7 & 28.5 & 21.3 & 32.2 & 22.5 & 66.8 & 59.0 & 0.0 & 56.9 & 9.0 & 27.7 & 36.7 \\

\textbf{DUDA}\textbf{$_\text{DAF}$} & & 93.1 & 65.3 & 73.1 & 40.0 & 18.8 & 52.3 & 45.4 & 46.2 & 58.7 & 40.6 & 65.8 & 54.3 & 30.6 & 79.3 & 51.3 & 3.0 & 86.3 & 42.9 & 36.3 & 51.8 \\

MIC \cite{hoyer2023mic} & & 91.7 & 61.8 & 70.5 & 44.1 & 17.8 & 51.0 & 19.6 & 39.0 & 45.1 & 34.5 & 54.0 & 51.0 & 14.6 & 33.8 & 75.3 & 0.0 & 82.1 & 24.6 & 29.3 & 44.2 \\

\textbf{DUDA}\textbf{$_\text{MIC}$} & & 94.5 & 72.4 & 80.9 & 50.7 & 21.8 & 60.3 & 33.3 & 53.0 & 58.0 & 38.4 & 69.0 & 62.1 & 53.0 & 80.4 & 72.9 & 11.5 & 86.1 & 38.1 & 41.3 & 56.7 \\ 

\hline
\multicolumn{22}{c}{\textbf{Clear-to-Adverse-Weather: Cityscapes$\rightarrow$ACDC (Test)}} \\
\hline

DAFormer \cite{hoyer2022daformer} & \multirow{4}{*}{\rotatebox[origin=c]{90}{MiT-B0}} & 79.0 & 36.0 & 66.1 & 26.9 & 23.3 & 41.9 & 47.2 & 46.0 & 80.2 & 45.4 & 85.4 & 38.4 & 13.2 & 69.6 & 37.4 & 33.3 & 28.7 & 18.5 & 37.0 & 44.9 \\

\textbf{DUDA}\textbf{$_\text{DAF}$} & & 64.0 & 55.2 & 83.0 & 40.0 & 33.9 & 48.2 & 27.8 & 54.5 & 74.1 & 51.3 & 60.0 & 54.1 & 26.9 & 80.6 & 51.6 & 45.7 & 80.3 & 30.6 & 46.5 & 53.1 \\

MIC \cite{hoyer2023mic} & & 66.7 & 48.9 & 74.3 & 42.6 & 23.4 & 47.5 &	58.9 & 57.0 & 82.4 & 53.4 & 67.5 & 43.2 & 18.0 & 75.0 & 51.1 & 42.5 & 66.2 & 21.6 & 42.1 & 51.7 \\

\textbf{DUDA}\textbf{$_\text{MIC}$} & & 90.2 & 65.7 & 87.5 & 48.8 &	35.0 & 53.4 & 59.1 & 62.9 & 75.1 & 58.8 & 87.0 & 62.3 & 41.5 & 85.0 & 60.3 & 67.8 & 86.1 & 42.4 & 55.4 & 64.4 \\

\hline

DAFormer \cite{hoyer2022daformer} & \multirow{4}{*}{\rotatebox[origin=c]{90}{MiT-B1}} & 80.6 & 39.0 & 73.2 & 31.9 & 26.2 & 44.1 & 49.3 & 52.0 & 69.5 & 48.1 & 85.3 & 44.8 & 17.4 & 67.7 & 36.3 & 44.6 & 58.5 & 25.1 & 45.6 & 49.4 \\

\textbf{DUDA}\textbf{$_\text{DAF}$} & & 64.6 & 57.5 & 83.6 & 40.8 & 34.8 & 50.2 & 28.9 & 56.2 & 74.7 & 53.7 & 60.1 & 56.9 & 30.8 & 81.6 & 52.6 & 47.4 & 82.8 & 35.0 & 48.0 & 54.7 \\

MIC \cite{hoyer2023mic} & & 56.0 & 52.4 & 81.0 & 45.0 & 31.1 & 51.0 & 60.4 & 58.0 & 73.8 & 54.5 & 58.5 & 59.2 & 39.4 & 77.5 & 58.7 & 55.3 & 78.7 & 38.0 & 53.0 & 56.9 \\

\textbf{DUDA}\textbf{$_\text{MIC}$} & & 90.7 & 66.4 & 88.1 & 48.5 & 37.3 & 55.9 & 60.2 & 65.7 & 75.7 & 59.4 & 87.0 & 66.1 & 46.4 & 86.4 & 63.6 & 69.0 & 89.5 & 48.3 & 59.7 & 66.5 \\

\hline

\end{tabular}}
\label{table:b0b1}
\end{table*}

\begin{table*}[t]
\centering
\renewcommand{\arraystretch}{1.4}
\setlength{\tabcolsep}{4pt}
\fontsize{6.5pt}{6.5pt}\selectfont

\caption {Comparison of DUDA with DAFormer and MIC in MiT-B2 and B4 backbones in class-wise IoU. mIoU for Synthia$\rightarrow$Cityscapes is averaged over 16 classes.}


\resizebox{1.0\textwidth}{!}{
\begin{tabular}{>{\arraybackslash}p{1.4cm}|>{\centering\arraybackslash}m{0.10cm}|>{\centering\arraybackslash}m{0.40cm}>{\centering\arraybackslash}m{0.60cm}>{\centering\arraybackslash}m{0.50cm}>{\centering\arraybackslash}m{0.40cm}>{\centering\arraybackslash}m{0.50cm}>{\centering\arraybackslash}m{0.40cm}>{\centering\arraybackslash}m{0.70cm}>{\centering\arraybackslash}m{0.40cm}>{\centering\arraybackslash}m{0.50cm}>{\centering\arraybackslash}m{0.70cm}>{\centering\arraybackslash}m{0.30cm}>{\centering\arraybackslash}m{0.60cm}>{\centering\arraybackslash}m{0.50cm}>{\centering\arraybackslash}m{0.30cm}>{\centering\arraybackslash}m{0.50cm}>{\centering\arraybackslash}m{0.30cm}>{\centering\arraybackslash}m{0.50cm}>{\centering\arraybackslash}m{0.60cm}>{\centering\arraybackslash}m{0.40cm}|>{\centering\arraybackslash}m{0.50cm}}
\hline

Method & & Road & S.Walk & Build. & Wall & Fence & Pole & Tr.Light & Sign & Veget. & Terrain & Sky & Person & Rider & Car & Truck & Bus & Train & M.bike & Bike & mIoU \\

\hline
\multicolumn{22}{c}{\textbf{Synthetic-to-Real: GTA$\rightarrow$Cityscapes (Val.)}} \\
\hline

DAFormer \cite{hoyer2022daformer} & \multirow{4}{*}{\rotatebox[origin=c]{90}{MiT-B2}} & 94.9 & 63.8 & 88.6 & 47.2 & 34.9 & 48.3 & 54.1 & 57.9 & 89.6 & 49.6 & 91.0 & 67.8 & 39.9 & 90.7 & 61.2 & 67.1 & 59.0 & 50.4 & 58.4 & 63.9 \\

\textbf{DUDA}\textbf{$_\text{DAF}$} & & 97.0 & 76.9 & 89.8 & 53.4 & 48.5 & 52.7 & 55.5 & 64.5 & 90.3 & 44.5 & 93.1 & 72.2 & 45.3 & 93.0 & 79.9 & 83.3 & 68.2 & 54.7 & 63.0 & 69.8 \\

MIC \cite{hoyer2023mic} & & 96.8 & 76.4 & 90.9 & 57.4 & 51.3 & 58.8 & 63.9 & 70.6 & 91.4 & 50.0 & 94.1 & 77.0 & 50.0 & 93.9 & 80.0 & 84.2 & 70.2 & 59.1 & 65.1 & 72.7 \\

\textbf{DUDA}\textbf{$_\text{MIC}$} & & 97.5 & 80.7 & 91.7 & 61.7 & 57.0 & 60.6 & 64.3 & 71.3 & 91.8 & 51.5 & 94.0 & 79.6 & 56.1 & 94.5 & 83.7 & 90.0 & 80.1 & 61.2 & 68.1 & 75.5 \\

\hline

DAFormer \cite{hoyer2022daformer} & \multirow{4}{*}{\rotatebox[origin=c]{90}{MiT-B4}} & 93.9 & 59.8 & 88.9 & 49.6 & 44.4 & 48.9 & 55.7 & 56.8 & 89.3 & 49.3 & 92.2 & 71.5 & 42.1 & 91.7 & 59.8 & 77.0 & 67.3 & 57.5 & 60.0 & 66.1 \\

\textbf{DUDA}\textbf{$_\text{DAF}$} & & 97.0 & 77.2 & 90.1 & 54.7 & 51.3 & 53.0 & 57.0 & 65.1 & 90.5 & 45.2 & 93.0 & 73.2 & 45.9 & 93.2 & 81.4 & 82.5 & 64.6 & 60.1 & 64.7 & 70.5 \\

MIC \cite{hoyer2023mic} & & 96.4 & 75.7 & 91.8 & 61.0 & 58.4 & 59.8 & 65.3 & 73.2 & 92.0 & 52.7 & 93.9 & 79.2 & 52.5 & 93.6 & 76.7 & 80.9 & 74.5 & 65.6 & 67.6 & 74.3 \\

\textbf{DUDA}\textbf{$_\text{MIC}$} & & 97.5 & 80.7 & 92.0 & 63.6 & 59.5 & 61.4 & 65.5 & 72.0 & 91.9 & 51.8 & 94.1 & 80.4 & 57.3 & 94.7 & 87.0 & 91.1 & 82.9 & 64.9 & 68.9 & 76.7 \\

\hline
\multicolumn{22}{c}{\textbf{Synthetic-to-Real: Synthia$\rightarrow$Cityscapes (Val.)}} \\
\hline

DAFormer \cite{hoyer2022daformer} & \multirow{4}{*}{\rotatebox[origin=c]{90}{MiT-B2}} & 89.7 & 46.9 & 86.8 & 36.0 & 3.8 & 48.4 &	52.6 & 45.1 & 85.8 & - & 92.7 & 72.5 & 41.6 & 86.8 & - & 53.0 & - & 47.6 & 60.7 & 59.4 \\

\textbf{DUDA}\textbf{$_\text{DAF}$} & & 78.0 & 32.9 & 89.0 & 43.0 & 8.3 & 52.3 & 56.6 & 56.7 & 86.1 & - & 90.7 & 74.5 & 49.8 & 86.7 & - & 62.2 & - & 54.4 & 64.0 & 61.6 \\

MIC \cite{hoyer2023mic} & & 91.2 & 58.5 & 89.0 & 44.0 & 3.1 & 57.8 & 65.3 & 64.8 & 88.7 & - & 94.3 & 79.5 & 53.4 & 89.1 & - & 56.9 & - & 61.7 & 64.5 & 66.4 \\

\textbf{DUDA}\textbf{$_\text{MIC}$} & & 84.9 & 51.3 & 90.0 & 47.7 &	8.7 & 61.7 & 67.6 & 64.4 & 83.1 & - & 95.0 & 81.5 & 60.4 & 89.2 & - & 62.2 & - & 64.6 & 65.1 & 67.3 \\

\hline

DAFormer \cite{hoyer2022daformer} & \multirow{4}{*}{\rotatebox[origin=c]{90}{MiT-B4}} & 85.9 & 41.9 & 88.4 & 38.4 & 6.1 & 50.1 & 54.9 & 56.7 & 87.4 & - & 85.2 & 72.7 & 45.5 & 87.1 & - & 51.8 & - & 51.6 & 54.9 & 59.9 \\

\textbf{DUDA}\textbf{$_\text{DAF}$} & & 78.0 & 32.9 & 88.9 & 43.0 & 8.0 & 52.3 & 57.7 & 57.0 & 86.2 & - & 90.8 & 74.7 & 50.3 & 86.9 & - & 66.4 & - & 54.6 & 64.0 & 62.0 \\

MIC \cite{hoyer2023mic} & & 86.3 & 49.5 & 88.4 & 39.9 &	9.6 & 60.2 & 67.8 & 63.5 & 88.8 & - & 94.2 & 80.1 & 56.2 & 89.5 & - & 53.3 & - & 65.1 & 63.5 & 66.0 \\

\textbf{DUDA}\textbf{$_\text{MIC}$} & & 85.5 & 51.5 & 90.2 & 45.5 & 9.5 & 62.4 & 69.1 & 65.2 & 84.2 & - & 95.0 & 82.0 & 61.5 & 89.9 & - & 69.7 & - & 67.0 & 65.3 & 68.3 \\

\hline
\multicolumn{22}{c}{\textbf{Day-to-Nighttime: Cityscapes$\rightarrow$DarkZurich (Test)}} \\
\hline

DAFormer \cite{hoyer2022daformer} & \multirow{4}{*}{\rotatebox[origin=c]{90}{MiT-B2}} & 92.3 & 57.7 & 66.5 & 28.6 & 18.0 & 51.3 & 9.7 & 40.4 & 43.7 & 27.9 & 46.7 & 42.7 & 36.8 & 74.9 & 63.6 & 0.0 & 77.3 & 36.5 & 34.0 & 44.7 \\

\textbf{DUDA}\textbf{$_\text{DAF}$} & & 93.6 & 68.1 & 75.4 & 45.4 & 17.2 & 53.8 & 45.6 & 49.9 & 58.7 & 39.8 & 66.1 & 50.9 & 47.5 & 81.5 & 53.9 & 3.2 & 89.3 & 55.4 & 37.4 & 54.4 \\

MIC \cite{hoyer2023mic} & & 92.6 & 70.4 & 81.3 & 53.6 & 21.1 & 57.3 & 48.1 & 53.2 & 65.1 & 39.6 & 79.0 & 58.4 & 53.1 & 53.5 & 83.5 & 0.0 & 86.1 & 42.3 & 36.1 & 56.5 \\

\textbf{DUDA}\textbf{$_\text{MIC}$} & & 95.0 & 75.1 & 82.1 & 53.6 & 24.2 & 61.6 & 35.0 & 56.7 & 58.1 & 43.1 & 69.2 & 64.5 & 59.9 & 81.3 & 81.0 & 6.1 & 90.4 & 53.2 & 43.0 & 59.6 \\

\hline

DAFormer \cite{hoyer2022daformer} & \multirow{4}{*}{\rotatebox[origin=c]{90}{MiT-B4}} & 92.7 & 63.5 & 65.9 & 34.7 & 11.5 & 48.1 & 17.0 & 44.4 & 44.4 & 25.1 & 39.0 & 54.5 & 52.7 & 76.7 & 47.6 & 2.7 & 89.0 & 42.6 & 39.9 & 46.9 \\

\textbf{DUDA}\textbf{$_\text{DAF}$} & & 93.7 & 68.2 & 75.7 & 42.6 & 19.3 & 53.8 & 43.5 & 47.0 & 61.3 & 37.2 & 66.8 & 56.6 & 54.9 & 81.0 & 52.3 & 3.3 & 90.1 & 48.4 & 38.3 & 54.4 \\

MIC \cite{hoyer2023mic} & & 95.3 & 76.7 & 83.0 & 55.4 & 25.0 & 63.0 & 35.4 & 57.5 & 59.1 & 44.2 & 70.5 & 66.3 & 55.1 & 81.2 & 80.8 & 12.7 & 90.5 & 42.5 & 44.5 & 59.9 \\

\textbf{DUDA}\textbf{$_\text{MIC}$} & & 95.4 & 77.2 & 82.9 & 55.6 & 25.6 & 62.8 & 35.7 & 57.8 & 59.1 & 43.9 & 70.5 & 66.2 & 58.6 & 81.2 & 81.8 & 13.1 & 91.4 & 42.4 & 44.4 & 60.3 \\ 

\hline
\multicolumn{22}{c}{\textbf{Clear-to-Adverse-Weather: Cityscapes$\rightarrow$ACDC (Test)}} \\
\hline

DAFormer \cite{hoyer2022daformer} & \multirow{4}{*}{\rotatebox[origin=c]{90}{MiT-B2}} & 55.7 & 40.3 & 83.8 & 42.2 & 31.8 & 48.1 & 39.9 & 50.5 & 73.7 & 48.2 & 50.6 & 56.1 & 31.0 & 78.6 & 53.9 & 55.5 & 73.0 & 36.0 & 43.5 & 52.2 \\

\textbf{DUDA}\textbf{$_\text{DAF}$} & & 64.1 & 57.1 & 84.1 & 44.7 & 36.9 & 51.8 & 30.3 & 58.1 & 75.0 & 53.6 & 59.5 & 60.3 & 35.8 & 82.6 & 58.5 & 51.9 & 84.2 & 40.6 & 50.0 & 56.8 \\

MIC \cite{hoyer2023mic} & & 53.4 & 55.4 & 81.7 & 53.9 & 37.8 & 55.3 & 59.5 & 62.3 & 80.0 & 55.7 & 56.7 & 63.8 & 39.1 & 82.7 & 71.3 & 67.2 & 81.9 & 43.9 & 55.9 & 60.9 \\

\textbf{DUDA}\textbf{$_\text{MIC}$} & & 91.0 & 67.4 & 88.7 & 50.7 & 39.4 & 57.8 & 60.9 & 67.2 & 76.2 & 60.7 & 87.0 & 69.5 & 48.1 & 88.1 & 71.7 & 78.4 & 90.3 & 51.3 & 59.8 & 68.6 \\

\hline

DAFormer \cite{hoyer2022daformer} & \multirow{4}{*}{\rotatebox[origin=c]{90}{MiT-B4}} & 69.0 & 34.9 & 84.4 & 44.3 & 32.4 & 50.9 & 32.0 & 57.0 & 72.2 & 41.6 & 72.6 & 58.5 & 35.3 & 81.0 & 54.1 & 66.1 & 81.1 & 38.3 & 49.1 & 55.5 \\

\textbf{DUDA}\textbf{$_\text{DAF}$} & & 63.2 & 57.9 & 85.0 & 47.6 & 36.6 & 52.2 & 29.6 & 58.2 & 75.1 & 54.4 & 57.6 & 61.8 & 36.9 & 83.3 & 59.4 & 58.7 & 85.7 & 41.9 & 51.0 & 57.7 \\

MIC \cite{hoyer2023mic} & & 52.8 & 62.9 & 86.1 & 58.8 & 41.3 & 55.9 & 53.1 & 59.0 & 74.9 & 58.1 & 47.9 & 69.6 & 46.6 & 86.3 & 75.7 & 84.0 & 89.9 & 52.7 & 61.3 & 64.0 \\

\textbf{DUDA}\textbf{$_\text{MIC}$} & & 91.4 & 68.8 & 89.3 & 52.3 & 40.4 & 59.2 & 61.3 & 68.6 & 76.4 & 62.1 & 87.1 & 71.5 & 48.6 & 89.3 & 76.7 & 83.8 & 90.6 & 55.5 & 61.5 & 70.2 \\

\hline

\end{tabular}}
\label{table:b2b4}
\end{table*}

\begin{table*}[t]
\centering

\renewcommand{\arraystretch}{1.4}
\setlength{\tabcolsep}{4pt}
\fontsize{6.5pt}{6.5pt}\selectfont

\caption {Comparison of DUDA with prior UDA Semantic Segmentation methods in SegFormer-based networks in class-wise IoU. mIoU for Synthia$\rightarrow$Cityscapes is averaged over 16 classes.}


\resizebox{1.0\textwidth}{!}{
\begin{tabular}{>{\arraybackslash}p{1.5cm}|>{\centering\arraybackslash}m{0.40cm}>{\centering\arraybackslash}m{0.60cm}>{\centering\arraybackslash}m{0.50cm}>{\centering\arraybackslash}m{0.40cm}>{\centering\arraybackslash}m{0.50cm}>{\centering\arraybackslash}m{0.40cm}>{\centering\arraybackslash}m{0.70cm}>{\centering\arraybackslash}m{0.40cm}>{\centering\arraybackslash}m{0.50cm}>{\centering\arraybackslash}m{0.70cm}>{\centering\arraybackslash}m{0.30cm}>{\centering\arraybackslash}m{0.60cm}>{\centering\arraybackslash}m{0.50cm}>{\centering\arraybackslash}m{0.30cm}>{\centering\arraybackslash}m{0.50cm}>{\centering\arraybackslash}m{0.30cm}>{\centering\arraybackslash}m{0.50cm}>{\centering\arraybackslash}m{0.60cm}>{\centering\arraybackslash}m{0.40cm}|>{\centering\arraybackslash}m{0.50cm}}
\hline

Method & Road & S.Walk & Build. & Wall & Fence & Pole & Tr.Light & Sign & Veget. & Terrain & Sky & Person & Rider & Car & Truck & Bus & Train & M.bike & Bike & mIoU \\

\hline

\multicolumn{21}{c}{\textbf{Synthetic-to-Real: GTA$\rightarrow$Cityscapes (Val.)}} \\

\hline

HRDA \cite{hoyer2022hrda} & 96.4 & 74.4 & 91.0 & 61.6 & 51.5 & 57.1 & 63.9 & 69.3 & 91.3 & 48.4 & 94.2 & 79.0 & 52.9 & 93.9 & 84.1 & 85.7 & 75.9 & 63.9 & 67.5 & 73.8 \\

DiGA \cite{shen2023diga} & 97.0 & 78.6 & 91.3 & 60.8 & 56.7 & 56.5 & 64.4 & 69.9 & 91.5 & 50.8 & 93.7 & 79.2 & 55.2 & 93.7 & 78.3 & 86.9 & 77.8 & 63.7 & 65.8 & 74.3 \\

GANDA \cite{liao2023geometry} & 96.5 & 74.8 & 91.4 & 61.7 & 57.3 & 59.2 & 65.4 & 68.8 & 91.5 & 49.9 & 94.7 & 79.6 & 54.8 & 94.1 & 81.3 & 86.8 & 74.6 & 64.8 & 68.2 & 74.5 \\

RTea \cite{zhao2023learning} & 97.1 & 75.2 & 92.6 & 63.5 & 51.8 & 58.2 & 66.5 & 71.2 & 91.1 & 49.0 & 96.8 & 81.5 & 54.2 & 94.2 & 84.8 & 86.6 & 75.7 & 62.2 & 66.7 & 74.7 \\

BLV \cite{wang2023balancing} & 96.7 & 76.6 & 91.5 & 61.2 & 56.9 & 59.4 & 62.2 & 72.8 & 91.5 & 51.2 & 94.3 & 77.5 & 54.7 & 93.5 & 83.2 & 84.7 & 79.7 & 68.1 & 67.6 & 74.9 \\

IR$^{2}$F-RMM \cite{gong2023continuous} & 97.5 & 80.0 & 91.0 & 60.0 & 53.3 & 56.2 & 63.9 & 72.4 & 91.7 & 51.0 & 94.2 & 79.0 & 51.1 & 94.3 & 84.7 & 86.7 & 75.9 & 62.6 & 67.8 & 74.4 \\

CDAC \cite{wang2023cdac} & 97.1 & 78.7 & 91.8 & 59.6 & 57.1 & 59.1 & 66.1 & 72.2 & 91.8 & 53.1 & 94.5 & 79.4 & 51.6 & 94.6 & 84.9 & 87.8 & 78.7 & 64.9 & 67.6 & 75.3 \\

PiPa \cite{chen2023pipa} & 96.8 & 76.3 & 91.6 & 63.0 & 57.7 & 60.0 & 65.4 & 72.6 & 91.7 & 51.8 & 94.8 & 79.7 & 56.4 & 94.4 & 85.9 & 88.4 & 78.9 & 63.5 & 67.2 & 75.6 \\

MIC \cite{hoyer2023mic} & 97.4 & 80.1 & 91.7 & 61.2 & 56.9 & 59.7 & 66.0 & 71.3 & 91.7 & 51.4 & 94.3 & 79.8 & 56.1 & 94.6 & 85.4 & 90.3 & 80.4 & 64.5 & 68.5 & 75.9 \\

MICDrop \cite{yang2025micdrop} & 97.6 & 81.5 & 92.0 & 62.8 & 59.4 & 62.6 & 62.9 & 73.6 & 91.6 & 52.6 & 94.1 & 80.2 & 57.0 & 94.8 & 87.4 & 90.7 & 81.6 & 65.3 & 67.8 & 76.6 \\

\hline


\textbf{DUDA}\textbf{$_\text{MIC}$} (B4) & 97.5 & 80.7 & 92.0 & 63.6 & 59.5 & 61.4 & 65.5 & 72.0 & 91.9 & 51.8 & 94.1 & 80.4 & 57.3 & 94.7 & 87.0 & 91.1 & 82.9 & 64.9 & 68.9 & 76.7 \\

\textbf{DUDA}\textbf{$_\text{MIC}$} (B2) & 97.5 & 80.7 & 91.7 & 61.7 & 57.0 & 60.6 & 64.3 & 71.3 & 91.8 & 51.5 & 94.0 & 79.6 & 56.1 & 94.5 & 83.7 & 90.0 & 80.1 & 61.2 & 68.1 & 75.5 \\

\textbf{DUDA}\textbf{$_\text{MIC}$} (B1) & 97.3 & 79.6 & 91.0 & 54.4 & 53.9 & 59.0 & 62.2 & 70.8 & 91.6 & 50.2 & 93.9 & 78.3 & 53.7 & 94.2 & 82.3 & 84.5 & 70.8 & 58.1 & 67.0 & 73.3 \\

\textbf{DUDA}\textbf{$_\text{MIC}$} (B0) & 97.1 & 78.3 & 90.6 & 56.3 & 51.0 & 56.8 & 58.3 & 68.1 & 91.2 & 49.2 & 93.7 & 76.4 & 50.0 & 93.3 & 76.7 & 82.4 & 71.2 & 56.7 & 65.5 & 71.7 \\

\hline

\multicolumn{21}{c}{\textbf{Synthetic-to-Real: Synthia$\rightarrow$Cityscapes (Val.)}} \\

\hline
HRDA \cite{hoyer2022hrda} & 85.2 & 47.7 & 88.8 & 49.5 & 4.8 & 57.2 & 65.7 & 60.9 & 85.3 & - & 92.9 & 79.4 & 52.8 & 89.0 & - & 64.7 & - & 63.9 & 64.9 & 65.8 \\

DiGA \cite{shen2023diga} & 88.5 & 49.9 & 90.1 & 51.4 & 6.6 & 55.3 & 64.8 & 62.7 & 88.2 & - & 93.5 & 78.6 & 51.8 & 89.5 & - & 62.2 & - & 61.0 & 65.8 & 66.2 \\

GANDA \cite{liao2023geometry} & 89.1 & 50.6 & 89.7 & 51.4 & 6.7 & 59.4 & 66.8 & 57.7 & 86.7 & - & 93.8 & 80.6 & 56.9 & 90.7 & - & 64.8 & - & 62.6 & 65.0 & 67.0 \\

RTea \cite{zhao2023learning} & 87.8 & 49.0 & 90.3 & 50.3 & 5.5 & 58.6 & 66.0 & 61.4 & 86.8 & - & 93.1 & 79.5 & 53.1 & 89.5 & - & 65.1 & - & 63.7 & 64.6 & 66.5 \\

BLV \cite{wang2023balancing} & 87.6 & 47.9 & 90.5 & 50.4 & 6.9 & 57.1 & 64.3 & 65.3 & 86.9 & - & 93.4 & 78.9 & 54.9 & 89.1 & - & 62.9 & - & 65.2 & 66.8 & 66.8 \\

IR$^{2}$F-RMM \cite{gong2023continuous} & 90.4 & 54.9 & 89.4 & 48.0 & 7.4 & 59.0 & 65.5 & 63.2 & 87.8 & - & 94.1 & 80.5 & 55.8 & 90.0 & - & 65.9 & - & 64.5 & 66.8 & 67.7 \\ 

CDAC \cite{wang2023cdac} & 93.1 & 68.5 & 89.8 & 51.2 & 8.9 & 59.4 & 65.5 & 65.3 & 84.7 & - & 94.4 & 81.2 & 57.0 & 90.5 & - & 56.9 & - & 66.8 & 66.4 & 68.7 \\ 

PiPa \cite{chen2023pipa} & 88.6 & 50.1 & 90.0 & 53.8 & 7.7 & 58.1 & 67.2 & 63.1 & 88.5 & - & 94.5 & 79.7 & 57.6 & 90.8 & - & 70.2 & - & 65.1 & 66.9 & 68.2 \\ 

MIC \cite{hoyer2023mic} & 86.6 & 50.5 & 89.3 & 47.9 & 7.8 & 59.4 & 66.7 & 63.4 & 87.1 & - & 94.6 & 81.0 & 58.9 & 90.1 & - & 61.9 & - & 67.1 & 64.3 & 67.3 \\ 

MICDrop \cite{yang2025micdrop} & 82.8 & 42.6 & 90.5 & 51.6 & 9.6 & 61.0 & 65.7 & 65.0 & 89.1 & - & 95.0 & 81.1 & 59.7 & 90.6 & - & 68.3 & - & 67.4 & 66.5 & 67.9 \\

\hline


\textbf{DUDA}\textbf{$_\text{MIC}$} (B4) & 85.5 & 51.5 & 90.2 & 45.5 & 9.5 & 62.4 & 69.1 & 65.2 & 84.2 & - & 95.0 & 82.0 & 61.5 & 89.9 & - & 69.7 & - & 67.0 & 65.3 & 68.3 \\

\textbf{DUDA}\textbf{$_\text{MIC}$} (B2) & 84.9 & 51.3 & 90.0 & 47.7 &	8.7 & 61.7 & 67.6 & 64.4 & 83.1 & - & 95.0 & 81.5 & 60.4 & 89.2 & - & 62.2 & - & 64.6 & 65.1 & 67.3 \\

\textbf{DUDA}\textbf{$_\text{MIC}$} (B1) & 85.2 & 52.4 & 89.5 & 46.9 & 7.8 & 59.7 & 65.4 & 63.7 & 82.4 & - & 95.0 & 80.2 &	57.9 & 90.1 & - & 64.8 & - & 62.9 & 64.3 & 66.8 \\

\textbf{DUDA}\textbf{$_\text{MIC}$} (B0) & 85.7 & 52.5 & 88.9 & 43.9 &	8.6 & 56.8 & 62.0 & 61.2 & 82.9 & - & 94.5 & 78.4 & 53.1 & 89.7 & - & 62.1 & - & 60.7 & 63.0 & 65.3 \\

\hline

\multicolumn{21}{c}{\textbf{Day-to-Nighttime: Cityscapes$\rightarrow$DarkZurich (Test)}} \\

\hline
HRDA \cite{hoyer2022hrda} & 90.4 & 56.3 & 72.0 & 39.5 & 19.5 & 57.8 & 52.7 & 43.1 & 59.3 & 29.1 & 70.5 & 60.0 & 58.6 & 84.0 & 75.5 & 11.2 & 90.5 & 51.6 & 40.9 & 55.9 \\

IR$^{2}$F-RMM \cite{gong2023continuous} & 94.7 & 75.1 & 73.2 & 44.4 & 25.7 & 60.6 & 39.0 & 47.4 & 70.2 & 41.6 & 77.3 & 62.4 & 55.5 & 86.4 & 55.5 & 20.0 & 92.0 & 55.3 & 42.8 & 58.9 \\

MIC \cite{hoyer2023mic} & 94.8 & 75.0 & 84.0 & 55.1 & 28.4 & 62.0 & 35.5 & 52.6 & 59.2 & 46.8 & 70.0 & 65.2 & 61.7 & 82.1 & 64.2 & 18.5 & 91.3 & 52.6 & 44.0 & 60.2 \\

\hline

\textbf{DUDA}\textbf{$_\text{MIC}$} (B4) & 95.4 & 77.2 & 82.9 & 55.6 & 25.6 & 62.8 & 35.7 & 57.8 & 59.1 & 43.9 & 70.5 & 66.2 & 58.6 & 81.2 & 81.8 & 13.1 & 91.4 & 42.4 & 44.4 & 60.3 \\ 

\textbf{DUDA}\textbf{$_\text{MIC}$} (B2) & 95.0 & 75.1 & 82.1 & 53.6 & 24.2 & 61.6 & 35.0 & 56.7 & 58.1 & 43.1 & 69.2 & 64.5 & 59.9 & 81.3 & 81.0 & 6.1 & 90.4 & 53.2 & 43.0 & 59.6 \\

\textbf{DUDA}\textbf{$_\text{MIC}$} (B1) & 94.5 & 72.4 & 80.9 & 50.7 & 21.8 & 60.3 & 33.3 & 53.0 & 58.0 & 38.4 & 69.0 & 62.1 & 53.0 & 80.4 & 72.9 & 11.5 & 86.1 & 38.1 & 41.3 & 56.7 \\ 

\textbf{DUDA}\textbf{$_\text{MIC}$} (B0) & 94.7 & 73.7 & 79.5 & 49.5 & 17.7 & 57.3 & 32.0 & 49.0 & 57.1 & 39.7 & 68.2 & 58.2 & 49.1 & 79.9 & 78.9 & 1.8 & 86.2 & 31.4 & 38.7 & 54.9 \\

\hline

\multicolumn{21}{c}{\textbf{Clear-to-Adverse-Weather: Cityscapes$\rightarrow$ACDC (Test)}} \\

\hline
HRDA \cite{hoyer2022hrda} & 88.3 & 57.9 & 88.1 & 55.2 & 36.7 & 56.3 & 62.9 & 65.3 & 74.2 & 57.7 & 85.9 & 68.8 & 45.7 & 88.5 & 76.4 & 82.4 & 87.7 & 52.7 & 60.4 & 68.0 \\
CDAC \cite{wang2023cdac} & 87.0 & 56.7 & 84.5 & 53.5 & 34.3 & 54.6 & 43.6 & 51.4 & 71.7 & 58.6 & 85.4 & 68.7 & 45.7 & 89.0 & 70.9 & 81.5 & 90.1 & 47.6 & 59.0 & 64.9 \\
MIC \cite{hoyer2023mic} & 90.8 & 67.1 & 89.2 & 54.5 & 40.5 & 57.2 & 62.0 & 68.4 & 76.3 & 61.8 & 87.0 & 71.3 & 49.4 & 89.7 & 75.7 & 86.8 & 89.1 & 56.9 & 63.0 & 70.4 \\

\hline

\textbf{DUDA}\textbf{$_\text{MIC}$} (B4) & 91.4 & 68.8 & 89.3 & 52.3 & 40.4 & 59.2 & 61.3 & 68.6 & 76.4 & 62.1 & 87.1 & 71.5 & 48.6 & 89.3 & 76.7 & 83.8 & 90.6 & 55.5 & 61.5 & 70.2 \\

\textbf{DUDA}\textbf{$_\text{MIC}$} (B2) & 91.0 & 67.4 & 88.7 & 50.7 & 39.4 & 57.8 & 60.9 & 67.2 & 76.2 & 60.7 & 87.0 & 69.5 & 48.1 & 88.1 & 71.7 & 78.4 & 90.3 & 51.3 & 59.8 & 68.6 \\

\textbf{DUDA}\textbf{$_\text{MIC}$} (B1) & 90.7 & 66.4 & 88.1 & 48.5 & 37.3 & 55.9 & 60.2 & 65.7 & 75.7 & 59.4 & 87.0 & 66.1 & 46.4 & 86.4 & 63.6 & 69.0 & 89.5 & 48.3 & 59.7 & 66.5 \\

\textbf{DUDA}\textbf{$_\text{MIC}$} (B0) & 90.2 & 65.7 & 87.5 & 48.8 &	35.0 & 53.4 & 59.1 & 62.9 & 75.1 & 58.8 & 87.0 & 62.3 & 41.5 & 85.0 & 60.3 & 67.8 & 86.1 & 42.4 & 55.4 & 64.4 \\

\hline

\end{tabular}}
\label{table:class_wise_overview}
\end{table*}

\section{Ablation Study}
\label{sec:ablation_study}

Ablation studies are conducted in MiT-B0 backbone trained by DUDA$_\text{DAF}$ and DUDA$_\text{MIC}$ across the four different datasets. Similar to Table 4 of the main paper, the performance improvement by involving the four components, pre-adaptation (Pre-adapt) and fine-tuning with the cross-entropy (CE), KL divergence losses (KL), and inconsistency-based loss balancing (Incon.), are investigated. Table \ref{table:ablation_study_daformer_4data} and Table \ref{table:ablation_study_mic_4data} provide the results from DUDA$_\text{DAF}$ and DUDA$_\text{MIC}$, respectively. DUDA$_\text{DAF}$ with the four components consistently achieves the highest mIoU and mAcc in the four datasets. DUDA$_\text{MIC}$ with the four components in CS$\rightarrow$DZur and CS$\rightarrow$ACDC shows slightly lower mIoU but highest mAcc. 

In summary, DUDA generally shows continuous improvement by including each of the components.

\section{Additional Quantitative Results}
\label{sec:additional_quantitative_results}

\noindent\textbf{Comparison with Transformer-based Methods.} In Table 3 of the main paper, we compare DUDA and its baselines (DAFormer and MIC) in the MiT-B0, B1, B2, and B4 backbones in terms of mIoU scores. Here, we additionally provide the class-wise comparison in Tables \ref{table:b0b1} and \ref{table:b2b4}. It is important to note that, in GTA\(\rightarrow\)CS with MiT-B0 backbone, the methods without DUDA completely fail to segment the Train class, while DUDA$_\text{DAF}$ and DUDA$_\text{MIC}$ improve it by the IoU of almost \(~50\%\) and \(70\%\). Similarly, the accuracy of the Motorbike class shows a large improvement with DUDA (MiT-B0) across SYN\(\rightarrow\)CS, CS$\rightarrow$DZur, and CS$\rightarrow$ACDC. In a few cases, we see a performance drop, \textit{e.g.}, Road class in SYN\(\rightarrow\)CS and CS$\rightarrow$ACDC for DUDA$_\text{DAF}$. Overall, due to learning from higher-quality labels and inconsistency-based balancing, DUDA performs significantly better in most of the classes across benchmarks, with more substantial improvement generally observed in the minority classes. While the Train class is significantly improved ($\sim$50\%) in MiT-B0 and MiT-B1, the improvement is reduced since the baseline accuracy is high. The performance drop of the Road class in SYN$\rightarrow$CS and CS$\rightarrow$ACDC is again observed in MiT-B2 and MiT-B4. Overall, the observations and trends in the class-wise IoU of MiT-B0 and MiT-B1 are similar in the larger backbones.

In addition to the mIoU presented in Table 1 in Sec. 4 of the main paper, we provide the class-wise comparison of DUDA with the SOTA SegFormer-based methods in Table \ref{table:class_wise_overview}. It demonstrates the class-wise IoU of the SOTA methods and DUDA with MiT-B0, B1, B2, and B4 backbones across the four different UDA semantic segmentation benchmarks. Our DUDA$_\text{MIC}$ with MiT-B4 performs slightly better than recent SOTA methods, such as MIC \cite{hoyer2023mic} and MICDrop \cite{yang2025micdrop}, with MiT-B5 in on GTA$\rightarrow$CS and SYN$\rightarrow$CS. We believe this can be attributed to our inconsistency-weighted loss boosting accuracy on underperforming classes and the efficacy of learning from multiple teachers (LT and LS). DUDA also performs better than recent method CSI \cite{lim2024cross} in most experiments (\textit{e.g.}, with DAFormer base, mIoU of 69.8 w/ DUDA MiT-B2 vs. 67.9 w/ CSI MiT-B5 in GTA$\rightarrow$CS; mIoU of 61.6 w/ DUDA MiT-B2 vs 61.4 w/ CSI MiT-B5 in SYN$\rightarrow$CS). 

However, we omit comparison with MICDrop and CSI in the Tables to ensure fairness. The performance gain in MICDrop stems from additional geometric information, such as depth predictions, to enhance learning segmentation boundaries. Also, all the compared approaches in the Tables including ours assume consistent taxonomies between source and target domains, typical in traditional UDA semantic segmentation, whereas CSI considers inconsistent taxonomies between domains. Similarly, we do not directly compare with UDA methods leveraging foundation models \cite{ren2024cross, lim2024cross, yan2023sam4udass}. Lastly, InforMS \cite{wang2023informative} proposed Online Informative Class Sampling to dynamically adjust the weights of different semantic classes. However, it is particularly designed for daytime-nighttime adaptation scenarios, considering illumination discrepancies in the scene using spectrogram mean. In contrast, DUDA does not assume specific adaptation scenarios.

\noindent\textbf{Comparison with ResNet-based Methods.} The various ResNet-based methods reported in Table 2 in Sec. 4 (main paper) are compared with the class-wise IoU in Table \ref{table:class_wise_overview_dlv2}, including DUDA$_\text{MIC}$ with DeepLab-V2 (DLV2) backbone. In particular, the accuracy improvement is noticeable in GTA$\rightarrow$CS, and the Train class in DUDA$_\text{MIC}$ shows $\sim15\%$ higher IoU than other UDA methods. Similarly, DUDA$_\text{MIC}$ achieves $10\%$ higher IoU in certain classes, such as the Wall class in SYN$\rightarrow$CS and Car class in CS$\rightarrow$ACDC.

\begin{table*}[t]
\centering

\renewcommand{\arraystretch}{1.4}
\setlength{\tabcolsep}{4pt}
\fontsize{6.5pt}{6.5pt}\selectfont

\caption {Comparison of DUDA with prior UDA Semantic Segmentation methods in ResNet-based networks in class-wise IoU. mIoU for Synthia$\rightarrow$Cityscapes is averaged over 16 classes. SSG \cite{peng2023diffusion} and DAFormer \cite{hoyer2022daformer} in GTA$\rightarrow$Cityscapes are omitted since the class-wise IoU is not provided in the literature. *DAFormer results are implemented by ourselves.}


\resizebox{1.0\textwidth}{!}{
\begin{tabular}{>{\arraybackslash}p{1.7cm}|>{\centering\arraybackslash}m{0.40cm}>{\centering\arraybackslash}m{0.60cm}>{\centering\arraybackslash}m{0.50cm}>{\centering\arraybackslash}m{0.40cm}>{\centering\arraybackslash}m{0.50cm}>{\centering\arraybackslash}m{0.40cm}>{\centering\arraybackslash}m{0.70cm}>{\centering\arraybackslash}m{0.40cm}>{\centering\arraybackslash}m{0.50cm}>{\centering\arraybackslash}m{0.70cm}>{\centering\arraybackslash}m{0.30cm}>{\centering\arraybackslash}m{0.60cm}>{\centering\arraybackslash}m{0.40cm}>{\centering\arraybackslash}m{0.30cm}>{\centering\arraybackslash}m{0.50cm}>{\centering\arraybackslash}m{0.30cm}>{\centering\arraybackslash}m{0.40cm}>{\centering\arraybackslash}m{0.60cm}>{\centering\arraybackslash}m{0.40cm}|>{\centering\arraybackslash}m{0.40cm}}
\hline

Method & Road & S.Walk & Build. & Wall & Fence & Pole & Tr.Light & Sign & Veget. & Terrain & Sky & Person & Rider & Car & Truck & Bus & Train & M.bike & Bike & mIoU \\

\hline

\multicolumn{21}{c}{\textbf{Synthetic-to-Real: GTA$\rightarrow$Cityscapes (Val.)}} \\

\hline

ADVENT \cite{vu2019advent} & 89.4 & 33.1 & 81.0 & 26.6 & 26.8 & 27.2 & 33.5 & 24.7 & 83.9 & 36.7 & 78.8 & 58.7 & 30.5 & 84.8 & 38.5 & 44.5 & 1.7 & 31.6 & 32.4 & 45.5 \\

CBST \cite{zou2018unsupervised} & 89.6 & 58.9 & 78.5 & 33.0 & 22.3 & 41.4 & 48.2 & 39.2 & 83.6 & 24.3 & 65.4 & 49.3 & 20.2 & 83.3 & 39.0 & 48.6 & 12.5 & 20.3 & 35.3 & 47.0 \\

CRST \cite{zou2019confidence} & 91.7 & 45.1 & 80.9 & 29.0 & 23.4 & 43.8 & 47.1 & 40.9 & 84.0 & 20.0 & 60.6 & 64.0 & 31.9 & 85.8 & 39.5 & 48.7 & 25.0 & 38.0 & 47.0 & 49.8 \\

DACS \cite{tranheden2021dacs} & 89.9 & 39.7 & 87.9 & 30.7 & 39.5 & 38.5 & 46.4 & 52.8 & 88.0 & 44.0 & 88.8 & 67.2 & 35.8 & 84.5 & 45.7 & 50.2 & 0.0 & 27.3 & 34.0 & 52.1 \\

ProDA \cite{zhang2021prototypical} & 87.8 & 56.0 & 79.7 & 46.3 & 44.8 & 45.6 & 53.5 & 53.5 & 88.6 & 45.2 & 82.1 & 70.7 & 39.2 & 88.8 & 45.5 & 59.4 & 1.0 & 48.9 & 56.4 & 57.5 \\

Fredom \cite{truong2023fredom} & 90.9 & 54.1 & 87.8 & 44.1 & 32.6 & 45.2 & 51.4 & 57.1 & 88.6 & 42.6 & 89.5 & 68.8 & 40.0 & 89.7 & 58.4 & 62.6 & 55.3 & 47.7 & 58.1 & 61.3 \\

RTea \cite{zhao2023learning} & 95.4 & 67.1 & 87.9 & 46.1 & 44.0 & 46.0 & 53.8 & 59.5 & 89.7 & 49.8 & 89.8 & 71.5 & 40.5 & 90.8 & 55.0 & 57.9 & 22.1 & 47.7 & 62.5 & 61.9 \\

DiGA \cite{shen2023diga} & 95.6 & 67.4 & 89.8 & 51.6 & 38.1 & 52.0 & 59.0 & 51.5 & 86.4 & 34.5 & 87.7 & 75.6 & 48.8 & 92.5 & 66.5 & 63.8 & 19.7 & 49.6 & 61.6 & 62.7 \\

CONFETI \cite{li2023contrast} & 96.5 & 75.6 & 88.9 & 45.1 & 45.9 & 50.1 & 61.2 & 68.2 & 89.4 & 45.7 & 86.3 & 76.3 & 49.9 & 92.2 & 55.1 & 62.8 & 16.7 & 33.8 & 63.1 & 63.3 \\

HRDA \cite{hoyer2022hrda} & 96.2 & 73.1 & 89.7 & 43.2 & 39.9 & 47.5 & 60.0 & 60.0 & 89.9 & 47.1 & 90.2 & 75.9 & 49.0 & 91.8 & 61.9 & 59.3 & 10.2 & 47.0 & 65.3 & 63.0 \\

MIC \cite{hoyer2023mic} & 96.5 & 74.3 & 90.4 & 47.1 & 42.8 & 50.3 & 61.7 & 62.3 & 90.3 & 49.2 & 90.7 & 77.8 & 53.2 & 93.0 & 66.2 & 68.0 & 6.8 & 38.0 & 60.6 & 64.2 \\

\hline

\textbf{DUDA}\textbf{$_\text{MIC}$} (R101) & 97.7 & 80.9 & 91.1 & 49.8 & 55.5 & 57.7 & 62.6 & 70.0 & 91.4 & 50.9 & 94.1 & 78.6 & 56.7 & 94.2 & 81.9 & 85.5 & 71.2 & 59.8 & 66.9 & 73.5 \\
\textbf{DUDA}\textbf{$_\text{MIC}$} (R50) & 97.4 & 79.4 & 91.1 & 54.6 & 54.2 & 56.3 & 62.1 & 69.2 & 91.1 & 47.3 & 93.7 & 76.5 & 53.1 & 93.9 & 79.0 & 84.5 & 73.6 & 59.7 & 66.3 & 72.8 \\
\textbf{DUDA}\textbf{$_\text{MIC}$} (R18) & 97.0 & 77.0 & 89.8 & 47.8 & 49.1 & 54.3 & 57.8 & 66.7 & 90.7 & 47.1 & 92.7 & 73.8 & 48.7 & 93.2 & 73.5 & 79.1 & 60.8 & 54.8 & 63.1 & 69.3 \\

\hline

\multicolumn{21}{c}{\textbf{Synthetic-to-Real: Synthia$\rightarrow$Cityscapes (Val.)}} \\

\hline

ADVENT \cite{vu2019advent} & 85.6 & 42.2 &79.7 & 8.7 & 0.4 & 25.9 & 5.4 & 8.1 & 80.4 & - & 84.1 & 57.9 & 23.8 & 73.3 & - & 36.4 & - & 14.2 & 33.0 & 41.2 \\

CBST \cite{zou2018unsupervised} & 53.6 & 23.7 & 75.0 & 12.5 & 0.3 & 36.4 & 23.5 & 26.3 & 84.8 & - & 74.7 & 67.2 & 17.5 & 84.5 & - & 28.4 & - & 15.2 & 55.8 & 42.5 \\

CRST \cite{zou2019confidence} & 67.7 & 32.2 & 73.9 & 10.7 & 1.6 & 37.4 & 22.2 & 31.2 & 80.8 & - & 80.5 & 60.8 & 29.1 & 82.8 & - & 25.0 & - & 19.4 & 45.3 & 43.8 \\

DACS \cite{tranheden2021dacs} & 80.6 & 25.1 & 81.9 & 21.5 & 2.9 & 37.2 & 22.7 & 24.0 & 83.7 & - & 90.8 & 67.6 & 38.3 & 82.9 & - & 38.9 & - & 28.5 & 47.6 & 48.3 \\

ProDA \cite{zhang2021prototypical} & 87.8 & 45.7 & 84.6 & 37.1 & 0.6 & 44.0 & 54.6 & 37.0 & 88.1 & - & 84.4 & 74.2 & 24.3 & 88.2 & - & 51.1 & - & 40.5 & 45.6 & 55.5\\

*DAFormer \cite{hoyer2022daformer} & 62.1 & 24.7 & 85.2 & 24.5 & 3.7 & 38.6 & 44.8 & 50.9 & 84.9 & - & 84.1 & 69.6 & 40.6 & 86.1 & - & 51.7 & - & 46.5 & 55.2 & 55.3 \\

GANDA \cite{liao2023geometry} & 87.1 & 45.8 & 86.1 & 28.9 & 4.8 & 37.1 & 40.6 & 45.0 & 87.0 & - & 87.9 & 69.1 & 39.8 & 89.9 & - & 59.8 & - & 33.8 & 57.2 & 56.3 \\

Fredom \cite{truong2023fredom} & 86.0 & 46.3 & 87.0 & 33.3 & 5.3 & 48.7 & 53.4 & 46.8 & 87.1 & - & 89.1 & 71.2 & 38.1 & 87.1 & - & 54.6 & - & 51.3 & 59.9 & 59.1 \\

RTea \cite{zhao2023learning} & 93.2 & 59.6 & 86.3 & 31.3 & 4.8 & 43.1 & 41.8 & 44.0 & 88.6 & - & 90.5 & 70.4 & 42.6 & 89.5 & - & 56.7 & - & 40.2 & 59.9 & 58.9 \\

DiGA \cite{shen2023diga} & 89.1 & 53.4 & 86.1 & 28.7 & 3.0 & 49.6 & 50.6 & 34.9 & 88.2 & - & 84.9 & 71.3 & 40.9 & 91.6 & - & 75.1 & - & 50.3 & 65.8 & 60.2 \\

CONFETI \cite{li2023contrast} & 83.8 & 44.6 & 86.9 & 15.4 & 3.7 & 44.3 & 56.9 & 55.5 & 84.9 & - & 86.2 & 73.8 & 46.8 & 90.1 & - & 57.1 & - & 46.0 & 63.2 & 58.7 \\

HRDA \cite{hoyer2022hrda} & 85.8 & 47.3 & 87.3 & 27.3 & 1.4 & 50.5 & 57.8 & 61.0 & 87.4 & - & 89.1 & 76.2 & 48.5 & 87.3 & - & 49.3 & - & 55.0 & 68.2 & 61.2 \\

MIC \cite{hoyer2023mic} & 84.7 & 45.7 & 88.3 & 29.9 &	2.8	& 53.3 & 61.0 & 59.5 & 86.9 & - & 88.8 & 78.2 & 53.3 & 89.4 & - & 58.8 & - & 56.0 & 68.3 & 62.8 \\

\hline

\textbf{DUDA}\textbf{$_\text{MIC}$} (R101) & 88.7 & 54.2 & 90.0 & 47.9 &	8.6 & 59.0 & 65.7 & 62.9 & 86.8 & - & 94.3 & 80.5 & 59.5 & 90.4 & - & 62.6 & - & 61.1 & 65.5 & 67.4 \\
\textbf{DUDA}\textbf{$_\text{MIC}$} (R50) & 88.3 & 50.5 & 89.7 & 48.6 & 9.1 & 57.6 & 65.3 & 62.3 & 89.3 & - & 94.0 & 79.1 & 58.2 & 90.2 & - & 65.5 & - & 62.6 & 65.6 & 67.2 \\
\textbf{DUDA}\textbf{$_\text{MIC}$} (R18) & 87.4 & 48.7 & 88.5 & 42.1 & 9.0 & 52.3 & 58.8 & 58.4 & 88.9 & - & 93.6 & 75.0 & 52.8 & 88.8 & - & 58.1 & - & 53.9 & 62.2 & 63.7 \\

\hline

\multicolumn{21}{c}{\textbf{Day-to-Nighttime: Cityscapes$\rightarrow$DarkZurich (Test)}} \\

\hline

ADVENT \cite{vu2019advent} & 85.8 & 37.9 & 55.5 & 27.7 & 14.5 & 23.1 & 14.0 & 21.1 & 32.1 & 8.7 & 2.0 & 39.9 & 16.6 & 64.0 & 13.8 & 0.0 & 58.8 & 28.5 & 20.7 & 29.7\\

MGCDA \cite{sakaridis2020map} & 80.3 & 49.3 & 66.2 & 7.8 & 11.0 & 41.4 & 38.9 & 39.0 & 64.1 & 18.0 & 55.8 & 52.1 & 53.5 & 74.7 & 66.0 & 0.0 & 37.5 & 29.1 & 22.7 & 42.5 \\

DANNet \cite{wu2021dannet} & 90.0 & 54.0 & 74.8 & 41.0 & 21.1 & 25.0 & 26.8 & 30.2 & 72.0 & 26.2 & 84.0 & 47.0 & 33.9 & 68.2 & 19.0 & 0.3 & 66.4 & 38.3 & 23.6 & 44.3 \\

*DAFormer \cite{hoyer2022daformer} & 84.2 & 56.5 & 67.4 & 32.5 & 14.8 & 46.1 & 32.6 & 44.7 & 33.8 & 23.3 & 1.8 & 50.2 & 43.0 & 74.7 & 69.4 & 8.5 & 54.3 & 28.6 & 36.0 & 44.2 \\

HRDA \cite{hoyer2022hrda} & 88.7 & 65.5 & 68.3 & 41.9 & 18.1 & 50.6 & 6.0 & 39.6 & 33.3 & 34.4 & 0.3 & 57.6 & 51.7 & 75.0 & 70.9 & 8.5 & 63.6 & 41.0 & 38.8 & 44.9 \\

MIC \cite{hoyer2023mic} & 82.8 & 69.6 & 75.5 & 44.0 & 21.0 & 51.1 & 43.4 & 48.3 & 39.3 & 37.1 & 0.0 & 59.4 & 53.6 & 73.6 & 74.2 & 9.2 & 78.7 & 40.0 & 37.2 & 49.4\\

\hline

\textbf{DUDA}\textbf{$_\text{MIC}$} (R101) & 94.1 & 72.2 & 78.0 & 45.6 & 23.8 & 58.0 & 32.7 & 52.8 & 56.4 & 33.1 & 68.9 & 63.1 & 55.8 & 76.4 & 85.4 & 12.1 & 61.9 & 38.7 & 40.3 & 55.2 \\
\textbf{DUDA}\textbf{$_\text{MIC}$} (R50) & 93.5 & 69.1 & 78.7 & 47.1 & 19.6 & 57.3 & 32.4 & 49.2 & 55.0 & 36.1 & 68.0 & 62.4 & 52.9 & 76.9 & 75.0 & 1.8 & 74.8 & 41.2 & 39.7 & 54.2 \\
\textbf{DUDA}\textbf{$_\text{MIC}$} (R18) & 92.8 & 66.0 & 76.0 & 42.7 & 20.1 & 53.4 & 30.5 & 46.5 & 51.7 & 33.8 & 65.4 & 53.8 & 38.7 & 71.3 & 53.8 & 3.1 & 68.0 & 32.7 & 35.7 & 49.3 \\

\hline

\multicolumn{21}{c}{\textbf{Clear-to-Adverse-Weather: Cityscapes$\rightarrow$ACDC (Test)}} \\

\hline

ADVENT \cite{vu2019advent} & 72.9 & 14.3 & 40.5 & 16.6 & 21.2 & 9.3 & 17.4 & 21.2 & 63.8 & 23.8 & 18.3 & 32.6 & 19.5 & 69.5 & 36.2 & 34.5 & 46.2 & 26.9 & 36.1 & 32.7 \\

MGCDA \cite{sakaridis2020map} & 73.4 & 28.7 & 69.9 & 19.3 & 26.3 & 36.8 & 53.0 & 53.3 & 75.4 & 32.0 & 84.6 & 51.0 & 26.1 & 77.6 & 43.2 & 45.9 & 53.9 & 32.7 & 41.5 & 48.7 \\

DANNet \cite{wu2021dannet} & 84.3 & 54.2 & 77.6 & 38.0 & 30.0 & 18.9 & 41.6 & 35.2 & 71.3 & 39.4 & 86.6 & 48.7 & 29.2 & 76.2 & 41.6 & 43.0 & 58.6 & 32.6 & 43.9 & 50.0 \\

*DAFormer \cite{hoyer2022daformer} & 76.3 & 48.1 & 78.0 & 34.7 & 26.9 & 38.3 & 50.7 & 52.8 & 70.0 & 45.1 & 78.4 & 54.5 & 28.3 & 78.1 & 47.9 & 43.6 & 70.3 & 22.6 & 49.2 & 52.3 \\

HRDA \cite{hoyer2022hrda} & 84.9 & 63.2 & 83.1 & 33.1 & 32.3 & 46.0 & 42.7 & 55.4 & 69.2 & 52.8 & 83.1 & 63.2 & 37.8 & 78.1 & 48.5 & 58.5 & 62.4 & 42.8 & 57.2 & 57.6 \\

MIC \cite{hoyer2023mic} & 88.7 & 63.9 & 84.1 & 38.4 & 35.7 & 45.7 & 51.5 & 60.3 & 72.7 & 52.3 & 85.8 & 62.5 & 39.8 & 84.7 & 37.7 & 68.7 & 71.9 & 46.0 & 56.5 & 60.4 \\

\hline
\textbf{DUDA}\textbf{$_\text{MIC}$} (R101) & 89.9 & 64.7 & 87.1 & 40.3 & 37.4 & 55.3 & 61.9 & 66.9 & 75.2 & 58.8 & 87.0 & 66.3 & 40.7 & 87.5 & 63.6 & 75.3 & 85.9 & 50.8 & 57.7 & 65.9 \\
\textbf{DUDA}\textbf{$_\text{MIC}$} (R50) & 89.3 & 63.2 & 86.4 & 40.3 & 37.1 & 54.3 & 59.9 & 65.8 & 75.2 & 57.5 & 87.0 & 65.9 & 39.5 & 85.9 & 56.4 & 66.4 & 79.4 & 47.6 & 56.1 & 63.9 \\
\textbf{DUDA}\textbf{$_\text{MIC}$} (R18) & 88.2 & 60.0 & 84.6 & 38.4 & 34.3 & 49.9 & 53.8 & 54.2 & 74.1 & 54.9 & 86.8 & 58.6 & 33.8 & 79.1 & 47.1 & 36.8 & 75.7 & 35.1 & 50.5 & 57.7 \\
\hline

\end{tabular}}
\label{table:class_wise_overview_dlv2}
\end{table*}

\section{Additional Discussion}
\label{sec:additional_discussion}

\noindent\textbf{Vanilla KD baseline using LT.} The vanilla KD baseline (row-3) in the ablation study tables (e.g., Table 4 of the main paper) uses LS for distillation. We observe comparable performance using LT for distillation. For example, with LT, mIoU for MiT-B1 degrades slightly (-0.07\%), and mIoU for MiT-B0 improves slightly (+0.04\%) in GTA$\rightarrow$CS compared to using LS.

\noindent\textbf{KL Loss in pre-adaptation.} KL loss enriches learning by capturing class correlations using output logits. Note, we incorporate KL loss in the pre-adaptation. In GTA$\rightarrow$CS for DUDA$_{\text{DAF}}$, the KL in the pre-adaptation improves the final mIoU of MiT-B0 (65.1$\rightarrow$65.2\%) and MiT-B1 (68.2$\rightarrow$68.5\%).

\noindent\textbf{MiT-B5 as both Large (LS and LT) and Small (SS) Networks.} As DUDA-B4 outperforms MIC-B5 in GTA$\rightarrow$CS and SYN$\rightarrow$CS (Table 1 in the main script), we evaluate DUDA-B5 to check whether the performance improvement continues. DUDA$_\text{MIC}$ with MiT-B5 achieves mIoU of 76.7 in GTA$\rightarrow$CS and 68.6 in SYN$\rightarrow$CS. DUDA$_\text{DAF}$ with MiT-B5 achieves mIoU of 70.9 in GTA$\rightarrow$CS and 61.6 in SYN$\rightarrow$CS. DUDA-B5 models perform very similarly to DUDA-B4 (\textit{i.e.,} marginally better overall), indicating the performance gain is now saturated. It makes sense since, in DUDA-B5, all three networks (LS, LT, and SS) are MiT-B5 with the same capacity.

\noindent\textbf{LS/LT as Teacher at Pre-adaptation and Fine-tuning.} As we have two large networks (LT and LS) to guide a lightweight network (SS), four different setups can be considered in KD teacher, \textit{e.g.}, LT/LS at the pre-adaptation and LT/LS at the fine-tuning. DUDA employs two large networks (LT and LS) in a multi-teacher setup, where different large networks are used at the pre-adaptation and fine-tuning phases. Our experiments show the multi-teacher setup performs slightly better than the same-teacher setup. Relative mIoU differences are as follows: (multi-teacher) LT$\rightarrow$LS (DUDA): +0.0\% LS$\rightarrow$LT: +0.1\% and (same-teacher) LT$\rightarrow$LT: -0.5\% LS$\rightarrow$LS: -0.2\%, showing the multi-teacher setup in DUDA is effective.

\noindent\textbf{Upper Bound of Small Model's Performance.} In order to better understand the contribution of the UDA side, we investigate the upper bound of small models when distilled by large models in a supervised setup. When distilled from a large MiT-B5 (trained in a \textit{supervised} setup on Cityscapes), the performance of the smaller models is: mIoU of 75.4 for MiT-B0, mIoU of 77.5 for MiT-B1. In contrast, our GTA$\rightarrow$Cityscapes DUDA$_\text{MIC}$ models, trained \textit{unsupervised}, exhibit only a marginal performance decrease (See Table 1 in the main paper, or Table \ref{table:class_wise_overview}): MiT-B0 mIoU 71.7, MiT-B1 mIoU 73.3.

\noindent\textbf{Static Inconsistency-based Weighting during Fine-tuning.} The general approach to balanced loss involves statically weighting the loss based on the distribution in the entire dataset, and our method follows this practice. Also, during the fine-tuning stage with inconsistency-based balanced loss, the class distribution remains constant since the teacher model is frozen.


%% file: sample-sigconf-authordraft.bbl

\begin{thebibliography}{79}


\ifx \showCODEN    \undefined \def \showCODEN     #1{\unskip}     \fi
\ifx \showDOI      \undefined \def \showDOI       #1{#1}\fi
\ifx \showISBNx    \undefined \def \showISBNx     #1{\unskip}     \fi
\ifx \showISBNxiii \undefined \def \showISBNxiii  #1{\unskip}     \fi
\ifx \showISSN     \undefined \def \showISSN      #1{\unskip}     \fi
\ifx \showLCCN     \undefined \def \showLCCN      #1{\unskip}     \fi
\ifx \shownote     \undefined \def \shownote      #1{#1}          \fi
\ifx \showarticletitle \undefined \def \showarticletitle #1{#1}   \fi
\ifx \showURL      \undefined \def \showURL       {\relax}        \fi
\providecommand\bibfield[2]{#2}
\providecommand\bibinfo[2]{#2}
\providecommand\natexlab[1]{#1}
\providecommand\showeprint[2][]{arXiv:#2}

\bibitem[Bai et~al\mbox{.}(2022)]%
        {bai2022dynamically}
\bibfield{author}{\bibinfo{person}{Haoli Bai}, \bibinfo{person}{Hongda Mao}, {and} \bibinfo{person}{Dinesh Nair}.} \bibinfo{year}{2022}\natexlab{}.
\newblock \showarticletitle{Dynamically pruning segformer for efficient semantic segmentation}. In \bibinfo{booktitle}{\emph{ICASSP 2022-2022 IEEE International Conference on Acoustics, Speech and Signal Processing (ICASSP)}}. IEEE, \bibinfo{pages}{3298--3302}.
\newblock


\bibitem[Beyer et~al\mbox{.}(2022)]%
        {beyer2022knowledge}
\bibfield{author}{\bibinfo{person}{Lucas Beyer}, \bibinfo{person}{Xiaohua Zhai}, \bibinfo{person}{Am{\'e}lie Royer}, \bibinfo{person}{Larisa Markeeva}, \bibinfo{person}{Rohan Anil}, {and} \bibinfo{person}{Alexander Kolesnikov}.} \bibinfo{year}{2022}\natexlab{}.
\newblock \showarticletitle{Knowledge distillation: A good teacher is patient and consistent}. In \bibinfo{booktitle}{\emph{Proc. of IEEE/CVF conference on computer vision and pattern recognition}}. \bibinfo{pages}{10925--10934}.
\newblock


\bibitem[Chen et~al\mbox{.}(2017)]%
        {chen2017deeplab}
\bibfield{author}{\bibinfo{person}{Liang-Chieh Chen}, \bibinfo{person}{George Papandreou}, \bibinfo{person}{Iasonas Kokkinos}, \bibinfo{person}{Kevin Murphy}, {and} \bibinfo{person}{Alan~L Yuille}.} \bibinfo{year}{2017}\natexlab{}.
\newblock \showarticletitle{Deeplab: Semantic image segmentation with deep convolutional nets, atrous convolution, and fully connected crfs}.
\newblock \bibinfo{journal}{\emph{IEEE transactions on pattern analysis and machine intelligence}} \bibinfo{volume}{40}, \bibinfo{number}{4} (\bibinfo{year}{2017}), \bibinfo{pages}{834--848}.
\newblock


\bibitem[Chen et~al\mbox{.}(2019)]%
        {chen2019domain}
\bibfield{author}{\bibinfo{person}{Minghao Chen}, \bibinfo{person}{Hongyang Xue}, {and} \bibinfo{person}{Deng Cai}.} \bibinfo{year}{2019}\natexlab{}.
\newblock \showarticletitle{Domain adaptation for semantic segmentation with maximum squares loss}. In \bibinfo{booktitle}{\emph{Proc. of IEEE/CVF International Conference on Computer Vision}}. \bibinfo{pages}{2090--2099}.
\newblock


\bibitem[Chen et~al\mbox{.}(2023)]%
        {chen2023pipa}
\bibfield{author}{\bibinfo{person}{Mu Chen}, \bibinfo{person}{Zhedong Zheng}, \bibinfo{person}{Yi Yang}, {and} \bibinfo{person}{Tat-Seng Chua}.} \bibinfo{year}{2023}\natexlab{}.
\newblock \showarticletitle{Pipa: Pixel-and patch-wise self-supervised learning for domain adaptative semantic segmentation}. In \bibinfo{booktitle}{\emph{Proc. of 31st ACM International Conference on Multimedia}}. \bibinfo{pages}{1905--1914}.
\newblock


\bibitem[Chen et~al\mbox{.}(2022)]%
        {chen2022mtp}
\bibfield{author}{\bibinfo{person}{Xinghao Chen}, \bibinfo{person}{Yiman Zhang}, {and} \bibinfo{person}{Yunhe Wang}.} \bibinfo{year}{2022}\natexlab{}.
\newblock \showarticletitle{MTP: multi-task pruning for efficient semantic segmentation networks}. In \bibinfo{booktitle}{\emph{2022 IEEE International Conference on Multimedia and Expo (ICME)}}. IEEE, \bibinfo{pages}{1--6}.
\newblock


\bibitem[Choi et~al\mbox{.}(2019)]%
        {choi2019pseudo}
\bibfield{author}{\bibinfo{person}{Jaehoon Choi}, \bibinfo{person}{Minki Jeong}, \bibinfo{person}{Taekyung Kim}, {and} \bibinfo{person}{Changick Kim}.} \bibinfo{year}{2019}\natexlab{}.
\newblock \showarticletitle{Pseudo-labeling curriculum for unsupervised domain adaptation}.
\newblock \bibinfo{journal}{\emph{arXiv preprint arXiv:1908.00262}} (\bibinfo{year}{2019}).
\newblock


\bibitem[Cordts et~al\mbox{.}(2016)]%
        {cordts2016cityscapes}
\bibfield{author}{\bibinfo{person}{Marius Cordts}, \bibinfo{person}{Mohamed Omran}, \bibinfo{person}{Sebastian Ramos}, \bibinfo{person}{Timo Rehfeld}, \bibinfo{person}{Markus Enzweiler}, \bibinfo{person}{Rodrigo Benenson}, \bibinfo{person}{Uwe Franke}, \bibinfo{person}{Stefan Roth}, {and} \bibinfo{person}{Bernt Schiele}.} \bibinfo{year}{2016}\natexlab{}.
\newblock \showarticletitle{The cityscapes dataset for semantic urban scene understanding}. In \bibinfo{booktitle}{\emph{Proc. of IEEE conference on computer vision and pattern recognition}}. \bibinfo{pages}{3213--3223}.
\newblock


\bibitem[Dong et~al\mbox{.}(2021)]%
        {dong2021and}
\bibfield{author}{\bibinfo{person}{Jiahua Dong}, \bibinfo{person}{Yang Cong}, \bibinfo{person}{Gan Sun}, \bibinfo{person}{Zhen Fang}, {and} \bibinfo{person}{Zhengming Ding}.} \bibinfo{year}{2021}\natexlab{}.
\newblock \showarticletitle{Where and how to transfer: knowledge aggregation-induced transferability perception for unsupervised domain adaptation}.
\newblock \bibinfo{journal}{\emph{IEEE Transactions on Pattern Analysis and Machine Intelligence}} (\bibinfo{year}{2021}).
\newblock


\bibitem[Feng et~al\mbox{.}(2020)]%
        {feng2020admp}
\bibfield{author}{\bibinfo{person}{Xiaoyu Feng}, \bibinfo{person}{Zhuqing Yuan}, \bibinfo{person}{Guijin Wang}, {and} \bibinfo{person}{Yongpan Liu}.} \bibinfo{year}{2020}\natexlab{}.
\newblock \showarticletitle{Admp: An adversarial double masks based pruning framework for unsupervised cross-domain compression}.
\newblock \bibinfo{journal}{\emph{arXiv preprint arXiv:2006.04127}} (\bibinfo{year}{2020}).
\newblock


\bibitem[Gao et~al\mbox{.}(2021)]%
        {gao2021dsp}
\bibfield{author}{\bibinfo{person}{Li Gao}, \bibinfo{person}{Jing Zhang}, \bibinfo{person}{Lefei Zhang}, {and} \bibinfo{person}{Dacheng Tao}.} \bibinfo{year}{2021}\natexlab{}.
\newblock \showarticletitle{Dsp: Dual soft-paste for unsupervised domain adaptive semantic segmentation}. In \bibinfo{booktitle}{\emph{Proc. of 29th ACM international conference on multimedia}}. \bibinfo{pages}{2825--2833}.
\newblock


\bibitem[Gong et~al\mbox{.}(2023)]%
        {gong2023continuous}
\bibfield{author}{\bibinfo{person}{Rui Gong}, \bibinfo{person}{Qin Wang}, \bibinfo{person}{Martin Danelljan}, \bibinfo{person}{Dengxin Dai}, {and} \bibinfo{person}{Luc Van~Gool}.} \bibinfo{year}{2023}\natexlab{}.
\newblock \showarticletitle{Continuous Pseudo-Label Rectified Domain Adaptive Semantic Segmentation With Implicit Neural Representations}. In \bibinfo{booktitle}{\emph{Proc. of IEEE/CVF Conference on Computer Vision and Pattern Recognition}}. \bibinfo{pages}{7225--7235}.
\newblock


\bibitem[He et~al\mbox{.}(2016)]%
        {he2016deep}
\bibfield{author}{\bibinfo{person}{Kaiming He}, \bibinfo{person}{Xiangyu Zhang}, \bibinfo{person}{Shaoqing Ren}, {and} \bibinfo{person}{Jian Sun}.} \bibinfo{year}{2016}\natexlab{}.
\newblock \showarticletitle{Deep residual learning for image recognition}. In \bibinfo{booktitle}{\emph{Proc. of IEEE conference on computer vision and pattern recognition}}. \bibinfo{pages}{770--778}.
\newblock


\bibitem[He et~al\mbox{.}(2019)]%
        {he2019knowledge}
\bibfield{author}{\bibinfo{person}{Tong He}, \bibinfo{person}{Chunhua Shen}, \bibinfo{person}{Zhi Tian}, \bibinfo{person}{Dong Gong}, \bibinfo{person}{Changming Sun}, {and} \bibinfo{person}{Youliang Yan}.} \bibinfo{year}{2019}\natexlab{}.
\newblock \showarticletitle{Knowledge adaptation for efficient semantic segmentation}. In \bibinfo{booktitle}{\emph{Proc. of IEEE/CVF Conference on Computer Vision and Pattern Recognition}}. \bibinfo{pages}{578--587}.
\newblock


\bibitem[Hoyer et~al\mbox{.}(2022a)]%
        {hoyer2022daformer}
\bibfield{author}{\bibinfo{person}{Lukas Hoyer}, \bibinfo{person}{Dengxin Dai}, {and} \bibinfo{person}{Luc Van~Gool}.} \bibinfo{year}{2022}\natexlab{a}.
\newblock \showarticletitle{Daformer: Improving network architectures and training strategies for domain-adaptive semantic segmentation}. In \bibinfo{booktitle}{\emph{Proc. of IEEE/CVF Conference on Computer Vision and Pattern Recognition}}. \bibinfo{pages}{9924--9935}.
\newblock


\bibitem[Hoyer et~al\mbox{.}(2022b)]%
        {hoyer2022hrda}
\bibfield{author}{\bibinfo{person}{Lukas Hoyer}, \bibinfo{person}{Dengxin Dai}, {and} \bibinfo{person}{Luc Van~Gool}.} \bibinfo{year}{2022}\natexlab{b}.
\newblock \showarticletitle{HRDA: Context-aware high-resolution domain-adaptive semantic segmentation}. In \bibinfo{booktitle}{\emph{European Conference on Computer Vision}}. Springer, \bibinfo{pages}{372--391}.
\newblock


\bibitem[Hoyer et~al\mbox{.}(2023a)]%
        {hoyer2023domain}
\bibfield{author}{\bibinfo{person}{Lukas Hoyer}, \bibinfo{person}{Dengxin Dai}, {and} \bibinfo{person}{Luc Van~Gool}.} \bibinfo{year}{2023}\natexlab{a}.
\newblock \showarticletitle{Domain Adaptive and Generalizable Network Architectures and Training Strategies for Semantic Image Segmentation}.
\newblock \bibinfo{journal}{\emph{IEEE Transactions on Pattern Analysis and Machine Intelligence}} (\bibinfo{year}{2023}).
\newblock


\bibitem[Hoyer et~al\mbox{.}(2023b)]%
        {hoyer2023mic}
\bibfield{author}{\bibinfo{person}{Lukas Hoyer}, \bibinfo{person}{Dengxin Dai}, \bibinfo{person}{Haoran Wang}, {and} \bibinfo{person}{Luc Van~Gool}.} \bibinfo{year}{2023}\natexlab{b}.
\newblock \showarticletitle{MIC: Masked image consistency for context-enhanced domain adaptation}. In \bibinfo{booktitle}{\emph{Proc. of IEEE/CVF Conference on Computer Vision and Pattern Recognition}}. \bibinfo{pages}{11721--11732}.
\newblock


\bibitem[Jeong et~al\mbox{.}(2024)]%
        {jeong2024revisiting}
\bibfield{author}{\bibinfo{person}{Seongwon Jeong}, \bibinfo{person}{Jiyeong Kim}, \bibinfo{person}{Sungheui Kim}, {and} \bibinfo{person}{Dongbo Min}.} \bibinfo{year}{2024}\natexlab{}.
\newblock \showarticletitle{Revisiting Domain-Adaptive Semantic Segmentation via Knowledge Distillation}.
\newblock \bibinfo{journal}{\emph{IEEE Transactions on Image Processing}} (\bibinfo{year}{2024}).
\newblock


\bibitem[Jin et~al\mbox{.}(2022)]%
        {jin2022semi}
\bibfield{author}{\bibinfo{person}{Ying Jin}, \bibinfo{person}{Jiaqi Wang}, {and} \bibinfo{person}{Dahua Lin}.} \bibinfo{year}{2022}\natexlab{}.
\newblock \showarticletitle{Semi-supervised semantic segmentation via gentle teaching assistant}.
\newblock \bibinfo{journal}{\emph{Advances in Neural Information Processing Systems}}  \bibinfo{volume}{35} (\bibinfo{year}{2022}), \bibinfo{pages}{2803--2816}.
\newblock


\bibitem[Kalluri et~al\mbox{.}(2024)]%
        {kalluri2024uda}
\bibfield{author}{\bibinfo{person}{Tarun Kalluri}, \bibinfo{person}{Sreyas Ravichandran}, {and} \bibinfo{person}{Manmohan Chandraker}.} \bibinfo{year}{2024}\natexlab{}.
\newblock \showarticletitle{UDA-Bench: Revisiting Common Assumptions in Unsupervised Domain Adaptation Using a Standardized Framework}.
\newblock \bibinfo{journal}{\emph{arXiv preprint arXiv:2409.15264}} (\bibinfo{year}{2024}).
\newblock


\bibitem[Kim et~al\mbox{.}(2021)]%
        {kim2021comparing}
\bibfield{author}{\bibinfo{person}{Taehyeon Kim}, \bibinfo{person}{Jaehoon Oh}, \bibinfo{person}{NakYil Kim}, \bibinfo{person}{Sangwook Cho}, {and} \bibinfo{person}{Se-Young Yun}.} \bibinfo{year}{2021}\natexlab{}.
\newblock \showarticletitle{Comparing kullback-leibler divergence and mean squared error loss in knowledge distillation}.
\newblock \bibinfo{journal}{\emph{arXiv preprint arXiv:2105.08919}} (\bibinfo{year}{2021}).
\newblock


\bibitem[Kothandaraman et~al\mbox{.}(2021)]%
        {kothandaraman2021domain}
\bibfield{author}{\bibinfo{person}{Divya Kothandaraman}, \bibinfo{person}{Athira Nambiar}, {and} \bibinfo{person}{Anurag Mittal}.} \bibinfo{year}{2021}\natexlab{}.
\newblock \showarticletitle{Domain adaptive knowledge distillation for driving scene semantic segmentation}. In \bibinfo{booktitle}{\emph{Proc. of IEEE/CVF Winter Conference on Applications of Computer Vision}}. \bibinfo{pages}{134--143}.
\newblock


\bibitem[Kuzmin et~al\mbox{.}(2022)]%
        {kuzmin2022fp8}
\bibfield{author}{\bibinfo{person}{Andrey Kuzmin}, \bibinfo{person}{Mart Van~Baalen}, \bibinfo{person}{Yuwei Ren}, \bibinfo{person}{Markus Nagel}, \bibinfo{person}{Jorn Peters}, {and} \bibinfo{person}{Tijmen Blankevoort}.} \bibinfo{year}{2022}\natexlab{}.
\newblock \showarticletitle{Fp8 quantization: The power of the exponent}.
\newblock \bibinfo{journal}{\emph{Advances in Neural Information Processing Systems}}  \bibinfo{volume}{35} (\bibinfo{year}{2022}), \bibinfo{pages}{14651--14662}.
\newblock


\bibitem[Lee et~al\mbox{.}(2021)]%
        {lee2021unsupervised}
\bibfield{author}{\bibinfo{person}{Suhyeon Lee}, \bibinfo{person}{Junhyuk Hyun}, \bibinfo{person}{Hongje Seong}, {and} \bibinfo{person}{Euntai Kim}.} \bibinfo{year}{2021}\natexlab{}.
\newblock \showarticletitle{Unsupervised domain adaptation for semantic segmentation by content transfer}. In \bibinfo{booktitle}{\emph{Proc. of AAAI conference on Artificial Intelligence}}, Vol.~\bibinfo{volume}{35}. \bibinfo{pages}{8306--8315}.
\newblock


\bibitem[Li et~al\mbox{.}(2022)]%
        {li2022class}
\bibfield{author}{\bibinfo{person}{Ruihuang Li}, \bibinfo{person}{Shuai Li}, \bibinfo{person}{Chenhang He}, \bibinfo{person}{Yabin Zhang}, \bibinfo{person}{Xu Jia}, {and} \bibinfo{person}{Lei Zhang}.} \bibinfo{year}{2022}\natexlab{}.
\newblock \showarticletitle{Class-balanced pixel-level self-labeling for domain adaptive semantic segmentation}. In \bibinfo{booktitle}{\emph{Proc. of IEEE/CVF conference on computer vision and pattern recognition}}. \bibinfo{pages}{11593--11603}.
\newblock


\bibitem[Li et~al\mbox{.}(2021)]%
        {li2021bi}
\bibfield{author}{\bibinfo{person}{Shuang Li}, \bibinfo{person}{Fangrui Lv}, \bibinfo{person}{Binhui Xie}, \bibinfo{person}{Chi~Harold Liu}, \bibinfo{person}{Jian Liang}, {and} \bibinfo{person}{Chen Qin}.} \bibinfo{year}{2021}\natexlab{}.
\newblock \showarticletitle{Bi-classifier determinacy maximization for unsupervised domain adaptation}. In \bibinfo{booktitle}{\emph{Proc. of AAAI conference on artificial intelligence}}, Vol.~\bibinfo{volume}{35}. \bibinfo{pages}{8455--8464}.
\newblock


\bibitem[Li et~al\mbox{.}(2023)]%
        {li2023contrast}
\bibfield{author}{\bibinfo{person}{Tianyu Li}, \bibinfo{person}{Subhankar Roy}, \bibinfo{person}{Huayi Zhou}, \bibinfo{person}{Hongtao Lu}, {and} \bibinfo{person}{St{\'e}phane Lathuili{\`e}re}.} \bibinfo{year}{2023}\natexlab{}.
\newblock \showarticletitle{Contrast, Stylize and Adapt: Unsupervised Contrastive Learning Framework for Domain Adaptive Semantic Segmentation}. In \bibinfo{booktitle}{\emph{Proc. of IEEE/CVF Conference on Computer Vision and Pattern Recognition Workshop}}. \bibinfo{pages}{4868--4878}.
\newblock


\bibitem[Liao et~al\mbox{.}(2023)]%
        {liao2023geometry}
\bibfield{author}{\bibinfo{person}{Yinghong Liao}, \bibinfo{person}{Wending Zhou}, \bibinfo{person}{Xu Yan}, \bibinfo{person}{Zhen Li}, \bibinfo{person}{Yizhou Yu}, {and} \bibinfo{person}{Shuguang Cui}.} \bibinfo{year}{2023}\natexlab{}.
\newblock \showarticletitle{Geometry-aware network for domain adaptive semantic segmentation}. In \bibinfo{booktitle}{\emph{Proc. of AAAI Conference on Artificial Intelligence}}, Vol.~\bibinfo{volume}{37}. \bibinfo{pages}{8755--8763}.
\newblock


\bibitem[Lim and Kim(2024)]%
        {lim2024cross}
\bibfield{author}{\bibinfo{person}{Jeongkee Lim} {and} \bibinfo{person}{Yusung Kim}.} \bibinfo{year}{2024}\natexlab{}.
\newblock \showarticletitle{Cross-Domain Semantic Segmentation on Inconsistent Taxonomy using VLMs}.
\newblock \bibinfo{journal}{\emph{arXiv preprint arXiv:2408.02261}} (\bibinfo{year}{2024}).
\newblock


\bibitem[Liu et~al\mbox{.}(2021)]%
        {liu2021bapa}
\bibfield{author}{\bibinfo{person}{Yahao Liu}, \bibinfo{person}{Jinhong Deng}, \bibinfo{person}{Xinchen Gao}, \bibinfo{person}{Wen Li}, {and} \bibinfo{person}{Lixin Duan}.} \bibinfo{year}{2021}\natexlab{}.
\newblock \showarticletitle{Bapa-net: Boundary adaptation and prototype alignment for cross-domain semantic segmentation}. In \bibinfo{booktitle}{\emph{Proc. of IEEE/CVF international conference on computer vision}}. \bibinfo{pages}{8801--8811}.
\newblock


\bibitem[Loiseau et~al\mbox{.}(2025)]%
        {loiseau2025reliability}
\bibfield{author}{\bibinfo{person}{Thibaut Loiseau}, \bibinfo{person}{Tuan-Hung Vu}, \bibinfo{person}{Mickael Chen}, \bibinfo{person}{Patrick P{\'e}rez}, {and} \bibinfo{person}{Matthieu Cord}.} \bibinfo{year}{2025}\natexlab{}.
\newblock \showarticletitle{Reliability in semantic segmentation: Can we use synthetic data?}. In \bibinfo{booktitle}{\emph{European Conference on Computer Vision}}. Springer, \bibinfo{pages}{442--459}.
\newblock


\bibitem[Meng et~al\mbox{.}({[n.\,d.]})]%
        {mengslimmable}
\bibfield{author}{\bibinfo{person}{Rang Meng}, \bibinfo{person}{Weijie Chen}, \bibinfo{person}{Shicai Yang}, \bibinfo{person}{Jie Song}, \bibinfo{person}{Luojun Lin}, \bibinfo{person}{Di Xie}, \bibinfo{person}{Shiliang Pu}, \bibinfo{person}{Xinchao Wang}, \bibinfo{person}{Mingli Song}, {and} \bibinfo{person}{Yueting Zhuang}.} \bibinfo{year}{[n.\,d.]}\natexlab{}.
\newblock \showarticletitle{Slimmable Domain Adaptation (Supplementary Materials)}.
\newblock  (\bibinfo{year}{[n.\,d.]}).
\newblock


\bibitem[Murez et~al\mbox{.}(2018)]%
        {murez2018image}
\bibfield{author}{\bibinfo{person}{Zak Murez}, \bibinfo{person}{Soheil Kolouri}, \bibinfo{person}{David Kriegman}, \bibinfo{person}{Ravi Ramamoorthi}, {and} \bibinfo{person}{Kyungnam Kim}.} \bibinfo{year}{2018}\natexlab{}.
\newblock \showarticletitle{Image to image translation for domain adaptation}. In \bibinfo{booktitle}{\emph{Proc. of IEEE conference on computer vision and pattern recognition}}. \bibinfo{pages}{4500--4509}.
\newblock


\bibitem[Oh et~al\mbox{.}(2022)]%
        {oh2022non}
\bibfield{author}{\bibinfo{person}{Sangyun Oh}, \bibinfo{person}{Hyeonuk Sim}, \bibinfo{person}{Jounghyun Kim}, {and} \bibinfo{person}{Jongeun Lee}.} \bibinfo{year}{2022}\natexlab{}.
\newblock \showarticletitle{Non-Uniform Step Size Quantization for Accurate Post-Training Quantization}. In \bibinfo{booktitle}{\emph{European Conference on Computer Vision}}. Springer, \bibinfo{pages}{658--673}.
\newblock


\bibitem[Pan et~al\mbox{.}(2024)]%
        {pan2024moda}
\bibfield{author}{\bibinfo{person}{Fei Pan}, \bibinfo{person}{Xu Yin}, \bibinfo{person}{Seokju Lee}, \bibinfo{person}{Axi Niu}, \bibinfo{person}{Sungeui Yoon}, {and} \bibinfo{person}{In~So Kweon}.} \bibinfo{year}{2024}\natexlab{}.
\newblock \showarticletitle{MoDA: Leveraging Motion Priors from Videos for Advancing Unsupervised Domain Adaptation in Semantic Segmentation}. In \bibinfo{booktitle}{\emph{Proc. of IEEE/CVF Conference on Computer Vision and Pattern Recognition}}. \bibinfo{pages}{2649--2658}.
\newblock


\bibitem[Peng et~al\mbox{.}(2023)]%
        {peng2023diffusion}
\bibfield{author}{\bibinfo{person}{Duo Peng}, \bibinfo{person}{Ping Hu}, \bibinfo{person}{Qiuhong Ke}, {and} \bibinfo{person}{Jun Liu}.} \bibinfo{year}{2023}\natexlab{}.
\newblock \showarticletitle{Diffusion-based Image Translation with Label Guidance for Domain Adaptive Semantic Segmentation}. In \bibinfo{booktitle}{\emph{Proc. of IEEE/CVF International Conference on Computer Vision}}. \bibinfo{pages}{808--820}.
\newblock


\bibitem[Qiu et~al\mbox{.}({[n.\,d.]})]%
        {qiu2024make}
\bibfield{author}{\bibinfo{person}{Shoumeng Qiu}, \bibinfo{person}{Jie Chen}, \bibinfo{person}{Xinrun Li}, \bibinfo{person}{Ru Wan}, {and} \bibinfo{person}{Xiangyang Xue}.} \bibinfo{year}{[n.\,d.]}\natexlab{}.
\newblock \showarticletitle{Make a Strong Teacher with Label Assistance: A Novel Knowledge Distillation Approach for Semantic Segmentation}.
\newblock  (\bibinfo{year}{[n.\,d.]}).
\newblock


\bibitem[Ren et~al\mbox{.}(2024)]%
        {ren2024cross}
\bibfield{author}{\bibinfo{person}{Wenqi Ren}, \bibinfo{person}{Ruihao Xia}, \bibinfo{person}{Meng Zheng}, \bibinfo{person}{Ziyan Wu}, \bibinfo{person}{Yang Tang}, {and} \bibinfo{person}{Nicu Sebe}.} \bibinfo{year}{2024}\natexlab{}.
\newblock \showarticletitle{Cross-Class Domain Adaptive Semantic Segmentation with Visual Language Models}. In \bibinfo{booktitle}{\emph{Proc. of 32nd ACM International Conference on Multimedia}}. \bibinfo{pages}{5005--5014}.
\newblock


\bibitem[Richter et~al\mbox{.}(2016)]%
        {richter2016playing}
\bibfield{author}{\bibinfo{person}{Stephan~R Richter}, \bibinfo{person}{Vibhav Vineet}, \bibinfo{person}{Stefan Roth}, {and} \bibinfo{person}{Vladlen Koltun}.} \bibinfo{year}{2016}\natexlab{}.
\newblock \showarticletitle{Playing for data: Ground truth from computer games}. In \bibinfo{booktitle}{\emph{Computer Vision--ECCV 2016: 14th European Conference, Amsterdam, The Netherlands, October 11-14, 2016, Proceedings, Part II 14}}. Springer, \bibinfo{pages}{102--118}.
\newblock


\bibitem[Ros et~al\mbox{.}(2016)]%
        {ros2016synthia}
\bibfield{author}{\bibinfo{person}{German Ros}, \bibinfo{person}{Laura Sellart}, \bibinfo{person}{Joanna Materzynska}, \bibinfo{person}{David Vazquez}, {and} \bibinfo{person}{Antonio~M Lopez}.} \bibinfo{year}{2016}\natexlab{}.
\newblock \showarticletitle{The synthia dataset: A large collection of synthetic images for semantic segmentation of urban scenes}. In \bibinfo{booktitle}{\emph{Proc. of IEEE conference on computer vision and pattern recognition}}. \bibinfo{pages}{3234--3243}.
\newblock


\bibitem[Saha et~al\mbox{.}(2021)]%
        {saha2021learning}
\bibfield{author}{\bibinfo{person}{Suman Saha}, \bibinfo{person}{Anton Obukhov}, \bibinfo{person}{Danda~Pani Paudel}, \bibinfo{person}{Menelaos Kanakis}, \bibinfo{person}{Yuhua Chen}, \bibinfo{person}{Stamatios Georgoulis}, {and} \bibinfo{person}{Luc Van~Gool}.} \bibinfo{year}{2021}\natexlab{}.
\newblock \showarticletitle{Learning to relate depth and semantics for unsupervised domain adaptation}. In \bibinfo{booktitle}{\emph{Proc. of IEEE/CVF Conference on Computer Vision and Pattern Recognition}}. \bibinfo{pages}{8197--8207}.
\newblock


\bibitem[Saito et~al\mbox{.}(2018)]%
        {saito2018maximum}
\bibfield{author}{\bibinfo{person}{Kuniaki Saito}, \bibinfo{person}{Kohei Watanabe}, \bibinfo{person}{Yoshitaka Ushiku}, {and} \bibinfo{person}{Tatsuya Harada}.} \bibinfo{year}{2018}\natexlab{}.
\newblock \showarticletitle{Maximum classifier discrepancy for unsupervised domain adaptation}. In \bibinfo{booktitle}{\emph{Proc. of IEEE conference on computer vision and pattern recognition}}. \bibinfo{pages}{3723--3732}.
\newblock


\bibitem[Sakaridis et~al\mbox{.}(2020)]%
        {sakaridis2020map}
\bibfield{author}{\bibinfo{person}{Christos Sakaridis}, \bibinfo{person}{Dengxin Dai}, {and} \bibinfo{person}{Luc Van~Gool}.} \bibinfo{year}{2020}\natexlab{}.
\newblock \showarticletitle{Map-guided curriculum domain adaptation and uncertainty-aware evaluation for semantic nighttime image segmentation}.
\newblock \bibinfo{journal}{\emph{IEEE Transactions on Pattern Analysis and Machine Intelligence}} \bibinfo{volume}{44}, \bibinfo{number}{6} (\bibinfo{year}{2020}), \bibinfo{pages}{3139--3153}.
\newblock


\bibitem[Sakaridis et~al\mbox{.}(2021)]%
        {sakaridis2021acdc}
\bibfield{author}{\bibinfo{person}{Christos Sakaridis}, \bibinfo{person}{Dengxin Dai}, {and} \bibinfo{person}{Luc Van~Gool}.} \bibinfo{year}{2021}\natexlab{}.
\newblock \showarticletitle{ACDC: The adverse conditions dataset with correspondences for semantic driving scene understanding}. In \bibinfo{booktitle}{\emph{Proc. of IEEE/CVF International Conference on Computer Vision}}. \bibinfo{pages}{10765--10775}.
\newblock


\bibitem[Schwonberg et~al\mbox{.}(2023)]%
        {schwonberg2023survey}
\bibfield{author}{\bibinfo{person}{Manuel Schwonberg}, \bibinfo{person}{Joshua Niemeijer}, \bibinfo{person}{Jan-Aike Term{\"o}hlen}, \bibinfo{person}{J{\"o}rg~P Sch{\"a}fer}, \bibinfo{person}{Nico~M Schmidt}, \bibinfo{person}{Hanno Gottschalk}, {and} \bibinfo{person}{Tim Fingscheidt}.} \bibinfo{year}{2023}\natexlab{}.
\newblock \showarticletitle{Survey on Unsupervised Domain Adaptation for Semantic Segmentation for Visual Perception in Automated Driving}.
\newblock \bibinfo{journal}{\emph{IEEE Access}} (\bibinfo{year}{2023}).
\newblock


\bibitem[Shen et~al\mbox{.}(2023)]%
        {shen2023diga}
\bibfield{author}{\bibinfo{person}{Fengyi Shen}, \bibinfo{person}{Akhil Gurram}, \bibinfo{person}{Ziyuan Liu}, \bibinfo{person}{He Wang}, {and} \bibinfo{person}{Alois Knoll}.} \bibinfo{year}{2023}\natexlab{}.
\newblock \showarticletitle{DiGA: Distil to Generalize and then Adapt for Domain Adaptive Semantic Segmentation}. In \bibinfo{booktitle}{\emph{Proc. of IEEE/CVF Conference on Computer Vision and Pattern Recognition}}. \bibinfo{pages}{15866--15877}.
\newblock


\bibitem[Shu et~al\mbox{.}(2021)]%
        {shu2021channel}
\bibfield{author}{\bibinfo{person}{Changyong Shu}, \bibinfo{person}{Yifan Liu}, \bibinfo{person}{Jianfei Gao}, \bibinfo{person}{Zheng Yan}, {and} \bibinfo{person}{Chunhua Shen}.} \bibinfo{year}{2021}\natexlab{}.
\newblock \showarticletitle{Channel-wise knowledge distillation for dense prediction}. In \bibinfo{booktitle}{\emph{Proc. of IEEE/CVF International Conference on Computer Vision}}. \bibinfo{pages}{5311--5320}.
\newblock


\bibitem[Son et~al\mbox{.}(2021)]%
        {son2021densely}
\bibfield{author}{\bibinfo{person}{Wonchul Son}, \bibinfo{person}{Jaemin Na}, \bibinfo{person}{Junyong Choi}, {and} \bibinfo{person}{Wonjun Hwang}.} \bibinfo{year}{2021}\natexlab{}.
\newblock \showarticletitle{Densely guided knowledge distillation using multiple teacher assistants}. In \bibinfo{booktitle}{\emph{Proc. of IEEE/CVF International Conference on Computer Vision}}. \bibinfo{pages}{9395--9404}.
\newblock


\bibitem[Tang et~al\mbox{.}(2023)]%
        {tang2023dynamic}
\bibfield{author}{\bibinfo{person}{Quan Tang}, \bibinfo{person}{Bowen Zhang}, \bibinfo{person}{Jiajun Liu}, \bibinfo{person}{Fagui Liu}, {and} \bibinfo{person}{Yifan Liu}.} \bibinfo{year}{2023}\natexlab{}.
\newblock \showarticletitle{Dynamic Token Pruning in Plain Vision Transformers for Semantic Segmentation}. In \bibinfo{booktitle}{\emph{Proc. of IEEE/CVF International Conference on Computer Vision}}. \bibinfo{pages}{777--786}.
\newblock


\bibitem[Tarvainen and Valpola(2017)]%
        {tarvainen2017mean}
\bibfield{author}{\bibinfo{person}{Antti Tarvainen} {and} \bibinfo{person}{Harri Valpola}.} \bibinfo{year}{2017}\natexlab{}.
\newblock \showarticletitle{Mean teachers are better role models: Weight-averaged consistency targets improve semi-supervised deep learning results}.
\newblock \bibinfo{journal}{\emph{Advances in neural information processing systems}}  \bibinfo{volume}{30} (\bibinfo{year}{2017}).
\newblock


\bibitem[Tranheden et~al\mbox{.}(2021)]%
        {tranheden2021dacs}
\bibfield{author}{\bibinfo{person}{Wilhelm Tranheden}, \bibinfo{person}{Viktor Olsson}, \bibinfo{person}{Juliano Pinto}, {and} \bibinfo{person}{Lennart Svensson}.} \bibinfo{year}{2021}\natexlab{}.
\newblock \showarticletitle{Dacs: Domain adaptation via cross-domain mixed sampling}. In \bibinfo{booktitle}{\emph{Proc. of IEEE/CVF Winter Conference on Applications of Computer Vision}}. \bibinfo{pages}{1379--1389}.
\newblock


\bibitem[Truong et~al\mbox{.}(2023)]%
        {truong2023fredom}
\bibfield{author}{\bibinfo{person}{Thanh-Dat Truong}, \bibinfo{person}{Ngan Le}, \bibinfo{person}{Bhiksha Raj}, \bibinfo{person}{Jackson Cothren}, {and} \bibinfo{person}{Khoa Luu}.} \bibinfo{year}{2023}\natexlab{}.
\newblock \showarticletitle{Fredom: Fairness domain adaptation approach to semantic scene understanding}. In \bibinfo{booktitle}{\emph{Proc. of IEEE/CVF Conference on Computer Vision and Pattern Recognition}}. \bibinfo{pages}{19988--19997}.
\newblock


\bibitem[Tsai et~al\mbox{.}(2019)]%
        {tsai2019domain}
\bibfield{author}{\bibinfo{person}{Yi-Hsuan Tsai}, \bibinfo{person}{Kihyuk Sohn}, \bibinfo{person}{Samuel Schulter}, {and} \bibinfo{person}{Manmohan Chandraker}.} \bibinfo{year}{2019}\natexlab{}.
\newblock \showarticletitle{Domain adaptation for structured output via discriminative patch representations}. In \bibinfo{booktitle}{\emph{Proc. of IEEE/CVF International Conference on Computer Vision}}. \bibinfo{pages}{1456--1465}.
\newblock


\bibitem[Vu et~al\mbox{.}(2019)]%
        {vu2019advent}
\bibfield{author}{\bibinfo{person}{Tuan-Hung Vu}, \bibinfo{person}{Himalaya Jain}, \bibinfo{person}{Maxime Bucher}, \bibinfo{person}{Matthieu Cord}, {and} \bibinfo{person}{Patrick P{\'e}rez}.} \bibinfo{year}{2019}\natexlab{}.
\newblock \showarticletitle{Advent: Adversarial entropy minimization for domain adaptation in semantic segmentation}. In \bibinfo{booktitle}{\emph{Proc. of IEEE/CVF conference on computer vision and pattern recognition}}. \bibinfo{pages}{2517--2526}.
\newblock


\bibitem[Wang et~al\mbox{.}(2023b)]%
        {wang2023cdac}
\bibfield{author}{\bibinfo{person}{Kaihong Wang}, \bibinfo{person}{Donghyun Kim}, \bibinfo{person}{Rogerio Feris}, {and} \bibinfo{person}{Margrit Betke}.} \bibinfo{year}{2023}\natexlab{b}.
\newblock \showarticletitle{CDAC: Cross-domain Attention Consistency in Transformer for Domain Adaptive Semantic Segmentation}. In \bibinfo{booktitle}{\emph{Proc. of IEEE/CVF International Conference on Computer Vision}}. \bibinfo{pages}{11519--11529}.
\newblock


\bibitem[Wang et~al\mbox{.}(2023d)]%
        {wang2023informative}
\bibfield{author}{\bibinfo{person}{Shiqin Wang}, \bibinfo{person}{Xin Xu}, \bibinfo{person}{Xianzheng Ma}, \bibinfo{person}{Kui Jiang}, {and} \bibinfo{person}{Zheng Wang}.} \bibinfo{year}{2023}\natexlab{d}.
\newblock \showarticletitle{Informative Classes Matter: Towards Unsupervised Domain Adaptive Nighttime Semantic Segmentation}. In \bibinfo{booktitle}{\emph{Proc. of 31st ACM International Conference on Multimedia}}. \bibinfo{pages}{163--172}.
\newblock


\bibitem[Wang et~al\mbox{.}(2023a)]%
        {wang2023balancing}
\bibfield{author}{\bibinfo{person}{Yuchao Wang}, \bibinfo{person}{Jingjing Fei}, \bibinfo{person}{Haochen Wang}, \bibinfo{person}{Wei Li}, \bibinfo{person}{Tianpeng Bao}, \bibinfo{person}{Liwei Wu}, \bibinfo{person}{Rui Zhao}, {and} \bibinfo{person}{Yujun Shen}.} \bibinfo{year}{2023}\natexlab{a}.
\newblock \showarticletitle{Balancing Logit Variation for Long-tailed Semantic Segmentation}. In \bibinfo{booktitle}{\emph{Proc. of IEEE/CVF Conference on Computer Vision and Pattern Recognition}}. \bibinfo{pages}{19561--19573}.
\newblock


\bibitem[Wang et~al\mbox{.}(2021)]%
        {wang2021uncertainty}
\bibfield{author}{\bibinfo{person}{Yuxi Wang}, \bibinfo{person}{Junran Peng}, {and} \bibinfo{person}{ZhaoXiang Zhang}.} \bibinfo{year}{2021}\natexlab{}.
\newblock \showarticletitle{Uncertainty-aware pseudo label refinery for domain adaptive semantic segmentation}. In \bibinfo{booktitle}{\emph{Proc. of IEEE/CVF international conference on computer vision}}. \bibinfo{pages}{9092--9101}.
\newblock


\bibitem[Wang et~al\mbox{.}(2023c)]%
        {wang2023cal}
\bibfield{author}{\bibinfo{person}{Zixin Wang}, \bibinfo{person}{Yadan Luo}, \bibinfo{person}{Zhi Chen}, \bibinfo{person}{Sen Wang}, {and} \bibinfo{person}{Zi Huang}.} \bibinfo{year}{2023}\natexlab{c}.
\newblock \showarticletitle{Cal-SFDA: Source-free domain-adaptive semantic segmentation with differentiable expected calibration error}. In \bibinfo{booktitle}{\emph{Proc. of 31st ACM International Conference on Multimedia}}. \bibinfo{pages}{1167--1178}.
\newblock


\bibitem[Wen et~al\mbox{.}(2024)]%
        {wen2024cdea}
\bibfield{author}{\bibinfo{person}{Shuyuan Wen}, \bibinfo{person}{Bingrui Hu}, {and} \bibinfo{person}{Wenchao Li}.} \bibinfo{year}{2024}\natexlab{}.
\newblock \showarticletitle{CDEA: Context-and Detail-Enhanced Unsupervised Learning for Domain Adaptive Semantic Segmentation}. In \bibinfo{booktitle}{\emph{Proc. of 32nd ACM International Conference on Multimedia}}. \bibinfo{pages}{2786--2794}.
\newblock


\bibitem[Wu et~al\mbox{.}(2021)]%
        {wu2021dannet}
\bibfield{author}{\bibinfo{person}{Xinyi Wu}, \bibinfo{person}{Zhenyao Wu}, \bibinfo{person}{Hao Guo}, \bibinfo{person}{Lili Ju}, {and} \bibinfo{person}{Song Wang}.} \bibinfo{year}{2021}\natexlab{}.
\newblock \showarticletitle{Dannet: A one-stage domain adaptation network for unsupervised nighttime semantic segmentation}. In \bibinfo{booktitle}{\emph{Proc. of IEEE/CVF Conference on Computer Vision and Pattern Recognition}}. \bibinfo{pages}{15769--15778}.
\newblock


\bibitem[Xie et~al\mbox{.}(2023)]%
        {xie2023sepico}
\bibfield{author}{\bibinfo{person}{Binhui Xie}, \bibinfo{person}{Shuang Li}, \bibinfo{person}{Mingjia Li}, \bibinfo{person}{Chi~Harold Liu}, \bibinfo{person}{Gao Huang}, {and} \bibinfo{person}{Guoren Wang}.} \bibinfo{year}{2023}\natexlab{}.
\newblock \showarticletitle{Sepico: Semantic-guided pixel contrast for domain adaptive semantic segmentation}.
\newblock \bibinfo{journal}{\emph{IEEE Transactions on Pattern Analysis and Machine Intelligence}} (\bibinfo{year}{2023}).
\newblock


\bibitem[Xie et~al\mbox{.}(2021)]%
        {xie2021segformer}
\bibfield{author}{\bibinfo{person}{Enze Xie}, \bibinfo{person}{Wenhai Wang}, \bibinfo{person}{Zhiding Yu}, \bibinfo{person}{Anima Anandkumar}, \bibinfo{person}{Jose~M Alvarez}, {and} \bibinfo{person}{Ping Luo}.} \bibinfo{year}{2021}\natexlab{}.
\newblock \showarticletitle{SegFormer: Simple and efficient design for semantic segmentation with transformers}.
\newblock \bibinfo{journal}{\emph{Advances in Neural Information Processing Systems}}  \bibinfo{volume}{34} (\bibinfo{year}{2021}), \bibinfo{pages}{12077--12090}.
\newblock


\bibitem[Yan et~al\mbox{.}(2023)]%
        {yan2023sam4udass}
\bibfield{author}{\bibinfo{person}{Weihao Yan}, \bibinfo{person}{Yeqiang Qian}, \bibinfo{person}{Hanyang Zhuang}, \bibinfo{person}{Chunxiang Wang}, {and} \bibinfo{person}{Ming Yang}.} \bibinfo{year}{2023}\natexlab{}.
\newblock \showarticletitle{Sam4udass: When sam meets unsupervised domain adaptive semantic segmentation in intelligent vehicles}.
\newblock \bibinfo{journal}{\emph{IEEE Transactions on Intelligent Vehicles}} \bibinfo{volume}{9}, \bibinfo{number}{2} (\bibinfo{year}{2023}), \bibinfo{pages}{3396--3408}.
\newblock


\bibitem[Yang et~al\mbox{.}(2025)]%
        {yang2025micdrop}
\bibfield{author}{\bibinfo{person}{Linyan Yang}, \bibinfo{person}{Lukas Hoyer}, \bibinfo{person}{Mark Weber}, \bibinfo{person}{Tobias Fischer}, \bibinfo{person}{Dengxin Dai}, \bibinfo{person}{Laura Leal-Taix{\'e}}, \bibinfo{person}{Marc Pollefeys}, \bibinfo{person}{Daniel Cremers}, {and} \bibinfo{person}{Luc Van~Gool}.} \bibinfo{year}{2025}\natexlab{}.
\newblock \showarticletitle{Micdrop: masking image and depth features via complementary dropout for domain-adaptive semantic segmentation}. In \bibinfo{booktitle}{\emph{European Conference on Computer Vision}}. Springer, \bibinfo{pages}{329--346}.
\newblock


\bibitem[Yang and Soatto(2020)]%
        {yang2020fda}
\bibfield{author}{\bibinfo{person}{Yanchao Yang} {and} \bibinfo{person}{Stefano Soatto}.} \bibinfo{year}{2020}\natexlab{}.
\newblock \showarticletitle{Fda: Fourier domain adaptation for semantic segmentation}. In \bibinfo{booktitle}{\emph{Proc. of IEEE/CVF conference on computer vision and pattern recognition}}. \bibinfo{pages}{4085--4095}.
\newblock


\bibitem[Yu et~al\mbox{.}(2019)]%
        {yu2019accelerating}
\bibfield{author}{\bibinfo{person}{Chaohui Yu}, \bibinfo{person}{Jindong Wang}, \bibinfo{person}{Yiqiang Chen}, {and} \bibinfo{person}{Zijing Wu}.} \bibinfo{year}{2019}\natexlab{}.
\newblock \showarticletitle{Accelerating deep unsupervised domain adaptation with transfer channel pruning}. In \bibinfo{booktitle}{\emph{2019 International Joint Conference on Neural Networks (IJCNN)}}. IEEE, \bibinfo{pages}{1--8}.
\newblock


\bibitem[Yuan et~al\mbox{.}(2020)]%
        {yuan2020revisiting}
\bibfield{author}{\bibinfo{person}{Li Yuan}, \bibinfo{person}{Francis~EH Tay}, \bibinfo{person}{Guilin Li}, \bibinfo{person}{Tao Wang}, {and} \bibinfo{person}{Jiashi Feng}.} \bibinfo{year}{2020}\natexlab{}.
\newblock \showarticletitle{Revisiting knowledge distillation via label smoothing regularization}. In \bibinfo{booktitle}{\emph{Proc. of IEEE/CVF Conference on Computer Vision and Pattern Recognition}}. \bibinfo{pages}{3903--3911}.
\newblock


\bibitem[Yvinec et~al\mbox{.}(2023)]%
        {yvinec2023spiq}
\bibfield{author}{\bibinfo{person}{Edouard Yvinec}, \bibinfo{person}{Arnaud Dapogny}, \bibinfo{person}{Matthieu Cord}, {and} \bibinfo{person}{Kevin Bailly}.} \bibinfo{year}{2023}\natexlab{}.
\newblock \showarticletitle{Spiq: Data-free per-channel static input quantization}. In \bibinfo{booktitle}{\emph{Proc. of IEEE/CVF Winter Conference on Applications of Computer Vision}}. \bibinfo{pages}{3869--3878}.
\newblock


\bibitem[Zhang et~al\mbox{.}(2021)]%
        {zhang2021prototypical}
\bibfield{author}{\bibinfo{person}{Pan Zhang}, \bibinfo{person}{Bo Zhang}, \bibinfo{person}{Ting Zhang}, \bibinfo{person}{Dong Chen}, \bibinfo{person}{Yong Wang}, {and} \bibinfo{person}{Fang Wen}.} \bibinfo{year}{2021}\natexlab{}.
\newblock \showarticletitle{Prototypical pseudo label denoising and target structure learning for domain adaptive semantic segmentation}. In \bibinfo{booktitle}{\emph{Proc. of IEEE/CVF conference on computer vision and pattern recognition}}. \bibinfo{pages}{12414--12424}.
\newblock


\bibitem[Zhang et~al\mbox{.}(2019)]%
        {zhang2019category}
\bibfield{author}{\bibinfo{person}{Qiming Zhang}, \bibinfo{person}{Jing Zhang}, \bibinfo{person}{Wei Liu}, {and} \bibinfo{person}{Dacheng Tao}.} \bibinfo{year}{2019}\natexlab{}.
\newblock \showarticletitle{Category anchor-guided unsupervised domain adaptation for semantic segmentation}.
\newblock \bibinfo{journal}{\emph{Advances in neural information processing systems}}  \bibinfo{volume}{32} (\bibinfo{year}{2019}).
\newblock


\bibitem[Zhao et~al\mbox{.}(2023)]%
        {zhao2023learning}
\bibfield{author}{\bibinfo{person}{Dong Zhao}, \bibinfo{person}{Shuang Wang}, \bibinfo{person}{Qi Zang}, \bibinfo{person}{Dou Quan}, \bibinfo{person}{Xiutiao Ye}, \bibinfo{person}{Rui Yang}, {and} \bibinfo{person}{Licheng Jiao}.} \bibinfo{year}{2023}\natexlab{}.
\newblock \showarticletitle{Learning Pseudo-Relations for Cross-domain Semantic Segmentation}. In \bibinfo{booktitle}{\emph{Proc. of IEEE/CVF International Conference on Computer Vision}}. \bibinfo{pages}{19191--19203}.
\newblock


\bibitem[Zhao et~al\mbox{.}(2024)]%
        {zhao2024unsupervised}
\bibfield{author}{\bibinfo{person}{Xingchen Zhao}, \bibinfo{person}{Niluthpol~Chowdhury Mithun}, \bibinfo{person}{Abhinav Rajvanshi}, \bibinfo{person}{Han-Pang Chiu}, {and} \bibinfo{person}{Supun Samarasekera}.} \bibinfo{year}{2024}\natexlab{}.
\newblock \showarticletitle{Unsupervised domain adaptation for semantic segmentation with pseudo label self-refinement}. In \bibinfo{booktitle}{\emph{Proc. of IEEE/CVF Winter Conference on Applications of Computer Vision}}. \bibinfo{pages}{2399--2409}.
\newblock


\bibitem[Zheng and Yang(2021)]%
        {zheng2021rectifying}
\bibfield{author}{\bibinfo{person}{Zhedong Zheng} {and} \bibinfo{person}{Yi Yang}.} \bibinfo{year}{2021}\natexlab{}.
\newblock \showarticletitle{Rectifying pseudo label learning via uncertainty estimation for domain adaptive semantic segmentation}.
\newblock \bibinfo{journal}{\emph{International Journal of Computer Vision}} \bibinfo{volume}{129}, \bibinfo{number}{4} (\bibinfo{year}{2021}), \bibinfo{pages}{1106--1120}.
\newblock


\bibitem[Zhou et~al\mbox{.}(2022)]%
        {zhou2022context}
\bibfield{author}{\bibinfo{person}{Qianyu Zhou}, \bibinfo{person}{Zhengyang Feng}, \bibinfo{person}{Qiqi Gu}, \bibinfo{person}{Jiangmiao Pang}, \bibinfo{person}{Guangliang Cheng}, \bibinfo{person}{Xuequan Lu}, \bibinfo{person}{Jianping Shi}, {and} \bibinfo{person}{Lizhuang Ma}.} \bibinfo{year}{2022}\natexlab{}.
\newblock \showarticletitle{Context-aware mixup for domain adaptive semantic segmentation}.
\newblock \bibinfo{journal}{\emph{IEEE Transactions on Circuits and Systems for Video Technology}} \bibinfo{volume}{33}, \bibinfo{number}{2} (\bibinfo{year}{2022}), \bibinfo{pages}{804--817}.
\newblock


\bibitem[Zhu et~al\mbox{.}(2023)]%
        {zhu2023good}
\bibfield{author}{\bibinfo{person}{Jinjing Zhu}, \bibinfo{person}{Yunhao Luo}, \bibinfo{person}{Xu Zheng}, \bibinfo{person}{Hao Wang}, {and} \bibinfo{person}{Lin Wang}.} \bibinfo{year}{2023}\natexlab{}.
\newblock \showarticletitle{A Good Student is Cooperative and Reliable: CNN-Transformer Collaborative Learning for Semantic Segmentation}. In \bibinfo{booktitle}{\emph{Proc. of IEEE/CVF International Conference on Computer Vision}}. \bibinfo{pages}{11720--11730}.
\newblock


\bibitem[Zou et~al\mbox{.}(2018)]%
        {zou2018unsupervised}
\bibfield{author}{\bibinfo{person}{Yang Zou}, \bibinfo{person}{Zhiding Yu}, \bibinfo{person}{BVK Kumar}, {and} \bibinfo{person}{Jinsong Wang}.} \bibinfo{year}{2018}\natexlab{}.
\newblock \showarticletitle{Unsupervised domain adaptation for semantic segmentation via class-balanced self-training}. In \bibinfo{booktitle}{\emph{Proc. of European conference on computer vision (ECCV)}}. \bibinfo{pages}{289--305}.
\newblock


\bibitem[Zou et~al\mbox{.}(2019)]%
        {zou2019confidence}
\bibfield{author}{\bibinfo{person}{Yang Zou}, \bibinfo{person}{Zhiding Yu}, \bibinfo{person}{Xiaofeng Liu}, \bibinfo{person}{BVK Kumar}, {and} \bibinfo{person}{Jinsong Wang}.} \bibinfo{year}{2019}\natexlab{}.
\newblock \showarticletitle{Confidence regularized self-training}. In \bibinfo{booktitle}{\emph{Proc. of IEEE/CVF international conference on computer vision}}. \bibinfo{pages}{5982--5991}.
\newblock


\end{thebibliography}
